\documentclass{article}
\usepackage{ijcai22}

\usepackage{times}

\usepackage{soul}
\usepackage{url}
\usepackage[hidelinks]{hyperref}
\usepackage[utf8]{inputenc}
\usepackage[small]{caption}
\usepackage{graphicx}
\usepackage{amsmath}
\usepackage{booktabs}
\title{Option Transfer and SMDP Abstraction with Successor Features}

\author{
Dongge Han$^1$\and
Sebastian Tschiatschek$^2$\\
\affiliations
$^1$University of Oxford\\ 
$^2$University of Vienna, Faculty of Computer Science, Vienna, Austria\\
\emails
dongge.han.oxford@gmail.com,
sebastian.tschiatschek@univie.ac.at
}

\makeatletter
\def\blfootnote{\xdef\@thefnmark{}\@footnotetext}
\makeatother

\newcommand{\irlnaive}{\textsc{IRL-naive}}
\newcommand{\irlbatch}{\textsc{IRL-batch}}
\newcommand{\ok}{\textsc{OK}}

\newcommand\sbullet[1][.8]{\mathbin{\vcenter{\hbox{\scalebox{#1}{$\bullet$}}}}}

\usepackage{times}
\usepackage{amsmath,amsthm,amssymb,mathtools,nccmath}
\usepackage{subfig}
\usepackage{booktabs}
\usepackage[normalem]{ulem}
\useunder{\uline}{\ul}{}
\usepackage{multirow}
\usepackage{bbm, bm}
\usepackage{enumitem}
\usepackage{multibib}
\usepackage{silence}
\usepackage{algpseudocode}
\usepackage{algorithm}
\usepackage{todonotes}

\newcommand{\mdp}{\mathcal{M}}
\newcommand{\abstractmdp}{\bar{\mathcal{M}}}
\newcommand{\states}{\ensuremath{\mathcal{S}}}
\newcommand{\actions}{\ensuremath{\mathcal{A}}}
\newcommand{\dynamics}{\ensuremath{P}}
\newcommand{\reward}{\ensuremath{r}}

\renewcommand{\sf}{\ensuremath{\bm{\psi}}}
\newcommand{\sr}{\ensuremath{\mu}}
\newcommand{\feature}{\ensuremath{\bm{\theta}}}

\newcommand{\abstractdynamics}{\ensuremath{\bar{P}}}
\newcommand{\abstractreward}{\ensuremath{\bar{r}}}

\newcommand{\abstractsf}{\ensuremath{\bar{\sf}}}
\newcommand{\abstractstates}{\ensuremath{\bar{\mathcal{S}}}}
\newcommand{\abstractoptions}{\ensuremath{\bar{\mathcal{O}}}}
\newcommand{\abstractaction}[1]{\ensuremath{{\bar{#1}}}}

\newcommand{\abstractstate}[1]{\ensuremath{{\bar{#1}} }}
\newcommand{\abstractoption}[1]{\ensuremath{{\bar{#1}}}}
\newcommand{\options}{\ensuremath{\mathcal{O}}}
\newcommand{\abstractactions}{\ensuremath{\bar{\mathcal{A}}}}
\newcommand{\smap}{\ensuremath{f}}
\newcommand{\omap}{\ensuremath{g}}

\newcommand{\sa}{\ensuremath{{\bar{s}}}}
\newcommand{\oa}{\ensuremath{{\bar{o}}}}

\newcommand{\pia}{\ensuremath{{\bar{\pi}}}}

\newtheorem{definition}{Definition}[section]

\usepackage{thmtools} 
\usepackage{thm-restate}

\begin{document}

\maketitle

\begin{abstract}
Abstraction plays an important role in the generalisation of knowledge and skills and is key to sample efficient learning. In this work, we study joint temporal and state abstraction in reinforcement learning, where temporally-extended actions in the form of options induce temporal abstractions, while aggregation of similar states with respect to abstract options induces state abstractions. Many existing abstraction schemes ignore the interplay of state and temporal abstraction. Consequently, the considered option policies often cannot be directly transferred to new environments due to changes in the state space and transition dynamics. To address this issue, we propose a novel abstraction scheme building on successor features. This includes an algorithm for transferring abstract options across different environments and a state abstraction mechanism that allows us to perform efficient planning with the transferred options. 
\blfootnote{The appendix to this paper is available at \url{https://hdg94.github.io/assets/img/abstractions_appendix.pdf}.}
\end{abstract}

\section{Introduction}


Reinforcement learning (RL) has recently shown many remarkable successes~\cite{mnih2015human,silver2017mastering}. 
For efficient planning and learning in complex, long-horizon RL problems, it is often useful to allow RL agents to form different types and levels of abstraction~\cite{sutton1999between,li2006towards}.
On the one hand, temporal abstraction allows the efficient decomposition of complex problems into sub-problems. For example, the options framework~\cite{sutton1999between} enables agents to execute options, i.e., temporally-extended actions (e.g., \emph{pick up the key}), representing a sequence of primitive actions (e.g., \emph{move forward}).
On the other hand, state abstraction~\cite{li2006towards,abel2016near} is a common approach for forming abstract environment representations through aggregating similar states into abstract states allowing for efficient planning. 



In this paper we aim to combine the benefits of temporal and state abstractions for enabling transfer and reuse of options across environments. We build on the insight that environments with different dynamics and state representations
may share a similar abstract representation and
address the following question:
How can we transfer learned options to new environments and reuse them for efficient planning and exploration? 
This is a challenging question because options are typically described by policies that are not transferable across different environments due to different state representations and transition dynamics.
This issue is underlined by the fact that abstract options (e.g., \emph{open a door}) can often correspond to different policies in different environments.
To enable option transfer and reuse, we propose an abstract option representation that can be shared across environments.
We then present algorithms that ground abstract options in new environments.
Finally, we define a state abstraction mechanism that reuses the transferred options for efficient planning.

Concretely, to find a transferable abstract option representation, we propose abstract successor options, which represent options through their successor features (SF)~\cite{dayan1993improving,barreto2016successor}. An SF of an option is the vector of cumulative feature expectations of executing the option and thus can act as an abstract sketch of the goals the option should achieve. By defining shared features among the environments, the abstract successor options can be transferred across environments.
Abstract options then need to be grounded in the unseen environments. One way, therefore, is to find a ground option policy that maximises a reward defined as a weighted sum over the SF~\cite{barreto2019option}. However, careful hand-crafting on the reward weights is needed to produce an intended option's behaviour due to interference between the different feature components. 
Alternatively, we formulate option grounding as a feature-matching problem where the feature expectations of the learned option should match the abstract option. Under this formulation, we present two algorithms (\irlnaive{} and \irlbatch{}) which perform option grounding through inverse reinforcement learning (IRL). Our algorithms allow option transfer and can be used for exploration in environments with unknown dynamics.
Finally, to enable efficient planning and reuse of the abstract options, we propose successor homomorphism, a state abstraction mechanism that produces abstract environment models by incorporating temporal and state abstraction via abstract successor options. 
We demonstrate that the abstract models enable efficient planning and yield near-optimal performance in unseen environments.

\section{Background}
\label{sec:background}

\paragraph{Markov Decision Processes (MDP).}
We model an agent's decision making process as an MDP $\mdp = \langle \states, \actions, \dynamics, \reward, \gamma \rangle$, where $\states$ is a set of states the environment the agent is interacting with can be in, $\actions$ is a set of actions the agent can take, $\dynamics\colon \states \times \actions \rightarrow [0,1]^{|\states|}$ is a transition kernel describing transitions between states of the MDP upon taking actions, $\reward\colon \states \times \actions \rightarrow \mathbb{R}$ is a reward function, and $\gamma$ is a discount factor.
An agent's behavior can be characterized by a policy $\pi\colon \states \rightarrow [0,1]^{|\actions|}$, i.e., $\pi(a | s)$ is the probability of taking action $a$ in state $s$.\footnote{We only consider stationary policies in this paper.}

\paragraph{Successor Features (SF).}~\cite{barreto2016successor}
Given features $\feature\colon \states \times \actions \rightarrow \mathbb{R}^d$ associated with each state-action pair. The SF $\sf^\pi_{s_0}$ is the expected discounted sum of features of state-action pairs encountered when following policy $\pi$ starting from $s_0$, i.e.,
\begin{equation}
\sf^\pi_{s_0} =  \mathbb{E}_{\mdp,\pi}[\sum_{t=0}^\infty \gamma^t \feature(S_t, A_t) \mid S_0 = s_0].
\end{equation}

\noindent
\paragraph{Options and Semi-Markov Decision Processes (SMDPs).}
Temporally extended actions, i.e., sequences of actions over multiple time steps, are often represented as options $o=\langle I^o, \pi^o, \beta^o\rangle$, where $I^o \subseteq \states$ is a set of states that an option can be initiated at (initiation set), $\pi^o\colon \states \rightarrow [0,1]^{|\actions|}$ is the option policy, and $\beta^o\colon \states \rightarrow [0,1]$ is the termination condition, i.e., the probability that $o$ terminates in state $s \in \states$.
The transition dynamics and rewards induced by an option $o$ are
\begin{align*}
    P_{s, s'}^o &= \sum_{k=1}^\infty P(s, o, s', k)\gamma^k,\\
    r_{s}^o &= \mathbb{E}[r_{t+1} + \ldots + \gamma^k r_{t+k} \mid \mathcal{E}(o,s,t)], 
\end{align*}
where $P(s, o, s', k)$ is the probability of transiting from $s$ to $s'$ in $k$ steps when following option policy $\pi^o$ and terminating in $s'$, $\mathcal{E}(o,s,t)$ is the event that option $o$ is initiated at time $t$ in state $s$, and $t+k$ is the random time at which $o$ terminates.

An SMDP~\cite{sutton1999between} is an MDP with a set of options, i.e., $\mdp=\langle \states, \options, \dynamics, \reward, \gamma \rangle$, where $\dynamics\colon\states\times\options\rightarrow[0,1]^{|S|}$ are the options' transition dynamics, and $\reward\colon\states\times\options\rightarrow \mathbb{R}$ is the options' rewards. 
A family of variable-reward SMDPs~\cite{mehta2008transfer} (denoted as $\sf$-SMDP) is defined as $\mdp=\langle \states, \options, \dynamics, \sf, \gamma \rangle$, where $\sf_s^o$ is the SF of option $o$ starting at state $s$. A $\sf$-SMDP induces an SMDP if the reward is linear in the features, i.e., $r_s^o = w_r^T\sf_s^o$. 

\noindent
\paragraph{Inverse Reinforcement Learning (IRL).}
IRL is an approach to learning from demonstrations~\cite{abbeel2004apprenticeship} with unknown rewards.
A common assumption is that the rewards are linear in some features, i.e.,
$r_{s}^a = w_r^T\feature(s,a)$,
where $w_r \in \mathbb{R}^d$ is a real-valued weight vector specifying the reward of observing the different features.
Based on this assumption,~\cite{abbeel2004apprenticeship} observed that for a learner policy to perform as well as the expert's, it suffices that their feature expectations match. Therefore, IRL has been widely posed as a feature-matching problem~\cite{abbeel2004apprenticeship,syed2008apprenticeship,ho2016generative}, where the learner tries to match the expert's feature expectation.

\section{Options as Successor Features}


In this section, we first present abstract successor options, an abstract representation of options through successor features that can be shared across multiple MDPs $\{\mdp_i\}_{i=1}^k$.
Therefore, we assume a \textit{feature function} which maps state-action pairs for each MDP to a shared feature space, i.e., $\feature_{\mdp_i}\colon \states_i \times \actions_i \rightarrow \mathbb{R}^d$.
Such features are commonly adopted by prior works~\cite{barreto2016successor,syed2008apprenticeship,abbeel2004apprenticeship}, and can be often obtained through feature extractors such as an object detector~\cite{girshick2015fast}.
We then present two algorithms that enable us to transfer abstract successor options to new environments with and without known dynamics by grounding the abstract options.


\subsection{Abstract Successor Options}\label{subsec:abstract_options}
\emph{Abstract successor options} are vectors in $\mathbb{R}^d$, representing cumulative feature expectations that the corresponding ground options should realise.

\begin{definition}[Abstract Successor Options, Ground Options] Let $\mdp=\langle \states, \actions, \dynamics, \reward, \gamma \rangle$ be an MDP and $\feature_\mdp\colon\states\times\actions\rightarrow\mathbb{R}^d$ be the feature function. An \emph{abstract successor option} is a vector $\sf^{\abstractoption{o}} \in \mathbb{R}^d$. 
For brevity, we will often denote $\sf^{\abstractoption{o}} \in \mathbb{R}^d$ as $\abstractoption{o}$.
Let $g_s\colon \options\rightarrow \mathbb{R}^d$ denote a state-dependent mapping such that $o \mapsto \sf_s^o = \mathbb{E}[\sum_{\kappa=0}^k \gamma^\kappa\feature_\mdp(S_{t+\kappa}, A_{t+\kappa})| \mathcal{E}(o,s,t)]$.
The options $o \in g_s^{-1}(\abstractoption{o})$ are referred to as \emph{ground options} of $\abstractoption{o}$ in state $s$.
\end{definition}

The initiation set of an abstract successor option $\abstractoption{o}$ can now be naturally defined as $I^{\abstractoption{o}} \coloneqq \{s\in \states \mid \exists o\colon g_s(o) = \sf^{\abstractoption{o}}\}$. In the following section, we present an algorithm for grounding abstract successor options in an MDP $\mdp$ using IRL.

\subsection{Grounding Abstract Successor Options} \label{subsec:lifting_options}
The challenge of grounding abstract successor options, i.e., finding ground policies which realize an abstract successor option's feature vector, corresponds to a feature-matching problem.
This problem, although with a different aim, has been extensively studied in IRL, where a learner aims to match the expected discounted cumulative feature expectations of an expert~\cite{syed2008apprenticeship,abbeel2004apprenticeship}. Inspired by IRL, we present two algorithms for option grounding: \textsc{IRL-naive} and \textsc{IRL-batch}, where the latter algorithm enjoys better runtime performance while achieving similar action grounding performance (cf.\ experiments).

\begin{algorithm}[t]
  \footnotesize
\captionof{algorithm}{Option Grounding (\textsc{IRL-naive})}
\label{algo:lifting_original}
\begin{algorithmic}[1]
\Statex{\textbf{Input:} $\!\mdp\!\!=\!\!\langle \states, \actions, \dynamics, \reward, \gamma \rangle$, abstract option $\abstractoption{o}$, $\epsilon_\textnormal{thresh.}$ }
\Statex{\textbf{Output:} initiation set $I^\abstractoption{o}$, dict.\ of ground option policies $\Pi^\abstractoption{o}$, dict.\ of termination probabilities $\Xi^\abstractoption{o}$}
\State{\textit{// Construct augmented MDP}}
\State{$\states' \leftarrow \states \cup \{s_\textnormal{null}\}, \actions' \leftarrow \actions\cup \{a_\textnormal{T}\}, P' \leftarrow P $}
\State{$\forall s\in \states'\colon \dynamics'(s_\textnormal{null}| s, a_\textnormal{T}) = 1$} 
\State{$\forall a \in \actions'\colon P'(s_\textnormal{null} \mid s_\textnormal{null}, a) = 1$}
\State{\textit{// Find ground options (for \textnormal{IRL} see Algorithm~\ref{algo:lp_original}, Appendix~\ref{subsec:algorithm})}}
\ForAll {$s_\textnormal{start} \in \states$,}
    \State{$\epsilon, \pi^o_{s_\textnormal{start}} \leftarrow$ IRL($\states', \actions', \dynamics', \gamma, s_\textnormal{start}, \sf^\abstractoption{o}$)  }
    \If{$\epsilon \leq \epsilon_\textnormal{thresh.}$}
        \State{$I^\abstractoption{o} \leftarrow I^\abstractoption{o} \cup \{s_\textnormal{start}\}$}
        \State{$\Pi^\abstractoption{o}(s_\textnormal{start}) = \pi^o_{s_\textnormal{start}}$}
        \State{$\Xi^\abstractoption{o}(s_\textnormal{start}) = \beta^o$, where $\beta^o(s) = \pi^o_{s_\textnormal{start}}(a_\textnormal{T}|s)$}
    \EndIf
\EndFor
\State{\textbf{return} $I^\abstractoption{o}, \Pi^\abstractoption{o}, \Xi^\abstractoption{o}$}
\end{algorithmic}
\end{algorithm}

\paragraph{\textsc{IRL-naive} (Algorithm~\ref{algo:lifting_original})} is a naive algorithm for grounding an abstract option $\abstractoption{o}$.
The algorithm uses feature matching based IRL to find for each possible starting state $s_\textnormal{start}$ an option realizing the feature vector $\sf^{\abstractoption{o}}$.
First, we create an augmented MDP which enables termination of options. To do this, we augment the action space with a terminate action $a_\textnormal{T}$ and append a null state $s_\textnormal{null}$ to the state space. Specifically, $a_\textnormal{T}$ will take the agent from any ``regular'' state to the null state. Taking any action in the null state will lead to itself, and yield a zero feature, i.e., $\feature(s_\textnormal{null}, \cdot) = \mathbf{0}$. 
With the augmented MDP, we can compute an option policy via IRL. There exists several IRL formulations, and we adopt the linear programming (LP) approach by~\cite{syed2008apprenticeship} (Algorithm~\ref{algo:lp_original} in Appendix~\ref{subsec:algorithm}).
Specifically, the algorithm finds the discounted visitation frequencies for all state-action pairs that together match the abstract successor option, 
then the option policy $\pi^o$ (including option termination) can be deduced from the state-action visitation frequencies.
Finally, if the discrepancy $\epsilon$ of the successor feature of the learned option and the abstract option is below a specified threshold $\epsilon_\textnormal{thresh.}$, the start state $s_\textnormal{start}$ will be added to the initiation set. 

\paragraph{\textsc{IRL-batch} (Algorithm~\ref{algo:lifting_batch}, Appendix~\ref{subsec:algorithm}).}\textsc{IRL-naive} needs to solve an IRL problem for each state in $\states$ which is computationally demanding.
To improve the efficiency of abstract option grounding, we propose \textsc{IRL-batch}, a learning algorithm that performs IRL for starting states in batches. 
The main challenge 
is that the state-action visitation frequencies found by the LP for matching the abstract option's feature vector may be a mixture of different option policies from different starting states.
To approach this issue, \textsc{IRL-batch} uses a recursive procedure with a batched IRL component (cf.\ Algorithm~\ref{algo:lp_iterative} in Appendix~\ref{subsec:algorithm}) which effectively regularises the learned option policy.
This algorithm can significantly reduce the number of IRL problems that need to be solved while preserving near-optimal performance, cf. Table~\ref{table:key_door_option_fitting}.

\begin{figure*}[t]
	\centering
	\hfill
	\subfloat[\scriptsize Object-Rooms]{\includegraphics[width=0.2\linewidth, height=0.2\linewidth]{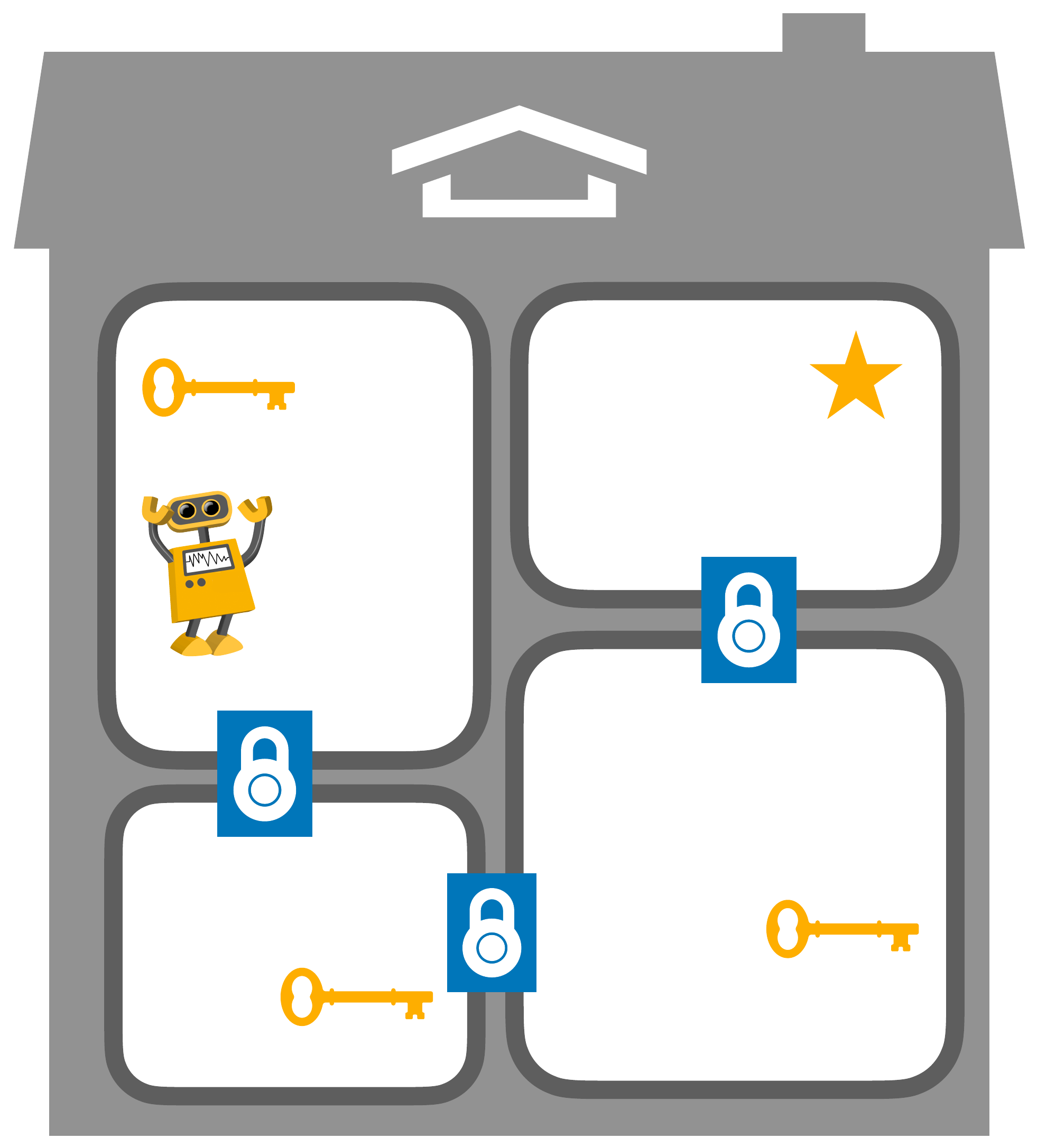}}%
	\hfill
	\subfloat[\scriptsize Abstract $\sf$-SMDP]{\includegraphics[width=0.25\linewidth,height=0.18\linewidth]{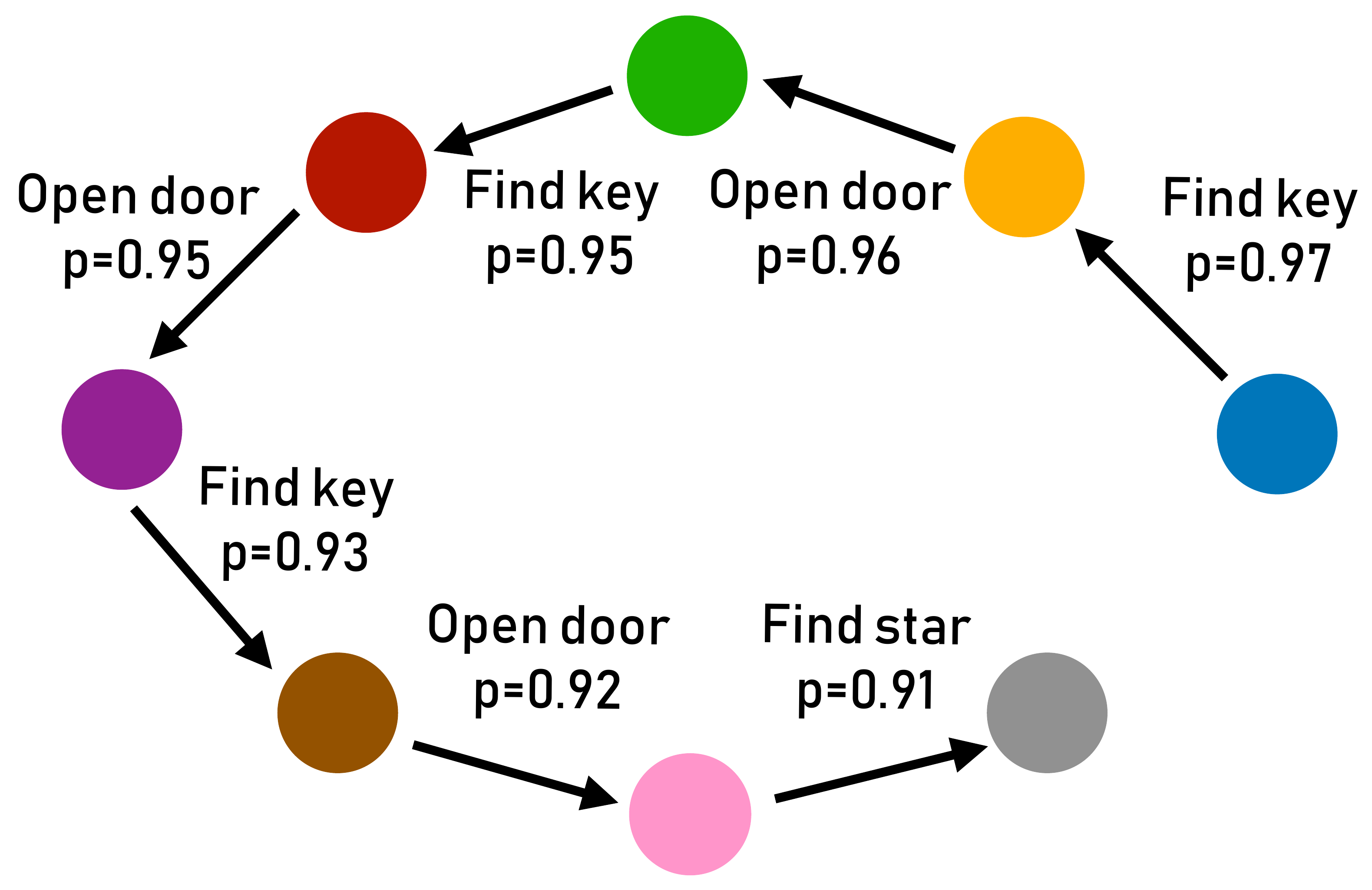}}%
	\hfill
	\subfloat[\scriptsize Successor Homomorphism]{\includegraphics[width=0.25\linewidth, height=0.2\linewidth]{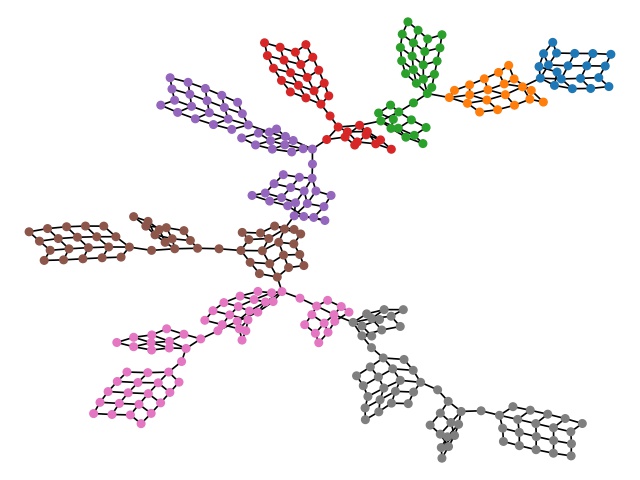}}%
	\hfill
	\subfloat[\scriptsize $Q$ (all)-Abstraction]{\includegraphics[width=0.25\linewidth, height=0.2\linewidth]{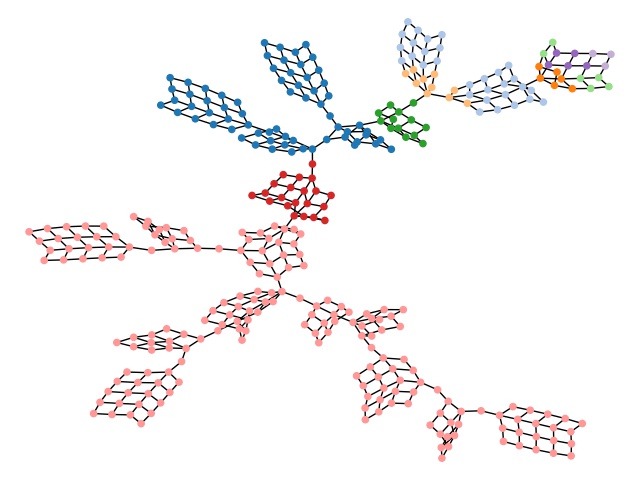}}
	\hfill
	\caption{MDP abstraction in the Object-Rooms domain. (b) Abstract $\sf$-SMDP induced by our successor homomorphism from the ground MDP as shown in (c), the abstract states in (b) correspond to aggregated ground states of the same colour in (c). (d) Abstraction induced by approximate $Q^*$-irrelevance abstraction (cf.\ Sec.~\ref{subseq:other_abstractions}) for \emph{find key}; the abstraction does not carry temporal semantics, and is not reusable for other tasks e.g., \emph{find star.}. Another example can be found in Figure~\ref{fig:abstract_mdp_graphs_larger} in Appendix~\ref{app:settings} and more details are in the experiments section.}
	\label{fig:abstract_mdp_graphs_smaller}
	\vspace{-2mm}
\end{figure*}

\subsection{Transfer Via Abstract Successor Options}
Since features are assumed to be shared across MDPs, an existing option from a source environment can be transferred to a target environment through first finding the abstract successor option $\abstractoption{o} = g_s(o)$ by computing the ground option's successor feature, then grounding $\abstractoption{o}$ in the target environment $o' \in g_s^{-1}(\abstractoption{o})$ using the \textsc{IRL-naive} or \textsc{IRL-batch} algorithm. 
When the transition dynamics of the target environment are unknown, exploration is needed before option grounding to construct the (approximate) MDP transition graph, e.g., through a random walk.
However, random walks can often get trapped in local communities~\cite{pons2005computing} and thus can be inefficient for exploration.
On the other hand, unlike solving a long-horizon task with sparse rewards, options are often relatively localised, and hence can be readily found with partially constructed transition graphs.
This enables simultaneous exploration and option grounding:
given a set of abstract options $\abstractoptions$,
we start with an iteration of a random walk which constructs an approximate MDP $\Tilde{\mdp}_0$. In each subsequent iteration $k$, we ground the abstract options using $\Tilde{\mdp}_{k-1}$, and update the MDP $\Tilde{\mdp}_k$ by exploring with a random walk using both primitive actions and the computed ground options. We show empirically that, using our approach, ground options can be learned quickly and significantly improve the exploration efficiency, cf.~Section~\ref{sec:grounding-unknown}. 

\section{Abstraction with Successor Homomorphism}
\label{sec:abstraction}

Parallel to temporal abstraction, state abstraction~\cite{dean1997model,li2006towards,abel2016near} aims to form abstract MDPs in order to reduce the complexity of planning while ensuring close-to-optimal performance compared with using the original MDP. Typically, an abstract MDP is formed by aggregating similar states into an abstract state.  
However, most prior methods do not take into account temporal abstraction thus often requiring strict symmetry relations among the aggregated states.
To overcome this limitation, we propose a state abstraction mechanism based on our abstract successor options and $\psi$-SMDPs (cf.\ Sec.~\ref{sec:background}), which naturally models the decision process with the abstract options.
In particular, we propose \emph{successor homomorphisms}, which define criteria for state aggregation and induced \emph{abstract $\psi$-SMDPs}. 
We show that planning with these abstract $\psi$-SMDPs yields near-optimal performance. Moreover, the abstract $\psi$-SMDPs can inherit meaningful semantics from the options, c.f.,  Fig.~\ref{fig:abstract_mdp_graphs_smaller}.

Specifically, an abstract $\psi$-SMDP is a tuple $\abstractmdp = \langle \abstractstates, \abstractoptions, \abstractdynamics, \abstractsf, \gamma \rangle$, where $\abstractstates$ is a set of abstract states, $\abstractoptions$ are the abstract options, whose transition dynamics between the abstract states are described by $\abstractdynamics_{\abstractstate{s}, \abstractstate{s}'}^{\abstractoption{o}}$, and $\abstractsf_{\abstractstate{s}}^\abstractoption{o}$ is the SF of executing $\bar{o}$ from $\bar{s}$. A successor homomorphism $h = (f(s), g_s(o), w_s)$ maps a (ground) $\psi$-SMDP to an abstract $\psi$-SMDP,  where $f(s)$ aggregates ground states with approximately equal option transition dynamics into abstract states, $g_s(o)$ is our state-dependent mapping between ground and abstract options, and $w_s$ is a weight function which weighs the aggregated ground states towards each abstract state. 

\begin{definition}[$\epsilon$-Approximate Successor Homomorphism]\label{def:epsilon_successor_homomorphism}
Let $h=(\smap(s), \omap_s(o), w_s)$ be a mapping from a ground $\sf$-SMDP $\mdp = \langle \states, \options, \dynamics, \sf, \gamma \rangle$ to an abstract $\sf$-SMDP $\abstractmdp = \langle \abstractstates, \abstractoptions, \abstractdynamics, \abstractsf, \gamma \rangle$, where $\smap\colon \states \rightarrow \abstractstates$, $\omap_s\colon \options \rightarrow \abstractoptions$ and $w\colon \states \rightarrow [0,1]$ s.t. $\forall \abstractstate{s}\in \abstractstates, \sum_{s\in \smap^{-1}(\abstractstate{s})}w_s = 1$. $h$ is an \emph{$(\epsilon_P, \epsilon_\psi)$-approximate successor homomorphism} if for $\epsilon_P, \epsilon_\psi>0$, $\forall s_1,s_2\in \states, o_1, o_2\in \options$,
\begin{align*}
&h(s_1, o_1) = h(s_2, o_2)\footnotemark \implies\!\!\! \\
&\quad \forall s'\in \states \; | \sum_{s''\in \smap^{-1}(\smap(s'))} \dynamics^{o_1}_{s_1, s''}\!\! - \!\dynamics^{o_2}_{s_2, s''}|\leq \epsilon_P,\\
&\quad \textnormal{and }
    \|\sf_{s_1}^{o_1} - \sf_{s_2}^{o_2}\|_1 \leq \epsilon_\psi.
\end{align*}
\footnotetext{$h(s_1, o_1) \!=\! h(s_2, o_2) \!\Leftrightarrow\! f(s_1)\!=\!f(s_2) \textnormal{ and } g_{s_1}(o_1)\!=\! g_{s_2}(o_2)$}
\end{definition}
Having defined the abstract states, the transition dynamics and features of the abstract $\psi$-SMDP $\abstractmdp$ can be computed by:
$\abstractdynamics_{\abstractstate{s}, \abstractstate{s}'}^{\abstractoption{o}} = \!\!\!\!\!\!\!\sum\limits_{s\in \smap^{-1}(\abstractstate{s})}\!\!\!\!\!w_s\!\!\sum\limits_{s' \in \smap^{-1}{(\abstractstate{s}')}}\!\!\!\!\!\dynamics_{s, s'}^{g^{-1}_s(\abstractoption{o})}
\textnormal{ , and }
\abstractsf_{\abstractstate{s}}^{\abstractoption{o}} = \!\!\!\!\!\sum\limits_{s\in \smap^{-1}(\abstractstate{s})} \!\!\!\!\!w_s \sf_{s}^{g^{-1}_s(\abstractoption{o})},
$

\noindent
where $\sum_{s''\in \smap^{-1}(\smap(s'))}\dynamics^o_{s, s''}$ refers to the transition probability from ground state $s$ to an abstract state $\abstractstate{s}' =f(s')$ following option $o$. Intuitively, two states mapping to the same abstract state have approximately equal option transition dynamics towards all abstract states, and the corresponding ground options induce similar successor features.
For efficient computation of the abstract model, the transition dynamics condition can be alternatively defined on the ground states s.t.\ $h(s_1, o_1) = h(s_2, o_2) \implies \forall s'\in S$,
\begin{align}\label{eq:ground}
 \vspace{-1mm}
 |\dynamics^{o_1}_{s_1, s'} - \dynamics^{o_2}_{s_2, s'}| \leq \epsilon_P, \textnormal{ and }\quad
 \|\sf_{s_1}^{o_1}  - \sf_{s_2}^{o_2}\|_1 \leq \epsilon_\psi,
 \vspace{-1mm}
\end{align}
which states that two states mapping to the same abstract state have approximately equal option transition dynamics towards all \emph{ground} states, and the corresponding ground options induce similar SF. In general, Def~\ref{def:epsilon_successor_homomorphism} and Eq.~\eqref{eq:ground} can result in different abstractions, but similar performance bounds could be derived. 
In cases in which multiple ground options map to the same abstract option, $g^{-1}_s(\abstractoption{o})$ picks one of the ground options. An example abstract $\sf$-SMDP induced by our approximate successor homomorphism is shown in Fig.~\ref{fig:abstract_mdp_graphs_smaller}.


Our successor homomorphism combines state and temporal abstraction, by aggregating states with similar multi-step transition dynamics and feature expectations, in contrast to one-step transition dynamics and rewards. This formulation not only relaxes the strong symmetry requirements among the aggregated states but also provides the induced abstract $\psi$-SMDP with semantic meanings obtained from the options.
Furthermore, each abstract $\sf$-SMDP instantiates a family of abstract SMDPs (cf. Def.~\ref{def:abstract_smdp}, Appendix~\ref{subsec:mdp_abstraction}) by specifying a task, i.e., reward weight $w_r$ on the features. Extending results from~\cite{abel2016near,abel2020value} to our setting, we can guarantee performance of planning with the abstract $\psi$-SMDP across different tasks.

\begin{restatable}{theorem}{thm}\label{thm} Let $w_r\colon \mathbb{R}^d \rightarrow \mathbb{R}$ be a linear reward vector such that $r_s^a=w_r^T \feature(s,a)$. Under this reward function, the value of an optimal abstract policy obtained through the $(\epsilon_P, \epsilon_\psi)$-approximate successor homomorphism is close to the optimal ground SMDP policy, where the difference is bounded by $\frac{2\kappa}{(1-\gamma)^2}$,  where $\kappa = |w_r|(2\epsilon_\psi + \frac{\epsilon_P|\abstractstates|\max_{s,a}|\feature(s,a)|}{1-\gamma})$. (c.f. appendix for the proof)
\end{restatable}

\section{Experiments}\label{sec:experiments}
\begin{figure*}[t]
    \vspace{-2mm}
	\centering
	\subfloat[State visitation frequencies]{\includegraphics[width=0.33\linewidth, height=0.20\linewidth]{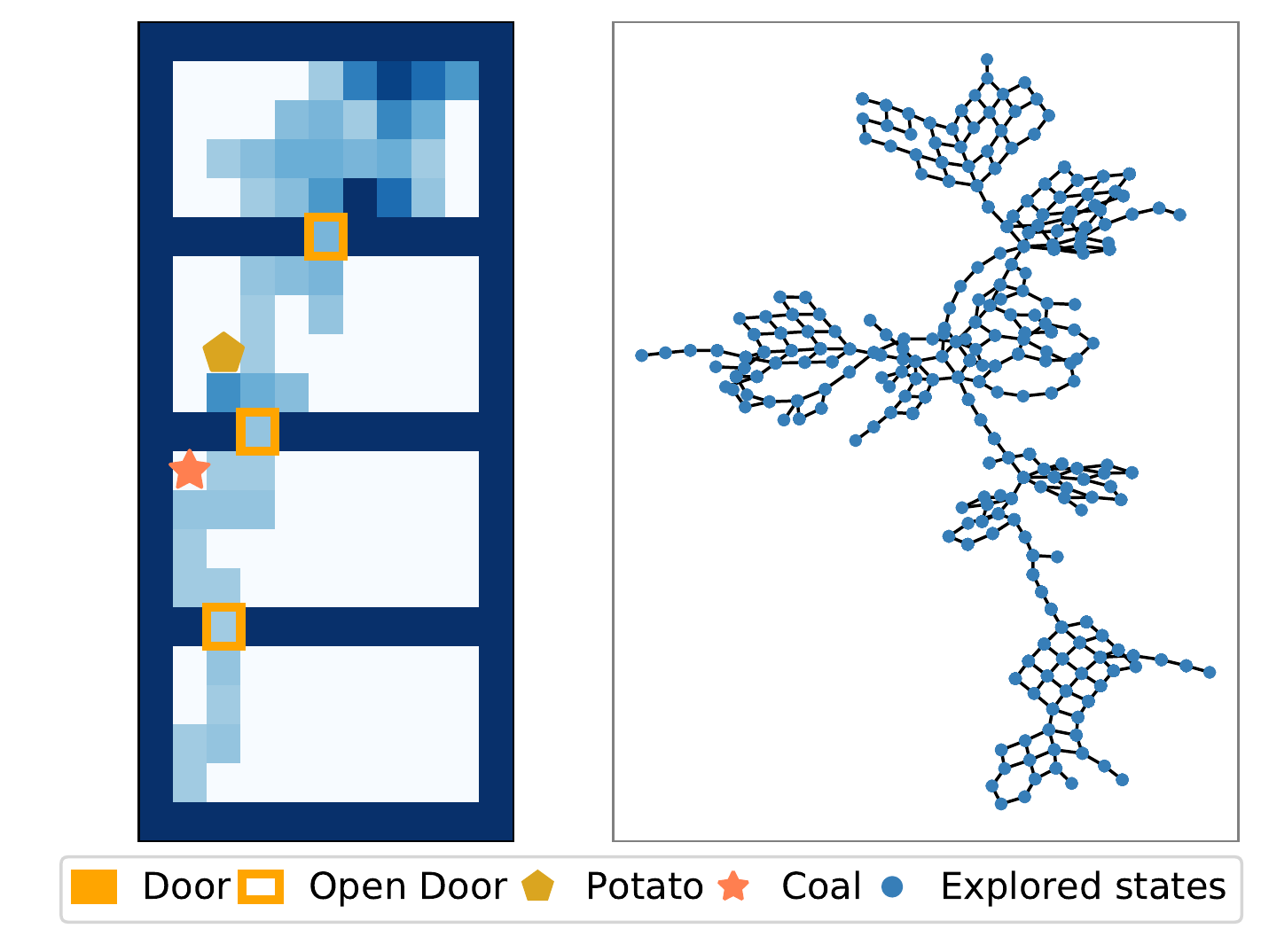}}%
	\hfill
	\subfloat[Setting]{\includegraphics[width=0.28\linewidth, height=0.2\linewidth]{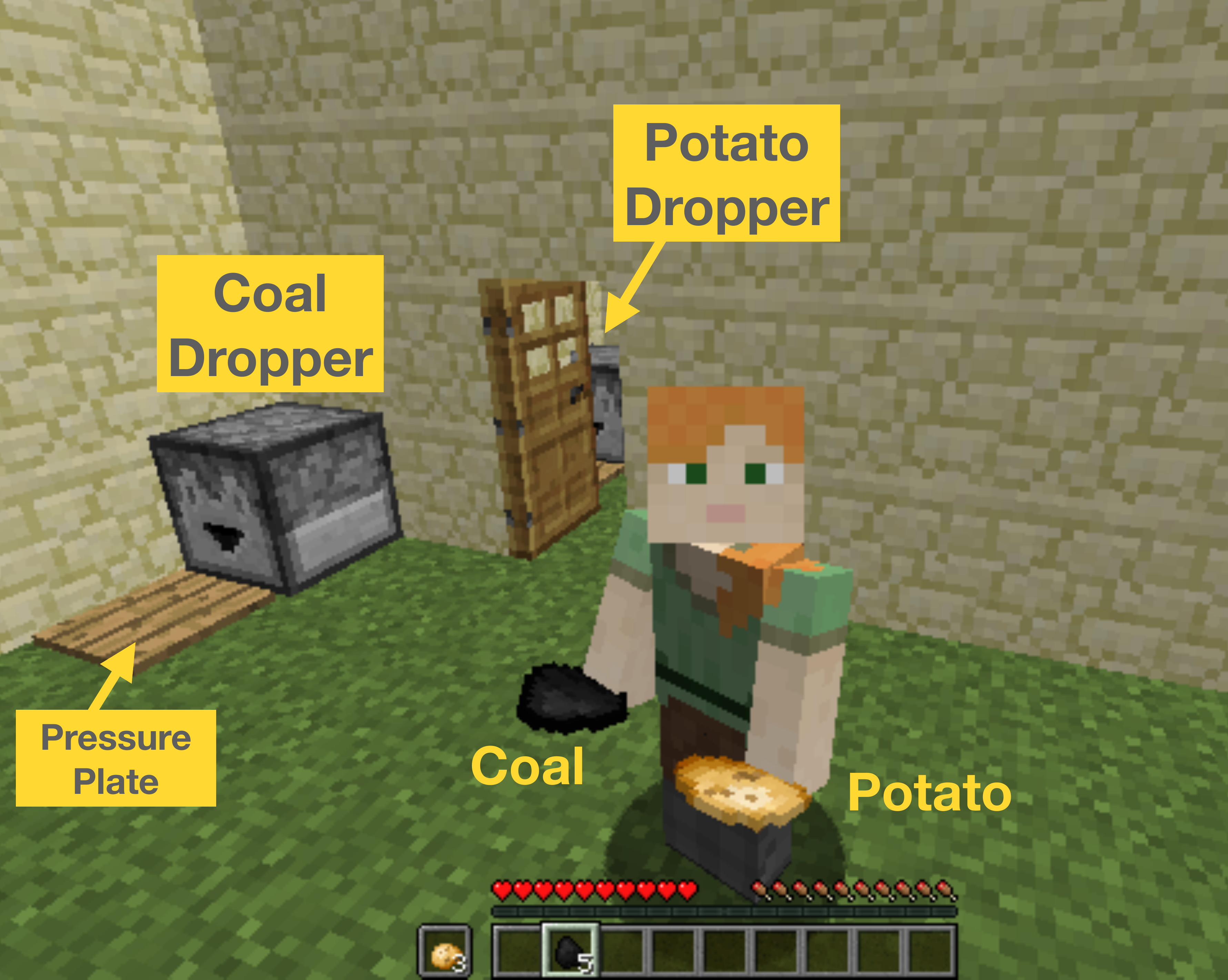}\label{fig:minecraft_baking_scene}}%
	\hfill
	\subfloat[States explored]{\includegraphics[width=0.25\linewidth, height=0.2\linewidth]{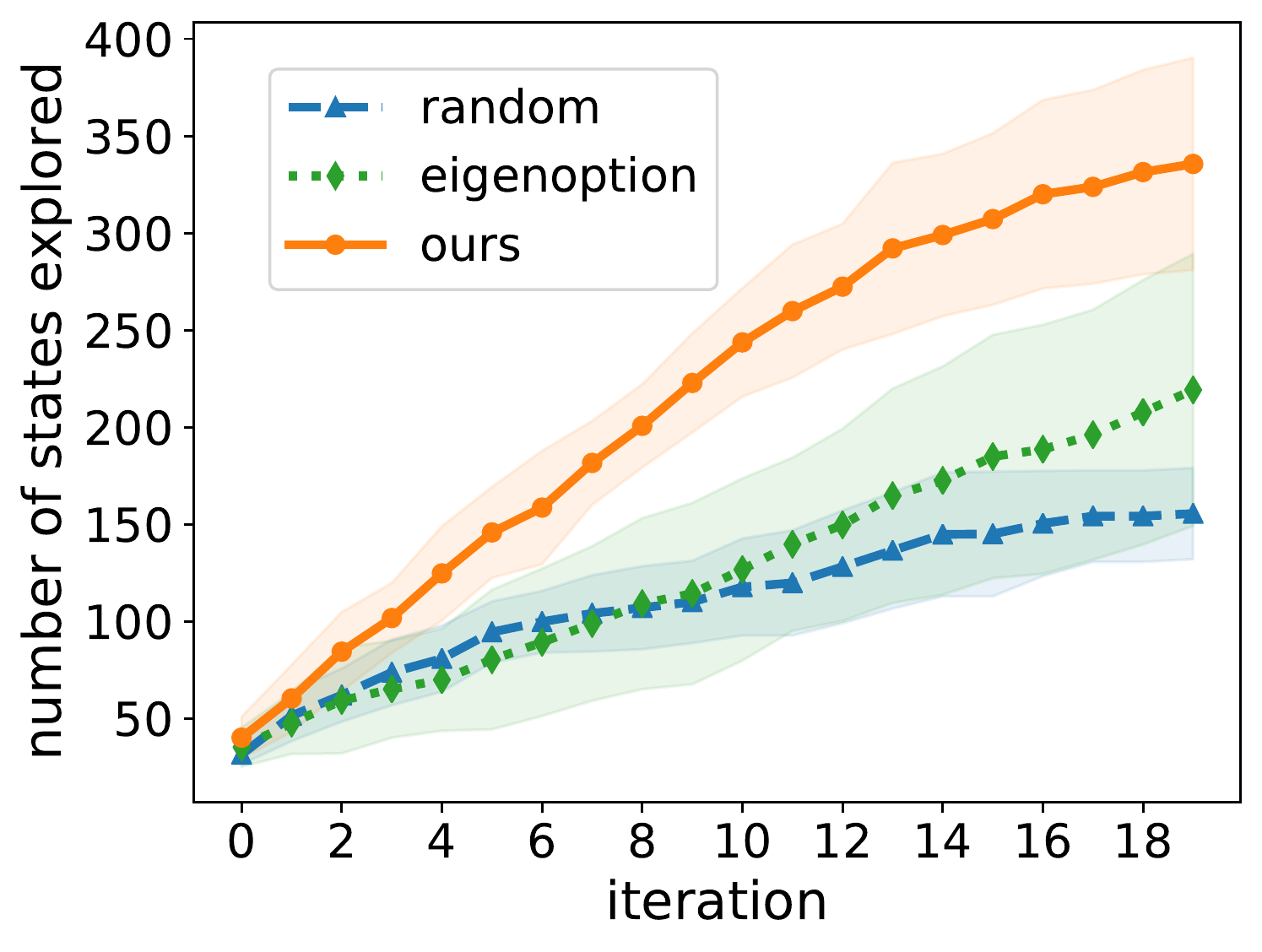}}%
	\vfill
	\vspace{-2mm}
	\subfloat[Baked potato obtained]{\includegraphics[width=0.23\linewidth, height=0.18\linewidth]{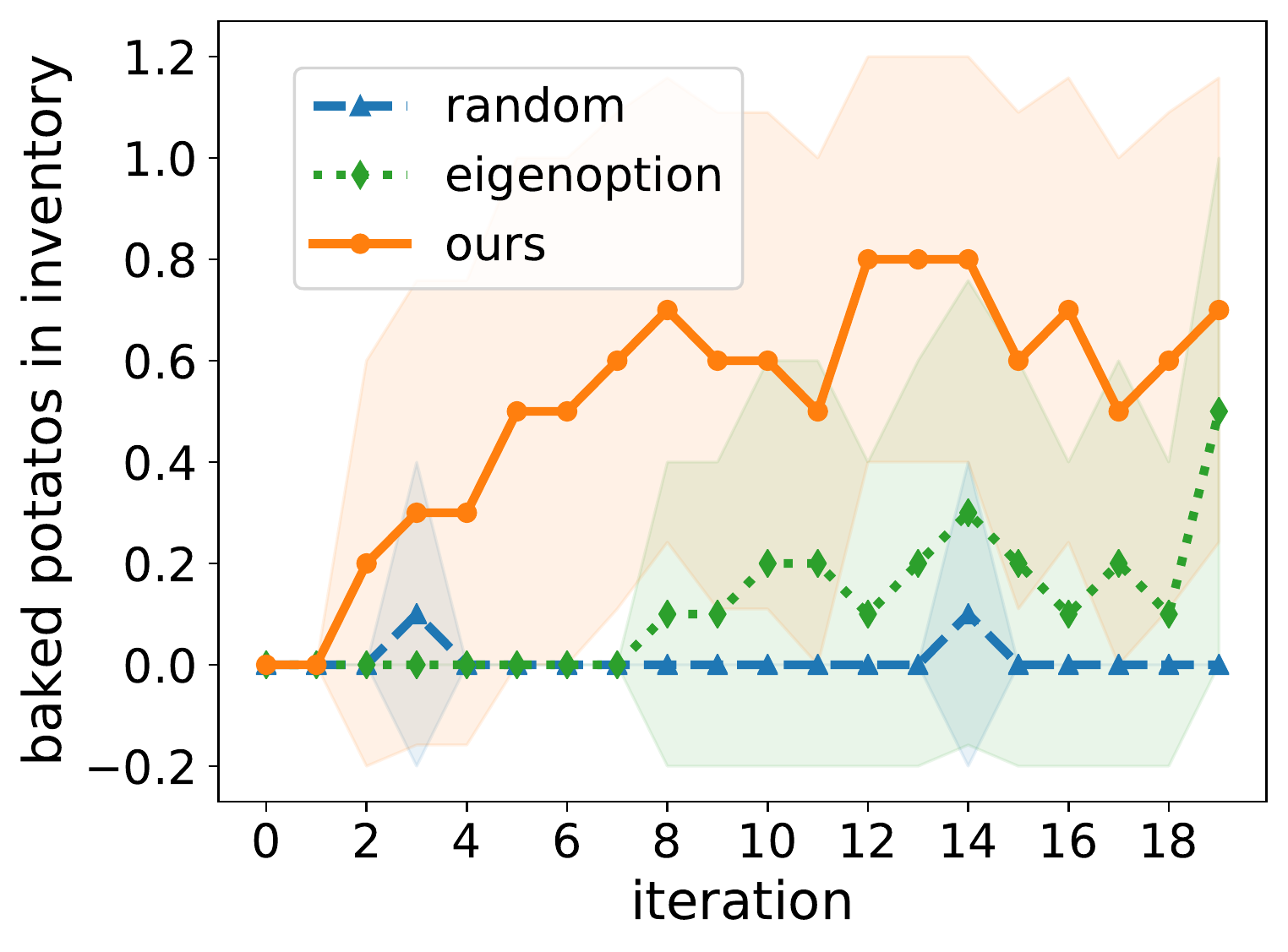}}%
	\hfill
	\subfloat[Coal obtained]{\includegraphics[width=0.23\linewidth, height=0.18\linewidth]{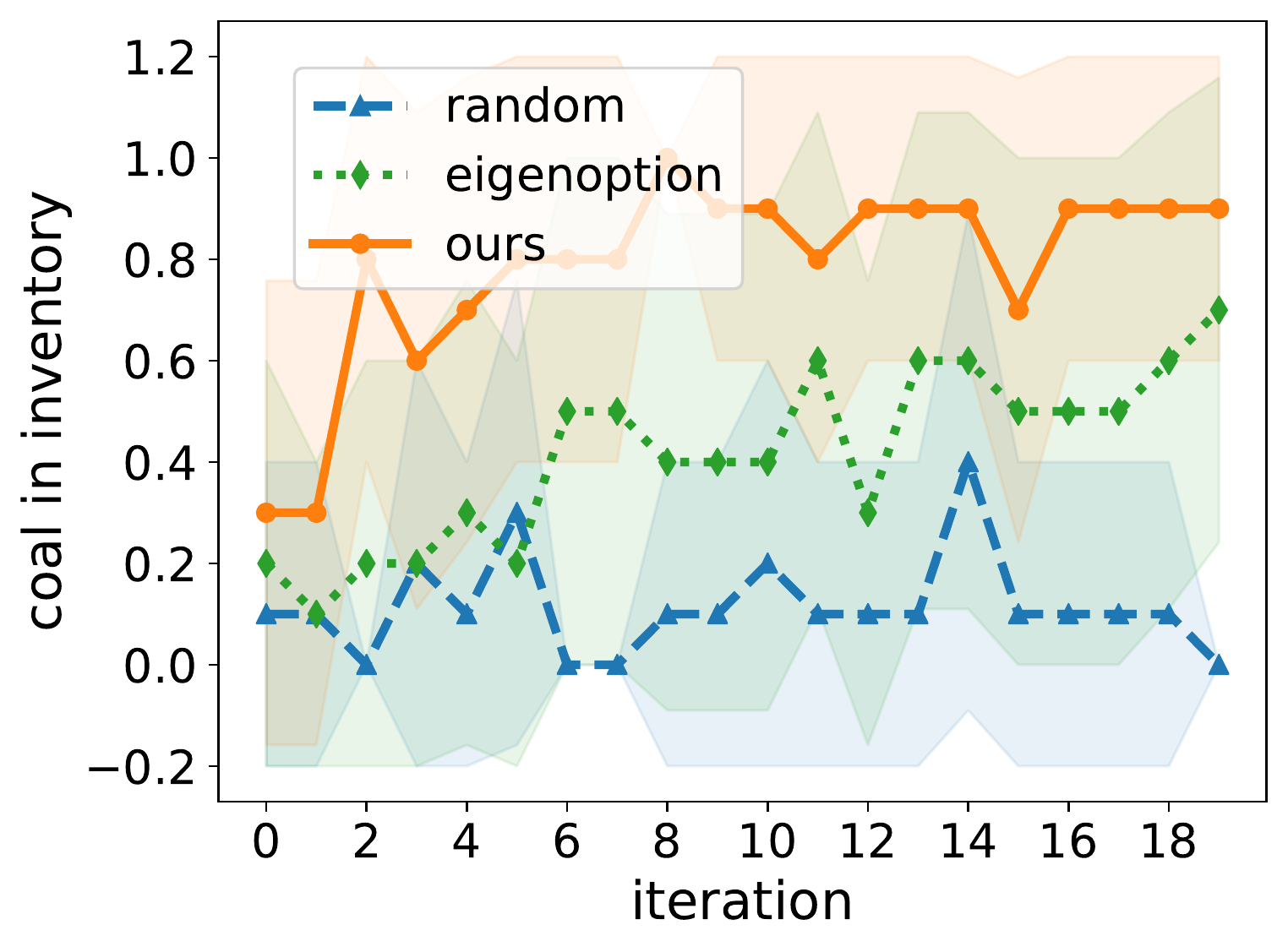}}%
	\hfill
	\subfloat[Potato obtained]{\includegraphics[width=0.23\linewidth, height=0.18\linewidth]{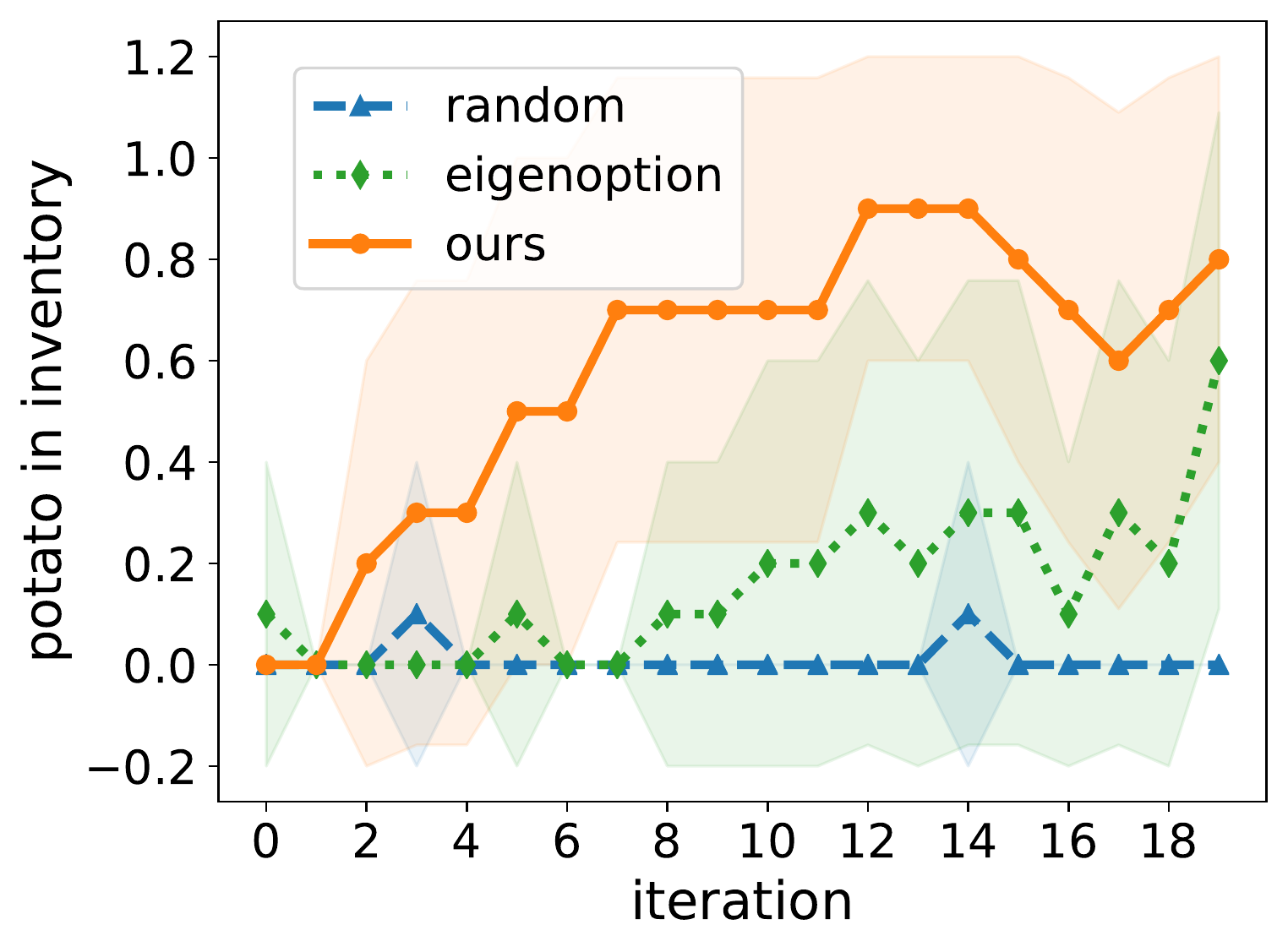}}%
	\hfill
	\subfloat[Doors opened]{\includegraphics[width=0.23\linewidth, height=0.18\linewidth]{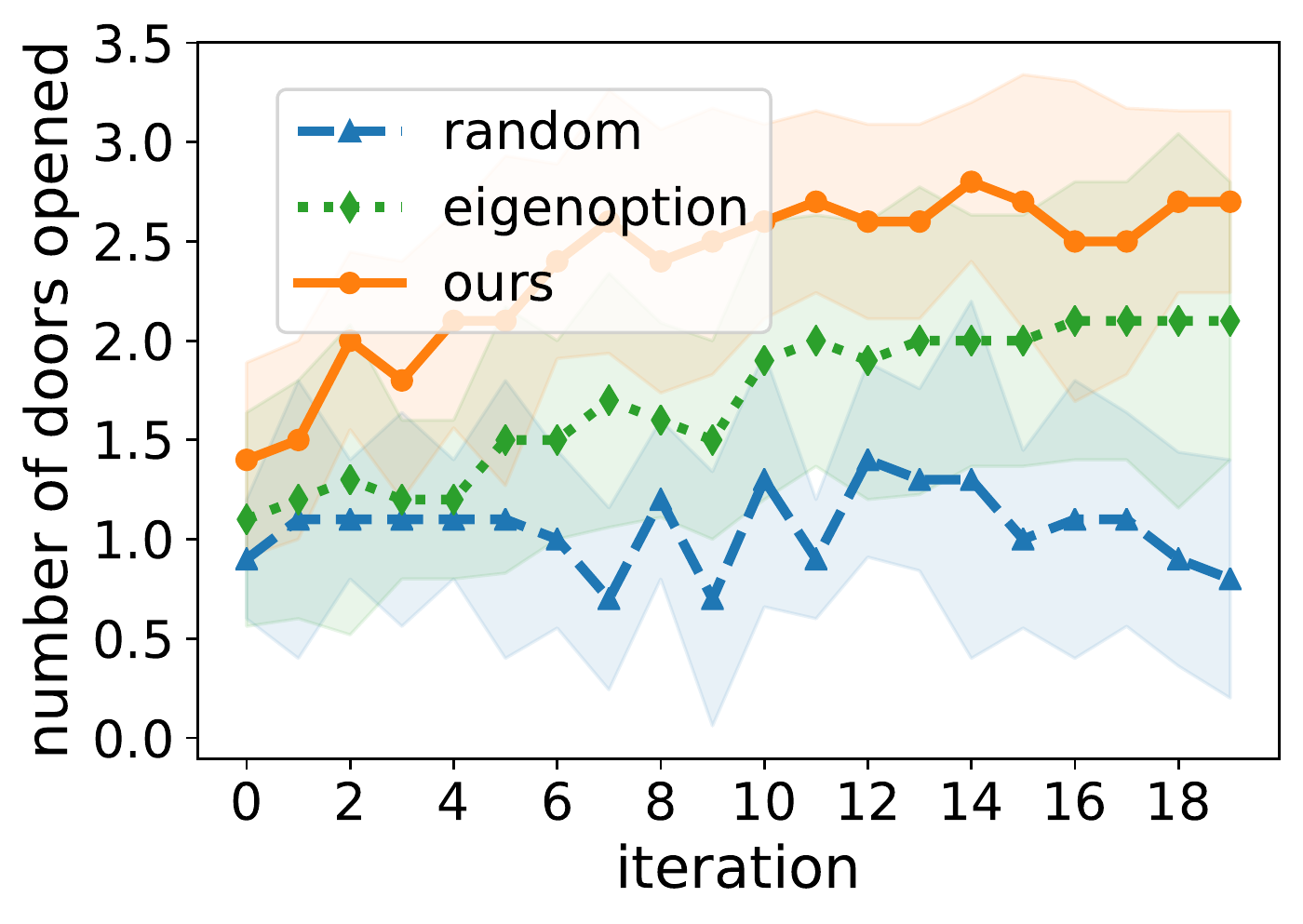}}%
	\hfill
	\caption{Exploring and grounding options in unknown environments in the Minecraft setting.} 
	\label{fig:minecraft_potato}
\end{figure*}

In this section, we empirically evaluate our proposed abstraction scheme for option transfer to new environments with known and unknown transition dynamics, and planning with abstract SMDPs. 
Additional details for all experiments are available in Appendix~\ref{subsec:experiment_details}.

\subsection{Option Transfer}
We evaluate the performance and efficiency of our option grounding algorithm for transferring options given by expert demonstrations to new environments. 

\noindent
\subsubsection{Transfer to environments with known dynamics}

\begin{table}[t]
\centering
\resizebox{1.\linewidth}{!}{%
\renewcommand{\arraystretch}{1.} 
\begin{tabular}{@{}cccccccc@{}}
\toprule
 & & \multicolumn{2}{c}{2 Rooms} & \multicolumn{2}{c}{3 Rooms} & \multicolumn{2}{c}{4 Rooms} \\ \cmidrule(l){3-4} \cmidrule(l){5-6} \cmidrule(l){7-8} 
 Goal & Algorithm & success & \textit{LP} & success & \textit{LP} & success & \textit{LP} \\ \midrule
\multirow{3}{*}{Find Key} & \irlnaive & 1.0 & 157 & 1.0 & 487 & 1.0 & 1463 \\
 & \irlbatch & 1.0 & \textbf{2} & 1.0 & \textbf{2} & 1.0 & \textbf{2} \\
 & \ok & 1.0 & - & 1.0 & - & 1.0 & - \\ \midrule
\multirow{3}{*}{Find Star} & \irlnaive & 1.0 & 157 & 1.0 & 487 & 1.0 & 1463 \\
 & \irlbatch & 1.0 & \textbf{2} & 1.0 & \textbf{2} & 1.0 & \textbf{2} \\
 & \ok & 1.0 & - & 1.0 & - & 1.0 & - \\ \midrule
\multirow{3}{*}{\begin{tabular}[c]{@{}c@{}}Find Key, \\ Open Door\end{tabular}} & \irlnaive & 1.0 & 157 & 1.0 & 487 & \textbf{1.0} & 1463 \\
 & \irlbatch & \textbf{1.0} & \textbf{3} & \textbf{1.0} & \textbf{15} & 0.95 & \textbf{29} \\
 & \ok & 0.08 & - & 0.56 & - & 0.71 & - \\ \midrule
\multirow{3}{*}{\begin{tabular}[c]{@{}c@{}}Find Key, \\ Open Door, \\ Find Star\end{tabular}} & \irlnaive & 1.0 & 157 & 1.0 & 487 & 1.0 & 1463 \\
 & \irlbatch & \textbf{1.0} & \textbf{4} & \textbf{1.0} & \textbf{16} & \textbf{1.0} & \textbf{7} \\
 & \ok & 0.0 & - & 0.51 & - & 0.66 & - \\ \bottomrule
\end{tabular}
}
\caption{Performance and efficiency of option grounding in the Object Rooms. 
We show the success rate (success) of the learned options for achieving all specified goals across all starting states in the initiation set, and the number of LPs used to find the option policy.}
\label{table:key_door_option_fitting}
\end{table}

We compare our option grounding algorithms with the baseline algorithm Option Keyboard (\ok{})~\cite{barreto2019option} on the Object-Rooms Setting.

\emph{\phantom{s}$\sbullet$ \irlnaive{} (Algorithm~\ref{algo:lifting_original}) and \irlbatch{} (Appendix; Algorithm~\ref{algo:lifting_batch})}. Our both algorithms \emph{transfer} an expert's demonstration from a 2-Room source environment to a target environment of 2-4 Rooms, by first encoding the expert's demonstration as an abstract option, then grounding the abstract option in the target environment.\\
\emph{\phantom{s}$\sbullet$ Option Keyboard (\ok{})}~\cite{barreto2019option}:  OK first trains primitive options (find-key, open-door, find-star) to maximise a cumulant (pseudo-reward on the history). The cumulants are hand-crafted in the target environment, where the agent is rewarded when terminating the option after achieving the goal. Then the composite options are synthesized via generalized policy improvement (GPI).

\emph{Object-Rooms Setting.} $N$ rooms are connected by doors with keys and stars inside. There are 6 actions: $\actions=\{$Up, Down, Left, Right, Pick up, Open$\}$. The agent can pick up the keys and stars, and use the keys to open doors. See Figure~\ref{fig:layouts} in Appendix~\ref{subsec:experiment_details} for an illustration.

\emph{Results.}
Table~\ref{table:key_door_option_fitting} shows the success rate of the computed options for achieving the required goals across all starting states in the initiation sets. 
Both \irlnaive{} and \irlbatch{} successfully transfer the options, moreover, \irlbatch{} significantly reduces the number of LPs solved. The \ok{} composed options have a low success rate for achieving the required goals, e.g., for grounding "find key, open door, find star" in the 3-Room setting, the learned options often terminate after achieving 1-2 subgoals.

\begin{figure*}[t]
	\centering
	\includegraphics[width=0.24\linewidth]{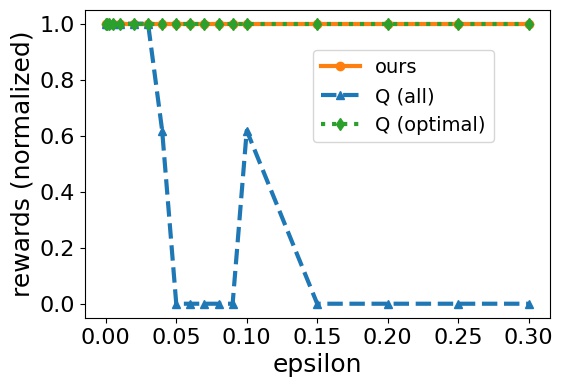}
	\hfill\vspace{-2mm}
	\includegraphics[width=0.24\linewidth]{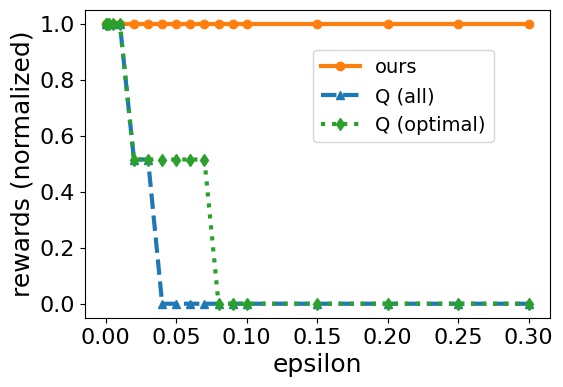}
	\hfill
	\includegraphics[width=0.24\linewidth]{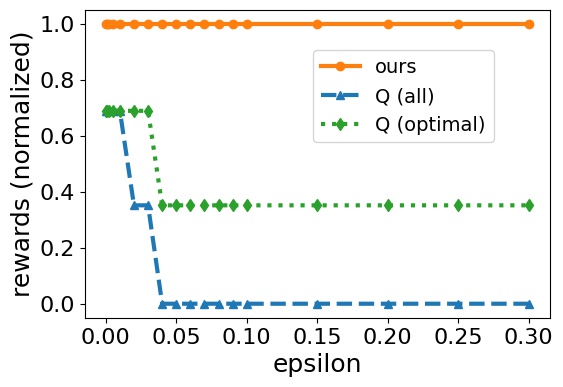}
	\hfill
	\includegraphics[width=0.24\linewidth]{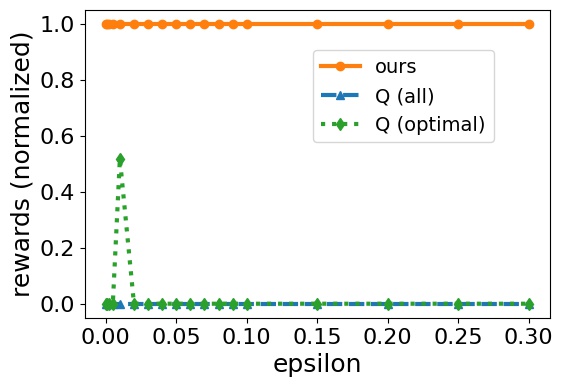}
    \vfill
	\subfloat[dense reward]{\includegraphics[width=0.24\linewidth]{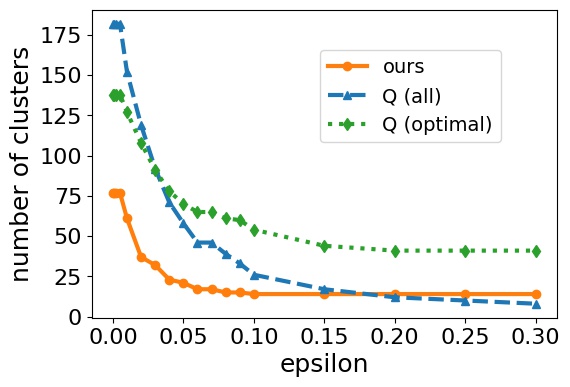}}%
	\hfill
	\subfloat[sparse reward]{\includegraphics[width=0.24\linewidth]{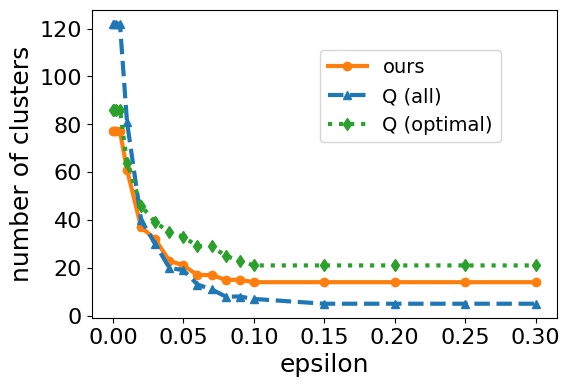}}%
	\hfill
	\subfloat[transfer (w. overlap)]{\includegraphics[width=0.24\linewidth]{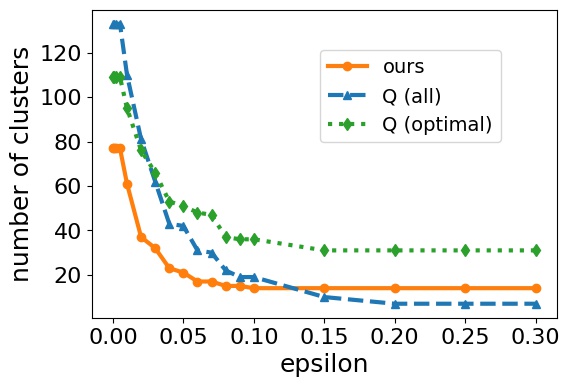}}%
	\hfill
	\subfloat[transfer (w.o. overlap)]{\includegraphics[width=0.24\linewidth]{img/abstraction/abstract_mdp_planning/nclusters_transfer_overlap.jpeg}}%
	\caption{Performance of planning with the abstract MDPs. The upper row shows the total rewards (normalized by the maximum possible total rewards) obtained, and the lower row shows the corresponding number of abstract states of the abstract MDPs. The $x$-axes are the distance thresholds $\epsilon$. \emph{transfer} refers to task transfer (i.e., different reward function). Further details are available in Appendix~\ref{subsec:experiment_details}.}
	\label{fig:abstract_mdp_planning}
	\vspace{-3mm}
\end{figure*}

\noindent
\subsubsection{Transfer to environments with unknown dynamics}
\label{sec:grounding-unknown}
To perform grounding with unknown transition dynamics, the agent simultaneously explores and computes ground options.


\emph{Bake-Rooms Setting (Minecraft)}. The considered environment is based on Malmo, a standard AI research platform using Minecraft~\cite{johnson2016malmo}, as shown in Figure~\ref{fig:minecraft_potato}: 4 rooms (R1-R4 from bottom to top) are connected by doors, and the agent starts in R1. To obtain a baked potato, the agent needs to open doors, collect coal from a coal dropper in R2, collect potato from a dropper in R3, and issue a \emph{craft} command to bake the potato. Agents observe their current location, objects in nearby $3\times3$ grids, and the items in inventory.

\emph{Baselines and learning.}
We compare our \irlbatch{} with the baselines: (i) \emph{random walk} and (ii) \emph{eigenoptions}~\cite{machado2017eigenoption}.
Each agent runs for 20 iterations, each consisting of 200 steps.
In the first iteration, all agents execute random actions. After each iteration, the agents construct an MDP graph based on collected transitions from all prior iterations. The eigenoption agent computes $3$ eigenoptions of smallest eigenvalues using the normalized graph Laplacian, while \irlbatch{} grounds the $3$ abstract options: 1. open and go to door, 2. collect coal, 3. collect potato. In the next iteration, agents perform a random walk with both primitive actions and the acquired options, update the MDP graph, compute new options, and so on.

\emph{Results.}
Figure~\ref{fig:minecraft_potato} (a) shows the state visitation frequency of our algorithm in the $14^\textnormal{th}$ iteration and a constructed transition graph (more details in Figure~\ref{fig:state-visitation-bake-rooms-ours}, Appendix~\ref{subsec:experiment_details}).
The trajectory shows that from R1 (bottom), the agent learns to open door, collect coal in R2, open door, collect potato in R3, and navigate around the rooms. (c) shows that our algorithm explores on average more than 50 percent more states than both baselines in 20 iterations. (d) - (g) shows the objects collected by the agents (max count is 1) and the number of doors opened. The agent using our algorithm learns to open door and collect coal within 2 iterations, and it learns to bake a potato 50 percent of the time within 5 iterations. In all cases, our algorithm outperforms the baselines. The results are averaged over 10 seeds, shaded regions show standard deviations. 

\subsection{Planning with Abstract SMDP}
We compare the performance of planning using our abstract $\psi$-SMDPs found by successor homomorphism, with two baseline abstraction methods~\cite{li2006towards,abel2016near} on the \emph{Object-Rooms} environment. 

\emph{1. Q (all)}: ($Q^*$-irrelevance abstraction), if $f(s_1)= f(s_2)$, then $\forall a, |Q^*(s_1, a) - Q^*(s_2, a)| \leq \epsilon$, where $Q^*(s,a)$ is the optimal Q-function for the considered MDP.

\emph{2. Q (optimal)}: ($a^*$-irrelevant abstraction), if $f(s_1)= f(s_2)$ then $s_1$ and $s_2$ share the same optimal action $a^*$ and $|Q^*(s_1, a^*) - Q^*(s_2, a^*)| \leq \epsilon$.

\emph{Training and results.} 3 abstract options are given: 1. open door, 2. find key, and 3. find star. To find the abstract model induced by our successor homomorphism, we first ground each abstract option. 
Then we form the abstract state by clustering the ground states $s$ according to the pairwise distance of their option termination state distributions $\max_{s', \abstractoption{o}}{|P_{s_1,s'}^{g_{s_1}^{-1}(\abstractoption{o})} - P_{s_2,s'}^{g_{s_2}^{-1}(\abstractoption{o})}|}$  through agglomerative clustering. 
Finally we compute the abstract transition dynamics and feature function.
For Q (all) and Q (optimal), we first perform Q-value iteration on a source task to obtain the optimal Q-values, then cluster the states by their pairwise differences, e.g., $|Q^*(s_1,\cdot) - Q^*(s_2, \cdot)|_\infty$ for the Q (all) approach, then compute the abstract transition dynamics. 


Figure~\ref{fig:abstract_mdp_planning} shows the results of using the induced abstract model for planning.
Our performs well across all tested settings with few abstract states. Since abstract $\psi$-SMDP does not depend on rewards, the abstract model is robust across tasks (with varying reward functions) and sparse rewards settings. Whereas abstraction schemes based on the reward function perform worse when the source task for performing abstraction is different from the target task for planning. 

\section{Related Work}\label{sec:related workd}

Agent-space options~\cite{konidaris2007building} are one of our conceptual inspirations. In our work, we tie the agent-space to the problem-space through features, and our option grounding algorithms allow the agents to transfer agent-space options across different problem spaces.

\textbf{Successor features (SF).} Successor representations (SR) were first introduced by~\cite{dayan1993improving}.
\cite{barreto2016successor} proposed SF which generalised SR. 
\cite{machado2017eigenoption} discovered eigenoptions from SF and showed their equivalence to options derived from the graph Laplacian. \cite{ramesh2019successor} discovered successor options via clustering over the SR.
The Option Keyboard~\cite{barreto2019option} is a pioneering work for option combinations: primitive options are first trained to maximise cumulants (i.e., pseudo-rewards on histories), then by putting preference weights over the cumulants, new options are synthesized to maximise the weighted sum of cumulants. For the purpose of option transfer and grounding,
however, this may yield imprecise option behaviours due to the interference between cumulants. In contrast, our successor feature-matching formulation generates precise option behaviours as demonstrated in Table~\ref{table:key_door_option_fitting}.

\textbf{MDP Abstraction.} MDP abstraction through bisimulation was first considered by~\cite{dean1997model}.~\cite{ravindran2002model} introduced MDP homomorphisms, which account for action equivalence with a state-dependent action mapping.~\cite{li2006towards} proposed a unified theory of abstraction, and approximate abstraction mechanisms were studied by~\cite{ferns2004metrics} and~\cite{abel2016near}. 
Most relevant to our work are SMDP homomorphisms~\cite{ravindran2003smdp} and theoretical formulations for reward-based abstractions in SMDPs in~\cite{abel2020value}.
Different from prior abstraction formulations which are reward-based, our feature-based successor homomorphism produces abstract models which are reusable across tasks.


\section{Conclusion}
\label{sec:conclusion}
We studied temporal and state abstraction in RL using SF. 
Specifically, we developed an abstract option representation with SF and presented algorithms that transfer existing ground options to new environments.
Based on the abstract options, we developed successor homomorphism, which produces abstract $\psi$-SMDPs that can be used to perform efficient planning with near-optimal performance guarantees. 
We demonstrated empirically that our algorithms transfer the options effectively and efficiently, and showed that our abstract $\psi$-SMDP models exhibit meaningful temporal semantics, with near-optimal planning performance across tasks.


\clearpage
\bibliography{reference}
\bibliographystyle{named}

\appendix
\appendix
\onecolumn
\section{Appendix}
\subsection{Algorithm}\label{subsec:algorithm}
In the main text, we have discussed a naive option grounding algorithm: 
\begin{itemize}
    \item \textbf{Algorithm~\ref{algo:lifting_original} (IRL-naive)}: naive option grounding algorithm which performs IRL over all starting states independently.
\end{itemize}
In this section, we present additional algorithms useful for option grounding:
\begin{itemize}
    \item \textbf{Algorithm~\ref{algo:lp_original} (IRL module for IRL-naive)}: the IRL algorithm adapted from ~\cite{syed2008apprenticeship}, and used by the IRL-naive option grounding algorithm (Algorithm~\ref{algo:lifting_original}) to find the option policies starting from each starting state.
    \item \textbf{Algorithm~\ref{algo:lifting_batch} (IRL-batch):} an efficient option grounding algorithm which improves IRL-naive and performs batched learning using IRL.
    \item \textbf{Algorithm~\ref{algo:lp_iterative}: (IRL module for IRL-batch)} the IRL module used by the IRL-batch option grounding algorithm (Algorithm~\ref{algo:lifting_batch}) to find option policies starting from a batch of starting states.
\end{itemize}

\subsubsection{IRL algorithm for the naive option grounding (Algorithm~\ref{algo:lp_original})}
First, we introduce the IRL module adapted from ~\cite{syed2008apprenticeship} and used by the IRL-naive option grounding algorithm:
\begin{algorithm}
\caption{IRL (used by Algorithm~\ref{algo:lifting_original} IRL-naive)}
\label{algo:lp_original}
\begin{algorithmic}[1]
\Statex{\textbf{Input:} Augmented MDP $\mdp=\langle \states', \actions', \dynamics', \reward, \gamma \rangle,$ starting state $s_\textnormal{start}$, state-action features $\feature\colon \states' \times \actions' \rightarrow \mathbbm{R}^d$, abstract option $\sf^{\abstractoption{o}} \in \mathbbm{R}^d$. }

\Statex{\textbf{Output:} ground and abstract option feature difference $\epsilon = \sum_{k=1}^{d} \epsilon_k$, option policy $\pi^o_{s_\textnormal{start}}$ }

\end{algorithmic}
\phantom{---}\text{Solve LP for visitation frequencies $\sr_{s,a}$ and feature matching errors $\epsilon_k$}
\begin{align}
    \min_{\sr,\epsilon_k} \quad & \sum_k \epsilon_k - \lambda_1 \sum_a\sr_{s_\textnormal{null}, a} + \lambda_2 \sum_s \mathbbm{1}_{\mu_{s,a_\textnormal{T}} > 0} \\
    \textnormal{s.t.} \quad 
     & \sum_a \sr_{s,a} = \mathbbm{1}_{s=s_\textnormal{start}} + \gamma \sum_{s', a} \dynamics'(s|s',a)\sr_{s',a} \qquad \forall s \in \states' \label{lp_naive:c1}\\
     & \sum_{s,a}\sr_{s,a}\feature_k(s,a) - \sf^{\abstractoption{o}}_k \leq \epsilon_k \qquad \phantom{-}\forall k\in 1, \ldots, d \label{lp_naive:c2}\\  
     & \sum_{s,a}\sr_{s,a}\feature_k(s,a) - \sf^{\abstractoption{o}}_k\geq -\epsilon_k \qquad \forall k\in 1, \ldots, d  \label{lp_naive:c3}\\
     & \sr_{s,a} \geq 0 \quad \forall s \in \states', a \in \actions'
\end{align}
\phantom{---}{Compute option policy 
    $\pi^o_{s_\textnormal{start}}(a|s) = \frac{\sr_{s,a}}{\sum_{a}\sr_{s,a}}
$}

\end{algorithm}

The algorithm finds the policy and termination condition of the option $\abstractoption{o}$ in the following two steps: 
\begin{enumerate}
    \item Compute the state-action visitation frequencies $\sr_{s,a}$ such that the corresponding expected feature vector (approximately) matches the abstract option feature $\sf^\abstractoption{o}$.
    \item Compute the option policy (which includes the termination condition) from $\sr_{s,a}$.
\end{enumerate}

\paragraph{Linear program.}
Adapting from prior work which uses linear programming approaches for solving MDPs and IRL~\cite{syed2008apprenticeship,malek2014linear,manne1960linear}, our LP aims to find the state-action visitation frequencies $\sr_{s,a}$ for all states and actions, which together (approximately) match the abstract option feature, i.e., $\sf^o_\textnormal{start} = \sum_{s,a} \sr_{s,a}\feature(s,a) \approx \sf^\abstractoption{o}$. 
In particular, $\| \sf^o_\textnormal{start} - \sf^\abstractoption{o} \|_1 \leq \sum_{k=1}^d \epsilon_k$.

\emph{\textbf{Inputs:} }
Recall that to enable the modelling of options and their termination conditions, the input augmented MDP $\mdp=\langle \states', \actions', \dynamics', \reward, \gamma \rangle$ was constructed in Algorithm~\ref{algo:lifting_original} by adding a null state $s_\textnormal{null}$ and termination action $a_\textnormal{T}$ such that $a_\textnormal{T}$ leads from any regular state to  $s_\textnormal{null}$. 

\emph{\textbf{Variables:}}
1. $\epsilon_k$: Upper bounds for the absolute difference between the (learner) learned ground option successor feature and the abstract option (expert) feature in the $k$-th dimension.

2. $\sr_{s,a}$: Expected cumulative state-action visitation of state-action pair $s,a$. Additionally, denote $\sr_s$ as the expected cumulative state visitation of $s$, i.e., $\forall s\in \states, a\in \actions$,
\begin{align}
    \sr_{s,a} = \mathbb{E}_{\mdp, \pi}[\sum_{t=0}^\infty\gamma^t\mathbbm{1}_{S_t=s, A_t=a}| s_0\sim P_\textnormal{start}], \quad \textnormal{and} \quad
    \sr_s = \mathbb{E}_{\mdp, \pi}[\sum_{t=0}^\infty\gamma^t\mathbbm{1}_{S_t=s}| s_0\sim P_\textnormal{start}]
\end{align}
where $P_\textnormal{start}$ is the distribution over starting states. In this naive option grounding algorithm the start state is a single state $s_\textnormal{start}$.
Observe that $\sr_s$ and $\sr_{s,a}$ are related by the policy $\pi\colon \states\rightarrow\actions$ as
\begin{align}
    \sr_{s,a} = \pi(a|s)\sr_s, \quad \textnormal{and}  \quad \sr_s = \sum_{a} \pi(a|s)\sr_{s} = \sum_{a}\sr_{s,a}.\label{eq:mu_relation}
\end{align}
And the Bellman flow constraint is given by either $\sr_s$ or $\sr_{s,a}$:
\begin{align}
    \sr_s = \mathbbm{1}_{s=s_\textnormal{start}} + \gamma \sum_{s',a}P(s|s', a)\pi(a|s')\sr_{s'} \iff \sum_a \sr_{s,a} = \mathbbm{1}_{s=s_\textnormal{start}} + \gamma \sum_{s', a} P(s|s',a)\sr_{s',a}\label{eq:bellman_flow}
\end{align} 
By Equation~\eqref{eq:mu_relation}, the policy $\pi$ corresponding to the state-action visitation frequencies is computed as
\begin{align}\pi_{s_\textnormal{start}}^o = \frac{\sr_{s,a}}{\sr_{s}} = \frac{\sr_{s,a}}{\sum_{a}\sr_{s,a}}.\label{eq:mu_policy}
\end{align}

\emph{\textbf{Objective function: } $\sum_k\epsilon_k - \lambda_1 \sum_a \mu_{s_\textnormal{null, a}} + \lambda_2 \sum_{s} \mathbbm{1}_{\mu_{s, a_\textnormal{T}} > 0}$} 

1. $\sum_k\epsilon_k$: the first term is the feature difference $\epsilon$ between the abstract and computed ground option.

2. $- \lambda_1 \sum_a \mu_{s_\textnormal{null, a}}$: The second term is a small penalty on the option length to encourage short options.
This is achieved by putting a small bonus (e.g., $\lambda_1 = 0.01$) on the expected cumulative visitation of the null state $s_\textnormal{null}$, which the agent reaches after using the terminate action.

3. $\lambda_2 \sum_{s} \mathbbm{1}_{\mu_{s, a_\textnormal{T}} > 0}$: The last term is a regularisation on the number of terminating states. This helps guide the LP to avoid finding mixtures of option policies which together match the feature expectation. 

In this way, the linear program can find an option policy which terminates automatically.

\emph{\textbf{Constraints: }}
Equation~\eqref{lp_naive:c1} is the Bellman flow constraint~\cite{syed2008apprenticeship,malek2014linear} which specifies how the state-action visitation frequencies are related by the transition dynamics, see Equation~\eqref{eq:bellman_flow}.
Equations~\eqref{lp_naive:c2} and~\eqref{lp_naive:c3} define the feature difference $\epsilon$ of the ground option $\sf^o_{s_\textnormal{start}} = \sum_{s,a}\sr_{s,a}\feature(s,a)$ and abstract option $\sf^\abstractaction{o}$ for the $k$-th dimension.

\emph{Step 2} derives the policy from the state-action visitation frequencies $\sr_{s,a}$, see Equation~\eqref{eq:mu_policy}.


\clearpage
\subsubsection{Grounding Abstract Options for Starting States in Batches (Algorithm~\ref{algo:lifting_batch})}
\begin{algorithm}[H]
\caption{Grounding Abstract Options (IRL-Batch)}
\label{algo:lifting_batch}
\begin{algorithmic}[1]
\Statex {\textbf{Input:} MDP $\mdp =\langle \states, \actions, \dynamics, \reward, \gamma \rangle$, abstract option $\sf^{\abstractoption{o}}$, $\epsilon_\textnormal{threshold}$ }
\Statex{\textbf{Output:} initiation set $I^\abstractoption{o}$, dictionary of ground option policies $\Pi^\abstractoption{o}$, termination probabilities $\Xi^\abstractoption{o}$}
\State{\textit{// Construct augmented MDP}}
\State{$\states' \leftarrow \states \cup \{s_\textnormal{null}\}, \actions' \leftarrow \actions\cup \{a_\textnormal{T}\}, P' \leftarrow P $}
\State{$\forall s\in \states'\colon \dynamics'(s_\textnormal{null} \mid s, a_\textnormal{T}) = 1$; $\forall a \in \actions'\colon P'(s_\textnormal{null} \mid s_\textnormal{null}, a) = 1$} 
\Statex{}
\State{\textit{// Find ground options for all starting states}}
\State{$I^\abstractoption{o}, \Pi^\abstractoption{o}, \Xi^\abstractoption{o}$ = \Call{Match-and-divide}{$ \states'\setminus\{s_\textnormal{null}\}$}}
\Statex{}
\State{\textit{// Recursive Function} }
\Function{Match-and-divide}{$c_\textnormal{start}$}
    \State{$\epsilon_{c_\textnormal{start}}, \Pi^o_{c_\textnormal{start}} \leftarrow$ IRL($\states', \actions', \dynamics', \gamma, c_\textnormal{start}, \sf^\abstractoption{o}$), \qquad \Comment{for IRL see Algorithm~\ref{algo:lp_iterative}}}
    \State{$c_\textnormal{match}$, $c_\textnormal{no-match}$, $C_\textnormal{ambiguous}$ = \Call{Classify}{$c_\textnormal{start}$, $\Pi^o_{c_\textnormal{start}}$, $\epsilon_{c_\textnormal{start}}$, $\epsilon_\textnormal{threshold}$}}
    \ForAll {$s_\textnormal{start} \in c_\textnormal{match}$}
        \State{$\pi^o_{s_\textnormal{start}} \leftarrow \Pi^o_{c_\textnormal{start}}(s_\textnormal{start})$}
        \State{$I^\abstractoption{o} \leftarrow I^\abstractoption{o} \cup \{s_\textnormal{start}\}$, $\Pi^\abstractoption{o}(s_\textnormal{start}) = \pi^o_{s_\textnormal{start}}$, $\Xi^\abstractoption{o}(s_\textnormal{start}) = \beta^o, \textnormal{ where } \beta^o(s) = \pi^o_{s_\textnormal{start}}$}
    \EndFor
    \If {$C_\textnormal{ambiguous} \neq \emptyset$}
        \ForAll {$c_i\in C_\textnormal{ambiguous}$}
            \State{ \Call{match-and-divide}{$c_i$}} \Comment{Note: $C_\textnormal{ambiguous}$ is a set of clusters}
        \EndFor
    \EndIf
    \State{\Return{$I^\abstractoption{o}, \Pi^\abstractoption{o}, \Xi^\abstractoption{o}$}}
\EndFunction

\Statex{}
\State{\textit{// Classify start states according to the policy found by IRL}}
\Function{Classify}{$c_\textnormal{start}, \Pi^o_{c_\textnormal{start}}$, $\epsilon_{c_\textnormal{start}}$,$\epsilon_\textnormal{threshold}$}
    \State{$c_\textnormal{match}, c_\textnormal{no-match}, c_\textnormal{ambiguous} \leftarrow \emptyset$}
    \ForAll {$s_\textnormal{start} \in c_\textnormal{start}$}
        \State{Execute $o$ from $s_\textnormal{start}$ to get successor feature $\sf^o_{s_\textnormal{start}}$ and terminating distribution $P^o_{s_\textnormal{start}, s'}$}
        \State{Compute feature difference $\epsilon^o_{s_\textnormal{start}} = |\sf^\abstractoption{o} - \sf^o_{s_\textnormal{start}}|$}
    \EndFor
     \State {$c_\textnormal{no-match} \leftarrow c_\textnormal{start}$ if $\min_{s_\textnormal{start}\in c_\textnormal{start}}\epsilon^o_{s_\textnormal{start}} > \epsilon_\textnormal{threshold}$ and $\epsilon_{c_\textnormal{start}} >\epsilon_\textnormal{threshold} $} 
    \State {$c_\textnormal{match} \leftarrow \{s_\textnormal{start}\in c_\textnormal{start}| \epsilon^o_{s_\textnormal{start}} \leq \epsilon_\textnormal{threshold}\}$}
    \State {$c_\textnormal{ambiguous} \leftarrow c_\textnormal{start}\setminus c_\textnormal{match}, c_\textnormal{no-match}$}
    \If{$c_\textnormal{ambiguous} \neq \emptyset$}
    \Comment{Cluster by termination distributions or successor features$^{(1)}$}
        \State{$C_\textnormal{ambiguous} \leftarrow \Call{Cluster}{c_\textnormal{ambiguous}, P^o_{s_\textnormal{start}, s'}}$} 
    \EndIf
    \State{\Return $c_\textnormal{match}$, $c_\textnormal{no-match}$, $C_\textnormal{ambiguous}$}
\EndFunction
\end{algorithmic}
\end{algorithm}

\textit{\footnotesize(1) In practice both approaches work well and address the challenge that the solution found could be a mixture of different option policies which individually yield different SF but together approximate the target SF. Clustering by SF approaches this issue by enforcing the SF of option policies from all $s_\textnormal{start}$ in a cluster to be similar and match the target. Clustering by termination distribution is an effective heuristic which groups together nearby $s_\textnormal{start}$, also resulting in options with similar SF. 
}

\vspace{5mm}
Based on the naive option grounding algorithm (Algorithm~\ref{algo:lifting_original}), we now introduce an efficient algorithm for grounding abstract options, which performs IRL over the start states in batches (Algorithm~\ref{algo:lifting_batch}).

\paragraph{Challenges.} The main challenges regarding performing batched IRL for grounding the options are: 1. By naively putting a uniform distribution over all possible starting states, the IRL LP cannot typically find the ground options which match the abstract option. Moreover, a closely related problem as well as one of the reasons for the first problem is 2. Since there are many different starting states, the state-action visitation frequencies found by the LP for matching the option feature may be a mixture of different option policies from the different starting states, and the induced ground option policies individually cannot achieve the successor feature of the abstract option.

\paragraph{Solutions.} For the first challenge, we introduce the batched IRL module (Algorithm~\ref{algo:lp_iterative}) to be used by IRL-batch. It flexibly learns a starting state distribution with entropy regularisation. 

For the second challenge, Algorithm~\ref{algo:lifting_batch} is a recursive approach where each recursion performs batched-IRL on a set of starting states. Then, by executing the options from each starting state using the transition dynamics, we prune the starting states which successfully match the abstract successor option's feature, and those where matching the abstract option is impossible. And cluster the remaining ambiguous states based on their option termination distribution (or their achieved successor features) and go to the next recursion. Intuitively, we form clusters of similar states (e.g., which are nearby and belong to a same community according to the option termination distribution). And running batched-IRL over a cluster of similar starting states typically returns a single ground option policy which applies to all these starting states.

\paragraph{Algorithm~\ref{algo:lifting_batch} (IRL-batch: Grounding Abstract Options in Batches).} The algorithm first constructs the augmented MDP with the null states and terminate actions in the same way as the naive algorithm. Then it uses a recursive function \emph{Match-And-Divide($c_\textnormal{start}$)}, which first computes the ground option policies corresponding to the set of starting states $c_\textnormal{start}$ through batched IRL (Algorithm~\ref{algo:lp_iterative}) over $c_\textnormal{start}$, then \emph{Classify} the starting states by their corresponding ground options' termination distributions or successor feature, into the following 3 categories: 1. $c_\textnormal{match}$: the start states where an abstract option can be initiated (i.e., there exists a ground option whose successor feature matches the abstract option); 2. $c_\textnormal{no-match}$: the start states where the abstract option cannot be initiated; and 3. $C_\textnormal{ambiguous}$: a set of clusters dividing the remaining ambiguous states. If $C_\textnormal{ambiguous}$ is not empty, then each cluster goes through the next recursion of \emph{Match-And-Divide}. Otherwise the algorithm terminates and outputs the initiation set, option policy and termination conditions.

\begin{algorithm}
\caption{IRL (used by Algorithm \ref{algo:lifting_batch} IRL-Batch)}
\label{algo:lp_iterative}
\begin{algorithmic}[1]
\Statex{\textbf{Input:} Augmented MDP $\mdp=\langle \states', \actions', \dynamics', \reward, \gamma \rangle,$, state-action features $\feature\colon \states' \times \actions' \rightarrow \mathbbm{R}^d$, abstract option $\sf^{\abstractoption{o}} \in \mathbbm{R}^d$, a set of starting states $c_\textnormal{start}\subseteq \states' \setminus \{s_\textnormal{null}\}$. }

\Statex{\textbf{Output:} (joint over $c_\textnormal{start}$) ground and abstract option feature difference $\epsilon = \sum_{k=1}^{d} \epsilon_k$, dictionary of ground option policy $\Pi^o(s_\textnormal{start})$ for all start states in $c_\textnormal{start}$ }
\end{algorithmic}
\phantom{-------}\emph{// Step 1: Solve LP for state-action visitation frequencies $\sr_{s,a}$}
\begin{align*}
    \min_{\sr,\epsilon_k, p^\textnormal{start}} \quad & \sum_k \epsilon_k - \lambda_1 \sum_a\sr_{s_\textnormal{null}, a} + \lambda_2 \sum_s p_{s}^\textnormal{start}\log p_{s}^\textnormal{start} + \lambda_3 \sum_{s} \mathbbm{1}_{\mu_{s, a_\textnormal{T}} > 0} & \\
    \textnormal{s.t.} \quad 
     & \sum_a \sr_{s,a} = p_{s}^\textnormal{start} + \gamma \sum_{s', a} \dynamics'(s|s',a)\sr_{s',a} & \qquad \forall s \in \states' \\
     & \sum_{s,a}\sr_{s,a}\feature_k(s,a) - \sf^{\abstractoption{o}}_k \leq \epsilon_k &  \forall k\in 1, \ldots, d \\  
     &\sum_{s,a}\sr_{s,a}\feature_k(s,a) - \sf^{\abstractoption{o}}_k\geq -\epsilon_k & \forall k\in 1, \ldots, d\\
     & \sr_{s,a} \geq 0 & \forall s \in \dynamics', a \in \actions'\\
     & \sum_{s\in c_\textnormal{start}}p_s^\textnormal{start} = 1 &\\
     &p_s^\textnormal{start} \geq 0
     & \forall s\in c_\textnormal{start}\\
     & p_s^\textnormal{start} = 0 &\forall s\notin c_\textnormal{start}
\end{align*}
\phantom{-------}\emph{// Step 2: Compute dictionary of option policies $\Pi^o$,  $\forall s_\textnormal{start} \in c_\textnormal{start}$:}
\begin{align*}
     \forall s\in \states,a\in \actions', \pi^o_{s_\textnormal{start}}(a|s) = \frac{\sr_{s,a}}{\sum_{a}\sr_{s,a}}, \quad \textnormal{ then add to dictionary} \quad 
     \Pi^o(s_\textnormal{start}) \leftarrow \pi^o_{s_\textnormal{start}}
\end{align*}
\end{algorithm}

\paragraph{Algorithm~\ref{algo:lp_iterative} (IRL module for IRL-batch).}
Compared with the IRL algorithm presented in Algorithm~\ref{algo:lifting_original}, this batched algorithm performs IRL over a batch of starting states. As we discussed in the above challenges, naively putting a uniform distribution over all possible starting states would typically fail. Therefore, we enable the LP to \emph{learn a starting state distribution $p^\textnormal{start}_s$} which best matches the abstract option feature. However, the LP would again reduce to a single starting state which best matches the abstract option. To counteract this effect, we add a small entropy regularisation to the objective to increase the entropy of the starting state distribution, i.e., we add $\lambda_2 \sum_s p_s^\textnormal{start}\log p_s^\textnormal{start}$.
Clearly, entropy is not a linear function.
In practise, the objective is implemented using a piecewise-linear approximation method provided by the Gurobi optimisation package.\footnote{\url{https://www.gurobi.com/}}

\subsection{Feature-based (Variable-reward) SMDP Abstraction}
In this section, we provide a general framework for feature-based abstraction using the variable-reward SMDPs. The state mapping and action mapping functions $f(s), g_s(o)$ can be defined to instantiate a new abstraction method.
\begin{definition}[Abstract $\sf$-SMDP]\label{def:abstract_sf_smdp} Let $\mdp = \langle \states, \options, \dynamics, \sf, \gamma \rangle$ be a ground $\sf$-SMDP. We say that $\abstractmdp = \langle \abstractstates, \abstractoptions, \abstractdynamics, \abstractsf, \gamma \rangle$ is an \emph{abstract $\sf$-SMDP of $\mdp$} if there exists (1) a state abstraction mapping $f\colon \states\rightarrow \abstractstates$ which maps each ground state to an abstract state, (2) a weight function $w\colon \states \rightarrow [0,1]$ over the ground states such that $\forall \abstractstate{s}\in \abstractstates, \sum_{s\in \smap^{-1}(\abstractstate{s})}w_s = 1$, (3) a state-dependent option abstraction mapping $g_s\colon \options \rightarrow \abstractoptions$, and (4) the abstract transition dynamics and features are\\
\centerline{
$\abstractdynamics_{\abstractstate{s}, \abstractstate{s}'}^{\abstractoption{o}} = \!\!\!\!\!\sum\limits_{s\in \smap^{-1}(\abstractstate{s})}\!\!\!\!\!w_s\sum\limits_{s' \in \smap^{-1}{(\abstractstate{s}')}}\!\!\!\!\!\dynamics_{s, s'}^{g^{-1}_s(\abstractoption{o})}
\textnormal{ , and } \quad
\abstractsf_{\abstractstate{s}}^{\abstractoption{o}} = \!\!\!\!\!\sum\limits_{s\in \smap^{-1}(\abstractstate{s})} \!\!\!\!\!w_s \sf_{s}^{g^{-1}_s(\abstractoption{o})}.
$}
\end{definition}

\subsection{Reward-based MDP and SMDP abstraction}\label{subsec:mdp_abstraction}
In the following we define the abstract MDP and abstract SMDPs, following the conventional notations of ~\cite{li2006towards,abel2016near,ravindran2003smdp}.

\begin{definition}[Abstract MDP]\label{def:abstract_mdp}
Let $\mdp = \langle \states, \actions, \dynamics, \reward, \gamma \rangle$ be a ground MDP. We say that $\abstractmdp = \langle \abstractstates, \actions, \abstractdynamics, \abstractreward, \gamma \rangle$ is an \emph{abstract MDP of $\mdp$} if there exists (1) a state abstraction mapping $f\colon \states\rightarrow \abstractstates$, which maps each ground state to an abstract state, (2) a weight function $w\colon \states \rightarrow [0,1]$ for the ground states such that $\forall \abstractstate{s}\in \abstractstates$, $ \sum_{s\in \smap^{-1}(\abstractstate{s})}w_s = 1$, and (3) a state-dependent action mapping $g_s\colon \actions\rightarrow\abstractactions$, the abstract transition dynamics and rewards are defined as
\centerline{$
\abstractdynamics_{\abstractstate{s}, \abstractstate{s}'}^\abstractaction{a} = \!\!\!\!\!\sum\limits_{s\in \smap^{-1}(\abstractstate{s})}\!\!\!\!\! w_s \sum\limits_{s'\in \smap^{-1}(\abstractstate{s}')} \!\!\!\!\!\dynamics_{s,s'}^{\omap_s^{-1}(\abstractaction{a})} \textnormal{ , and}\quad
\abstractreward_{\abstractstate{s}}^ \abstractaction{a} = \!\!\!\!\! \sum\limits_{s\in \smap^{-1}(\abstractstate{s})} \!\!\!\!\!w_s r_s^{\omap^{-1}_s(\abstractaction{a})}.$}
\end{definition}

In cases in which multiple ground actions map to the same abstract action, $g^{-1}_s(\abstractaction{a})$ picks one of the ground actions. $g_s$ is commonly defined as the identity mapping~\cite{li2006towards,abel2016near}. We now generalize this definition to abstract SMDPs:

\begin{definition}[Abstract SMDP]\label{def:abstract_smdp} Let $\mdp = \langle \states, \options, \dynamics, \reward, \gamma \rangle$ be a ground SMDP. We say that $\abstractmdp = \langle \abstractstates, \abstractoptions, \abstractdynamics, \abstractreward, \gamma \rangle$ is an \emph{abstract SMDP of $\mdp$} if there exists (1) a state abstraction mapping $f\colon \states\rightarrow \abstractstates$ which maps each ground state to an abstract state, (2) a weight function $w\colon \states \rightarrow [0,1]$ over the ground states such that $\forall \abstractstate{s}\in \abstractstates, \sum_{s\in \smap^{-1}(\abstractstate{s})}w_s = 1$, (3) a state-dependent option abstraction mapping $g_s\colon  \times\options \rightarrow \abstractoptions$, and (4) the abstract transition dynamics and rewards are\\
\centerline{
$\abstractdynamics_{\abstractstate{s}, \abstractstate{s}'}^{\abstractoption{o}} = \!\!\!\!\!\sum\limits_{s\in \smap^{-1}(\abstractstate{s})}\!\!\!\!\!w_s\sum\limits_{s' \in \smap^{-1}{(\abstractstate{s}')}}\!\!\!\!\!\dynamics_{s, s'}^{g^{-1}_s(\abstractoption{o})}
\textnormal{ , and } \quad
\abstractreward_{\abstractstate{s}}^{\abstractoption{o}} = \!\!\!\!\!\sum\limits_{s\in \smap^{-1}(\abstractstate{s})} \!\!\!\!\!w_s \reward_{s}^{g^{-1}_s(\abstractoption{o})}.
$}
\end{definition}
In cases in which multiple ground options map to the same abstract option, $g^{-1}_s(\abstractoption{o})$ picks one of the ground options, e.g., the option of shortest duration, maximum entropy, etc. 

\subsection{Relation to Other MDP Abstraction Methods}\label{subseq:other_abstractions}

The framework of MDP abstraction was first introduced by~\cite{dean1997model} through stochastic bisimulation, and~\cite{ravindran2002model} extended it  to MDP homomorphisms. 
Later, \cite{li2006towards} classified exact MDP abstraction into 5 categories and 
\cite{abel2016near} formulated their approximate counterparts: model-irrelevance, $Q^\pi$-irrelevance, $Q^*$-irrelevance, $a^*$-irrelevance and $\pi^*$-irrelevance abstractions.
Our successor homomorphism follows the formulation of MDP homomorphism~\cite{ravindran2003smdp}, which broadly fall into the category of model-irrelevance abstraction, where states are aggregated according to their one-step/multi-step transition dynamics and rewards.
On the other hand, abstraction schemes which aggregate states according to their Q-values are $Q^*$-irrelevance abstraction and $a^*$ abstraction, where $Q^*$ aggregate states according to all actions, while $a^*$ aggregate states with the same optimal action and Q-value, cf.\ Figure~\ref{fig:abstract_mdp_graphs_smaller} for an illustration of the induced abstract MDPs.
Different from prior abstraction formulations (Definition~\ref{def:abstract_mdp}) which are reward-based, our feature-based successor homomorphism produces abstract models with meaningful temporal semantics, and is robust under task changes. Furthermore, we include a generic formulation of feature-based (variable-reward) abstraction (Definition~\ref{def:abstract_sf_smdp}), which provides a basis for potential feature-based abstractions other than successor homomorphism.

\subsection{Proof for Theorem~\ref{thm}}
\thm*

\begin{proof}
Given $\epsilon$-Approximate Successor Homomorphism: $h = (\smap(s), \omap_s(o), w_s)$ from ground $\sf$-SMDP $\mdp=\langle \states, \options, \dynamics, \sf, \gamma\rangle$ to abstract $\sf$-SMDP $\abstractmdp=\langle \abstractstates, \abstractoptions, \abstractdynamics, \sf, \gamma\rangle$, such that $\forall s_1, s_2\in \states, o_1,o_2\in \options$, $h(s_1, o_1) = h(s_2, o_2) \implies$
\begin{align}
|\sum_{s_j\in \smap^{-1}(\sa')} P_{s_1, s_j}^{o_1} - \sum_{s_j\in \smap^{-1}(\sa')}P_{s_2, s_j}^{o_2} | &\leq \epsilon_P\\
|\sf_{s_1}^{o_1} - \sf_{s_2}^{o_2}| &\leq \epsilon_{\sf}
\end{align}
The abstract transition dynamics and features are
\begin{align}
    P_{\sa, \sa'}^{\oa} = \sum_{s\in \smap^{-1}(\sa)} w_{s}\sum_{s'\in \smap^{-1}(\sa')} P_{s, s'}^{\omap^{-1}_s(\oa)}, \textnormal{ and }
    \sf_{\sa}^{\oa}= \sum_{s\in \smap^{-1}(\sa)} w_{s} \sf_{s}^{\omap^{-1}_{s}(\oa)}. \label{eq_proof:abstraction}
\end{align}
We show that given features $\feature: \states\times\actions\rightarrow \mathbb{R}^d$ in the underlying (feature-based) MDP $\mdp=\langle\states, \actions, \dynamics, \feature, \gamma \rangle$, and linear reward function on the features $w_r\in\mathbb{R}^d$, the difference in value of the optimal policy in the induced abstract SMDP and the ground SMDP is bounded by $\frac{2\kappa}{(1-\gamma)^2}$, where $\kappa = |w_r|(2\epsilon_{\sf} + \frac{\epsilon_P|\abstractstates|\max_{s,a}|\feature(s,a)|}{1-\gamma}).$

To show the above error bound, we extend the proof of~\cite{abel2016near} for the error bound induced by approximate MDP abstraction, which follows the following three steps:

\emph{Step 1: Show that $\forall s_1\in\states, o_1\in \options, \sa = \smap(s_1), \oa = \omap(o_1) \implies |Q(\sa, \oa) - Q(s_1, o_1)|\leq \frac{\epsilon}{1-\gamma}.$}

\begin{align}\label{eq_proof:r}
    r_{\sa}^{\oa} 
        = w_r^T\sf_{\sa}^{\oa}
        = \sum_{s_i\in \smap^{-1}(\sa)} w_{s_i} w_r^T\sf_{s_i}^{\omap^{-1}_{s_i}(\oa)}
        = \sum_{s_i\in \smap^{-1}(\sa)} w_{s_i} r_{s_i}^{\omap^{-1}_{s_i}(\oa)} \textnormal{ denote } r_{s_i}^{\omap^{-1}_{s_i}(\oa)} \textnormal{ as } w_{s_i} r_{s_i}^{o_i}
\end{align}
\begin{align}
   \!\!\!\!\!\!\!\!\!\! &Q(\sa, \oa)
        = \mathbb{E}[r_{\sa}^{\oa} + \gamma^k \max_{\oa'}Q(\sa', \oa')] 
        \stackrel{\textnormal{Eq.}\eqref{eq_proof:abstraction},\eqref{eq_proof:r}}{=} 
        \!\!\!\!\!\sum_{s_i\in \smap^{-1}(\sa)}w_{s_i} r_{s_i}^{o_i} + \!\!\!\!\!\sum_{s_i\in \smap^{-1}(\sa)} w_{s_i} \sum_{\sa' \in \abstractstates'}\sum_{s_j\in \smap^{-1}(\sa')} \!\!\!\!\!P_{s_i, s_j}^{\omap^{-1}(\oa)} \max_{\oa'}Q(\sa', \oa')\\
    &Q(s_1, o_1)
        = \mathbb{E}[r_{s_1}^{o_1} + \gamma^k \max_{o'_j}Q(s'_j, o'_j)]
        = r_{s_1}^{o_1} + \sum_{\sa' \in \abstractstates'}\sum_{s_j\in \smap^{-1}(\sa')} P_{s_1, s_j}^{o_1} \max_{\oa'}Q(s_j, o'_j)
\end{align}
\begin{align}
    &|Q(\sa, \oa) - Q(s_1, o_1)|\label{eq_proof:q_diff}\\
        &= |\sum_{s_i\in \smap^{-1}(\sa)}w_{s_i} (r_{s_i}^{o_i} - r_{s_1}^{o_1}) + 
        \sum_{s_i\in \smap^{-1}(\sa)} w_{s_i} \sum_{\sa' \in \abstractstates'}\sum_{s_j\in \smap^{-1}(\sa')} \left( P_{s_i, s_j}^{\omap^{-1}_{s_i}(\oa)} \max_{\oa'}Q(\sa', \oa') - P_{s_1, s_j}^{o_1} \max_{o'_j} Q(s'_j, o'_j)\right)|\\
        &\leq \underbrace{|\sum_{s_i\in \smap^{-1}(\sa)}w_{s_i}w^T_r(\sf_{s_i}^{o_i}- \sf_{s_1}^{o_1})| + |\sum_{s_i\in \smap^{-1}(\sa)} w_{s_i} \sum_{\sa' \in \abstractstates'}\sum_{s_j\in \smap^{-1}(\sa')} \left( P_{s_i, s_j}^{\omap^{-1}_{s_i}(\oa)} \max_{\oa'}Q(\sa', \oa') - P_{s_1, s_j}^{o_1} \max_{\oa'} Q(\sa', \oa')\right)|}_{(1)}\\ 
        &+ \underbrace{|\sum_{s_i\in \smap^{-1}(\sa)} w_{s_i} \sum_{\sa' \in \abstractstates'}\sum_{s_j\in \smap^{-1}(\sa')} P_{s_1, s_j}^{o_1} \left(\max_{\oa'} Q(\sa', \oa')- \max_{o'_j} Q(s'_j, o'_j)\right)}_{(2)}|
\end{align}
\begin{align}
    (1)
        &\leq 2 |w_r|\epsilon_{\sf} + |\sum_{s_i\in \smap^{-1}(\sa)} w_{s_i} \sum_{\sa' \in \abstractstates'}\sum_{s_j\in \smap^{-1}(\sa')} \left( P_{s_i, s_j}^{\omap^{-1}_{s_i}(\oa)} - P_{s_1, s_j}^{o_1}\right) \max_{\oa'} Q(\sa', \oa')| \phantom{-}\\
            &\leq 2 |w_r|\epsilon_{\sf} + \sum_{\sa' \in \abstractstates'}\epsilon_P \max_{\oa'} Q(\sa', \oa')
        \leq 2 |w_r|\epsilon_{\sf}+|\abstractstates|\epsilon_P \frac{|w_r|\max_{s,a}|\feature(s,a)|}{1-\gamma}\\ 
        &\textnormal{ Denote } \kappa =  |w_r|\left(2\epsilon_{\sf}+\frac{|\abstractstates|\epsilon_P\max_{s,a}|\feature(s,a)|}{1-\gamma}\right)\\
    (2)
        &= \underbrace{|\sum_{s_i\in\smap^{-1}(\sa)}w_{s_i}}_{=1}\underbrace{\sum_{\sa' \in \abstractstates'}\sum_{s_j\in \smap^{-1}(\sa')} P_{s_1, s_j}^{o_1}}_{(3)\leq\gamma}\left( \max_{\oa'}Q(\sa', \oa') - \max_{o'_j}Q(s'_j, o'_j)\right)|\\
        & \leq \gamma \max_{s\in \states}|\max_{\oa}Q(\smap(s), \oa) - \max_{o}Q(s, o)| \quad \textnormal{Let }o^* = \arg\max_o Q(s,o)\\
        & \leq \gamma \max_{s\in \states}|Q(\smap(s), \omap_s(o^*)) - Q(s, o^*)| \quad \textnormal{ which goes back to Equation~\eqref{eq_proof:q_diff}}.
\end{align}
(3) is since the option lasts at least one step and terminates with probability 1:\\
$\sum_{\sa' \in \abstractstates'}\sum_{s_j\in \smap^{-1}(\sa')} \sum_{k=1}^{\infty} P_{s_1, s_j, k}^{o_1} = 1$\\
Hence (3) $ = \sum_{\sa' \in \abstractstates'}\sum_{s_j\in \smap^{-1}(\sa')} \sum_{k=1}^{\infty} \gamma^k P_{s_1, s_j,k}^{o_1} \leq \sum_{\sa' \in \abstractstates'}\sum_{s_j\in \smap^{-1}(\sa')} \gamma^{k=1} P_{s_1, s_j,k}^{o_1} = \gamma$.


\begin{align}
    |Q(\sa, \oa) - Q(s_1, o_1)| 
        &\leq (1) + (2) \\
        &\leq  \kappa + \gamma \underbrace{\max_{s\in\states}[|Q(\smap(s), \omap_s(o^*)) - Q(s, o^*)|}_{(4)}] \quad \textnormal{, expand (4) again using Equation~\eqref{eq_proof:q_diff}}\\
        &\leq \kappa + \gamma\left(\kappa + \gamma \max_{s\in\states}[|Q(\smap(s), \omap_s(o^*)) - Q(s, o^*)|]\right)\\
        &\leq  \kappa + \gamma\kappa + \gamma^2\kappa + \ldots \leq \frac{\kappa}{1-\gamma}
\end{align}

\emph{Step 2: The optimal option in the abstract MDP has a Q-value in the ground MDP that is nearly optimal, i.e.:}
\begin{equation}
    \forall s_1\in \states, Q(s_1, o^*_1) - Q(s_1, \omap^{-1}_{s_1}(\oa^*)) \leq \frac{2\kappa}{1-\gamma}
\end{equation}
where $o^*_1 = \arg\max_{o}Q(s_1, o) \textnormal{ and } \oa^* = \arg\max_{\oa}Q(\smap(s_1), \oa)$.

From step 1, we have $|Q(\sa, \oa) - Q(s_1, o_1)| \leq \frac{\kappa}{1-\gamma}$. \textnormal{Then, by definition of optimality, }
\begin{align}
    & Q(\sa, \omap_{s_1}(o^*_1)) \leq Q(\sa, \oa^*) \implies Q(\sa, \omap_{s_1}(o^*_1)) + \frac{\kappa}{1-\gamma} \leq Q(\sa, \oa^*) + \frac{\kappa}{1-\gamma}\\
    & \textnormal{Therefore, by step 1 then by optimality: } Q(s_1, o^*_1) \leq Q(\sa, \omap_{s_1}(o^*_1)) + \frac{\kappa}{1-\gamma} \leq Q(\sa, \oa^*) + \frac{\kappa}{1-\gamma} \label{eq_proof:qg_leq_qa}\\
    & \textnormal{Again by step 1: } Q(\sa, \oa^*)  \leq Q(s_1, g_{s_1}^{-1}(\oa^*)) + \frac{\kappa}{1-\gamma}\label{eq_proof:qa_leq_qg}\\
    &\textnormal{Therefore, } Q(s_1, o^*_1) \stackrel{Eq.\eqref{eq_proof:qg_leq_qa}}{\leq} Q(\sa, \oa^*) + \frac{\kappa}{1-\gamma} 
    \stackrel{Eq.\eqref{eq_proof:qa_leq_qg}}{\leq} Q(s_1, \omap^{-1}(\oa^*)) + \frac{2\kappa}{1-\gamma}
\end{align}

\emph{Step 3: The optimal abstract SMDP policy yields near optimal performance in the ground SMDP:}

Denote $\omap^{-1}(\pia)$ as the ground SMDP policy implementing the abstract SMDP policy $\pia$, i.e., at state $s$, the ground option corresponding to the abstract option $\oa$ chosen by the abstract policy is $\omap_s^{-1}(\oa)$.
\begin{align}
    &Q^{\pi^*}(s, o^*) - Q^{\omap^{-1}(\pia^*)} (s, \omap_s^{-1}(\oa^*))\label{eq_proof:q_pi}\\
        &\leq \frac{2\kappa}{1-\gamma} + Q^{\pi^*}(s, \omap_s^{-1}(\oa^*)) - Q^{\omap^{-1}_{s}(\pia^*)} (s, \omap^{-1}(\oa^*)) \textnormal{ , by step 2}\\
        &= \frac{2\kappa}{1-\gamma} + [r_{s}^{\omap^{-1}_s(\oa^*)} + \sum_{s'\in S} P_{s, s'}^{\omap^{-1}_s(\oa^*)} Q^{\pi^*}(s', o^*_{s'})] - [r_{s}^{\omap^{-1}_s(\oa^*)} + \sum_{s'\in S} P_{s, s'}^{\omap^{-1}_s(\oa^*)} Q^{\omap^{-1}(\pia^*)}(s', \omap^{-1}(o^*_{\sa'}))]\\
        &= \frac{2\kappa}{1-\gamma} +  \underbrace{\sum_{s'\in S} P_{s, s'}^{\omap^{-1}_s(\oa^*)}}_{\textnormal{(1)}\leq \gamma} [Q^{\pi^*}(s', o^*_{s'}) -  Q^{\omap^{-1}(\pia^*)}(s', \omap^{-1}_{s'}(o^*_{\sa'}))]\\
        &\leq \frac{2\kappa}{1-\gamma} + \gamma \max_{s'\in S}[|Q^{\pi^*}(s', o^*_{s'}) - Q^{\omap^{-1}(\pia^*)}(s', \omap^{-1}_{s'}(o^*_{\sa'}))|]\\
        &\stackrel{\textnormal{Eq.}\eqref{eq_proof:q_pi}}{\leq} \frac{2\kappa}{1-\gamma} + \gamma\left(\frac{2\kappa}{1-\gamma} + \gamma\max_{s'\in S}[Q^{\pi^*}(s', \omap^{-1}_{s'}(o^*_{\sa'})) - Q^{\omap^{-1}_{s'}(\pia^*)} (s', \omap^{-1}(o^*_{\sa'}))]\right) \\
        &\leq \frac{2\kappa}{1-\gamma} + \gamma \frac{2\kappa}{1-\gamma} + \gamma^2 \frac{2\kappa}{1-\gamma} \ldots\\
        &\leq \frac{2\kappa}{(1-\gamma)^2}
\end{align}
where (1) is because the option terminates with probability 1 and takes at least 1 step.
\end{proof}

\clearpage

\subsection{Additional Experiment Details}\label{subsec:experiment_details}

\subsubsection{Additional details on the experimental settings}\label{app:settings}
\begin{figure}[H]
    \centering
	\subfloat[2 Rooms, 157 states]{\includegraphics[width=0.21\linewidth, height=0.15\linewidth]{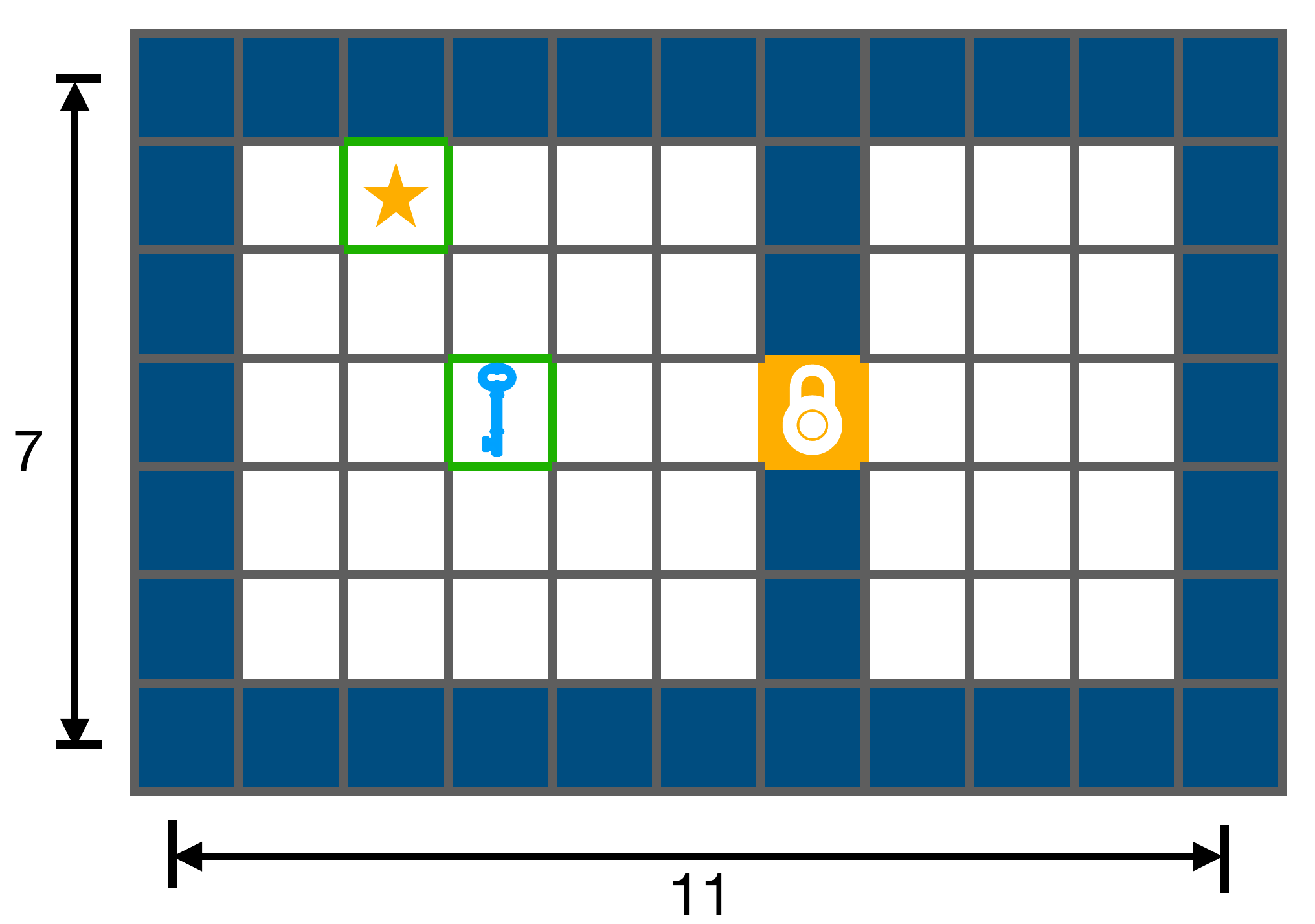}}%
	\hfill
	\subfloat[3 Rooms, 487 states]{\includegraphics[width=0.25\linewidth, height=0.15\linewidth]{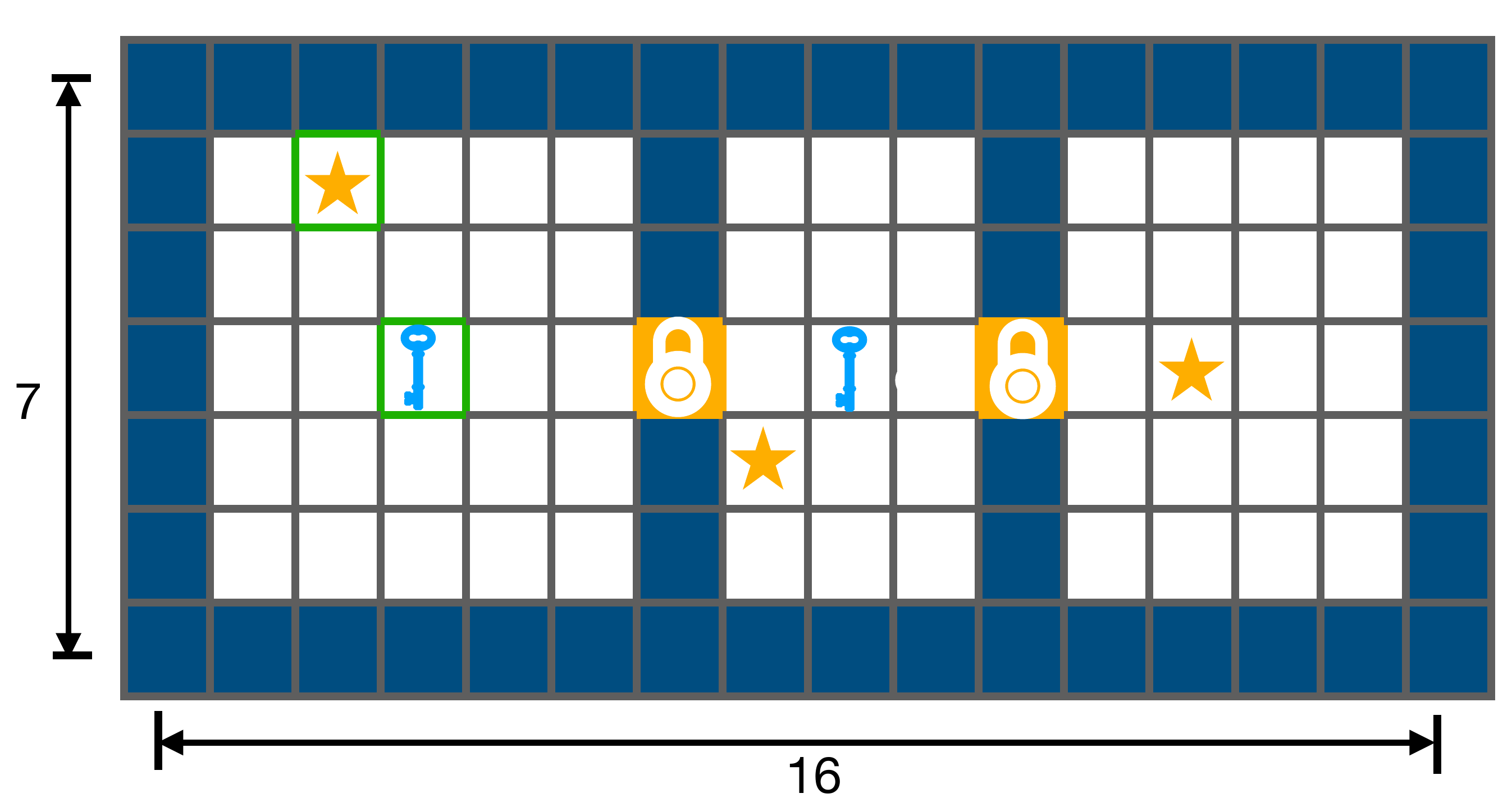}}%
	\hfill
	\subfloat[4 Rooms (small), 348 states ]{\includegraphics[width=0.25\linewidth, height=0.23\linewidth]{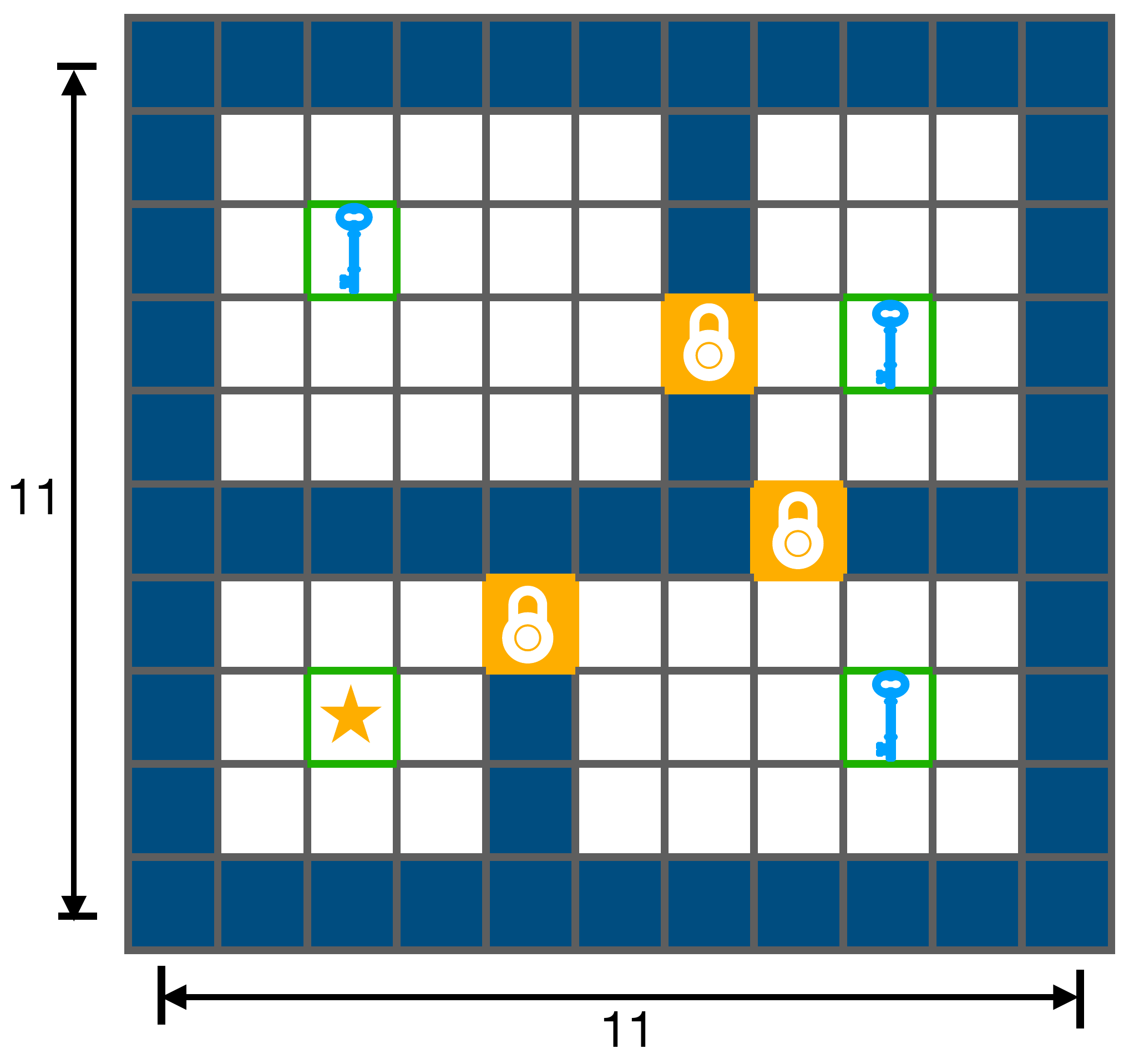}}%
	\hfill
	\subfloat[4 Rooms (large), 1463 states ]{\includegraphics[width=0.26\linewidth, height=0.24\linewidth]{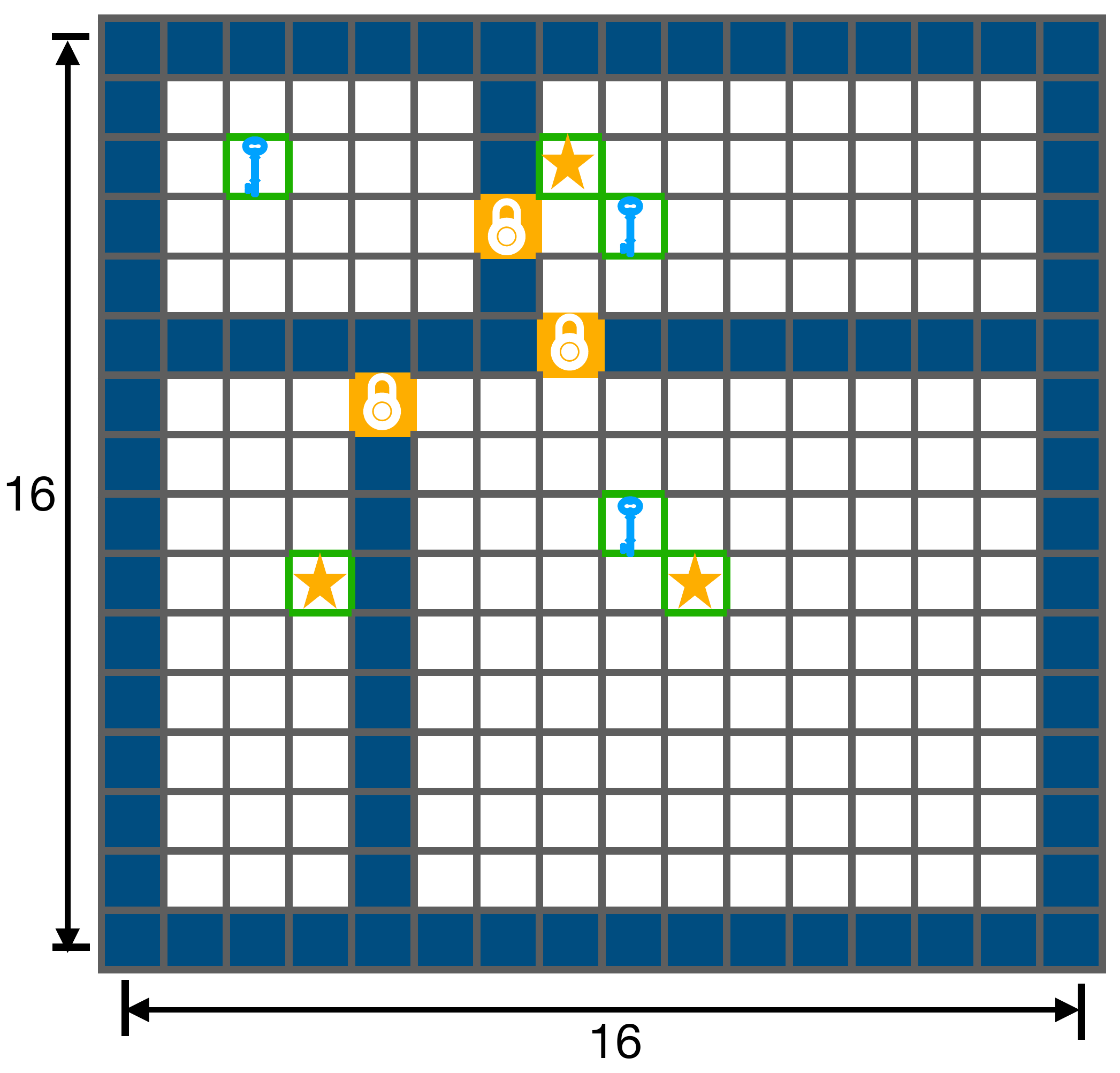}}%
	\caption{Object-room layouts used in our experiments. In each object room instance, blue grids are the walls, orange grids are doors, and there are two types of objects: stars and keys, which can be picked-up by the agent. The agent can pick up a key and use it to open a door.}
	\label{fig:layouts}
\end{figure}
\emph{Object-Rooms}: $N$ rooms are connected by doors with keys and stars inside. There are 6 actions, i.e., $\actions=\{$Up, Down, Left, Right, Pick up, Open$\}$. The agent can pick up the keys and stars, and use a key to open a door next to it. The agent starts from the (upper) left room.

\emph{Settings for Table~\ref{table:key_door_option_fitting}:}
The source room where the agent demonstrates and encodes the abstract options is the 2 Room variant.
The 2-4 target rooms setting used for grounding the options are the 2 Rooms in Figure~\ref{fig:layouts}~(a), 3 Rooms in Figure~\ref{fig:layouts}~(b), and 4 Rooms (large) in Figure~\ref{fig:layouts}~(d) variants. 

\emph{Settings for Figure~\ref{fig:abstract_mdp_graphs_smaller} and Figure~\ref{fig:abstract_mdp_planning_4R}:}
The abstract $\sf$-SMDP is generated using the setting 4 Room (small) in Figure~\ref{fig:layouts}~(c), where the first 3 rooms each contain a key and a star is in the final room. For Figure~\ref{fig:abstract_mdp_planning_4R}, the task specifications are as follows:
We refer to transfer as the task transfer, i.e., change of reward function. The total reward is discounted and normalized by the maximum reward achieved.
\begin{itemize}[topsep=0pt]
    \item dense reward (no transfer): the agent receives a reward for each key picked up, door opened, and star picked up, i.e., the reward vector $w_r = [1,1,1]$ over the features (key, open door, star).
    \item sparse reward (no transfer): the agent receives a reward for each door opened, i.e., the reward vector $w_r = [0,1,0]$.
    \item transfer (w. overlap): overlap refers to the overlap between reward function in the source and target task. In the source task, the agent receives a reward for each key picked up, and each door opened, i.e., the reward vector $w_r = [1,1,0]$. In the target task, the agent receives a reward for each door opened and star picked up, i.e., the reward vector $w_r = [0,1,1]$. 
    \item transfer (w.o. overlap): In the source task, the agent receives a reward for each star picked up, i.e., the reward vector $w_r = [0,0,1]$. In the target task, the agent receives a reward for each key picked up, i.e., the reward vector $w_r = [1,0,0]$. 
\end{itemize}

\emph{Settings for Figure~\ref{fig:abstract_mdp_planning} and Figure~\ref{fig:abstract_mdp_graphs_larger}:}
The abstract $\sf$-SMDP is generated using the 3 Room setting in Figure~\ref{fig:layouts}~(b), where the first 2 rooms each contains a key and a star, and the final room contains a star. For figure~\ref{fig:abstract_mdp_planning}, the task specifications is the same as the above description for Figure~\ref{fig:abstract_mdp_planning_4R}.

\subsubsection{Additional Experiment Results:}
In this section, we present additional experimental results.

\begin{figure}
    \centering
    \includegraphics[width=0.2\linewidth]{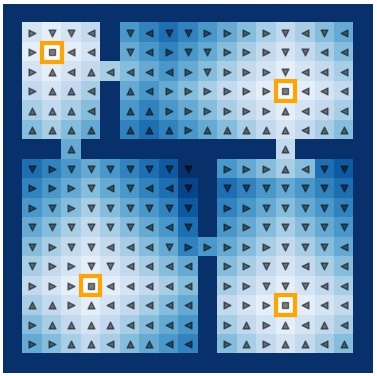}
	\caption{Ground options learned in the 4-Room domain via Algorithm~\ref{algo:lifting_batch} (presented in the Appendix). The features correspond to indicators of the room centers (marked in the orange square). Hence, the ground option should move to any one of the four room centres.
The option policy accurately takes the agent from each starting state to a nearby room centre and then terminates.}
    \label{fig:4_room}
\end{figure}

\paragraph{Classic 4-Rooms.}
Figure~\ref{fig:4_room} shows the ground option policy learned by \emph{IRL-batch} on the 4-Room domain.
The option is learned by solving $1$ linear program by \emph{IRL-batch}, i.e., the state-action visitation frequencies returned by the IRL program corresponds to an optimal policy for all starting states. Please refer to the figure for details of the option.

\begin{figure*}[t]
	\centering
	\subfloat[State visitation]{\includegraphics[width=0.25\linewidth, height=0.24\linewidth]{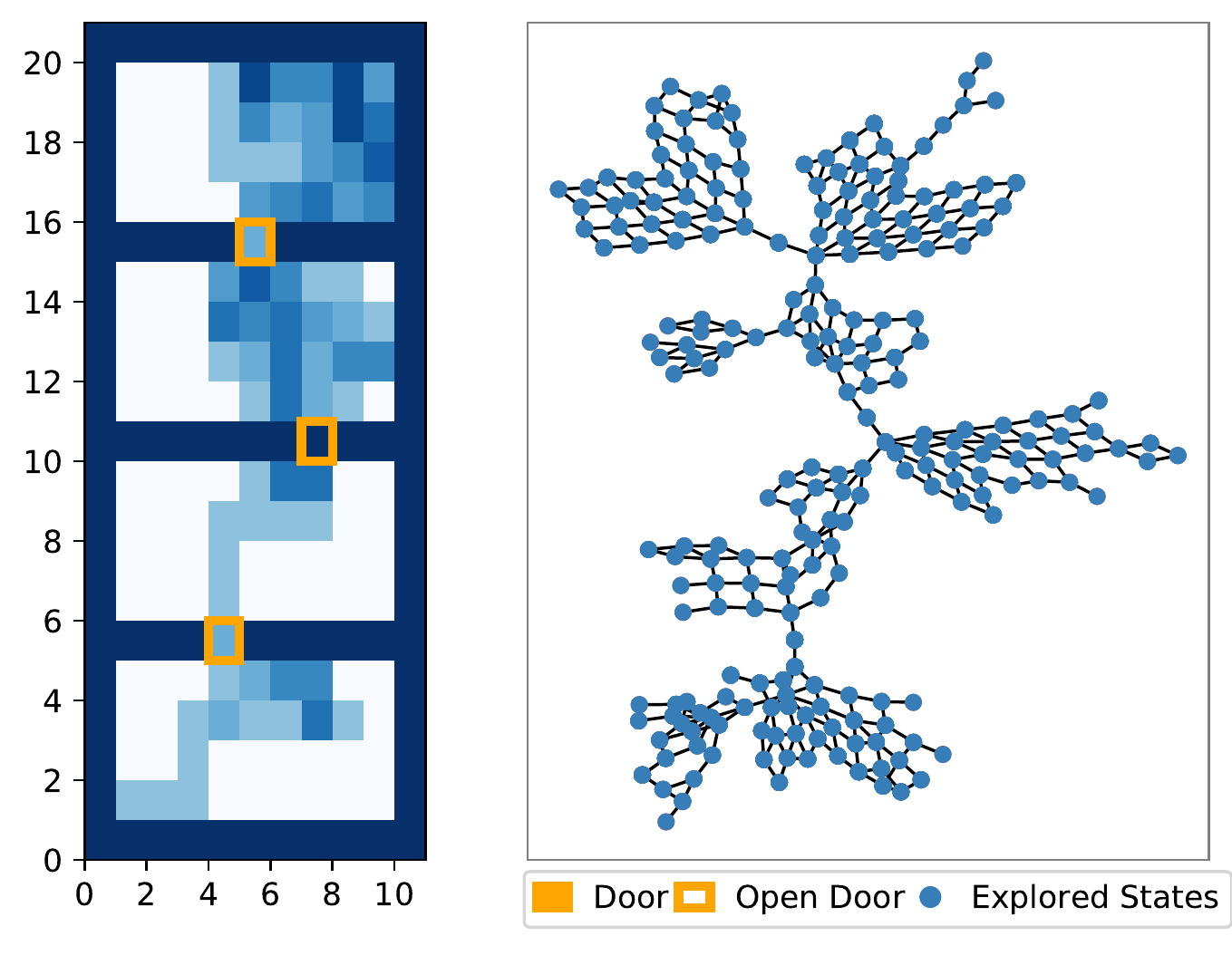}}%
	\hfill
	\subfloat[number of states explored]{\includegraphics[width=0.23\linewidth, height=0.22\linewidth]{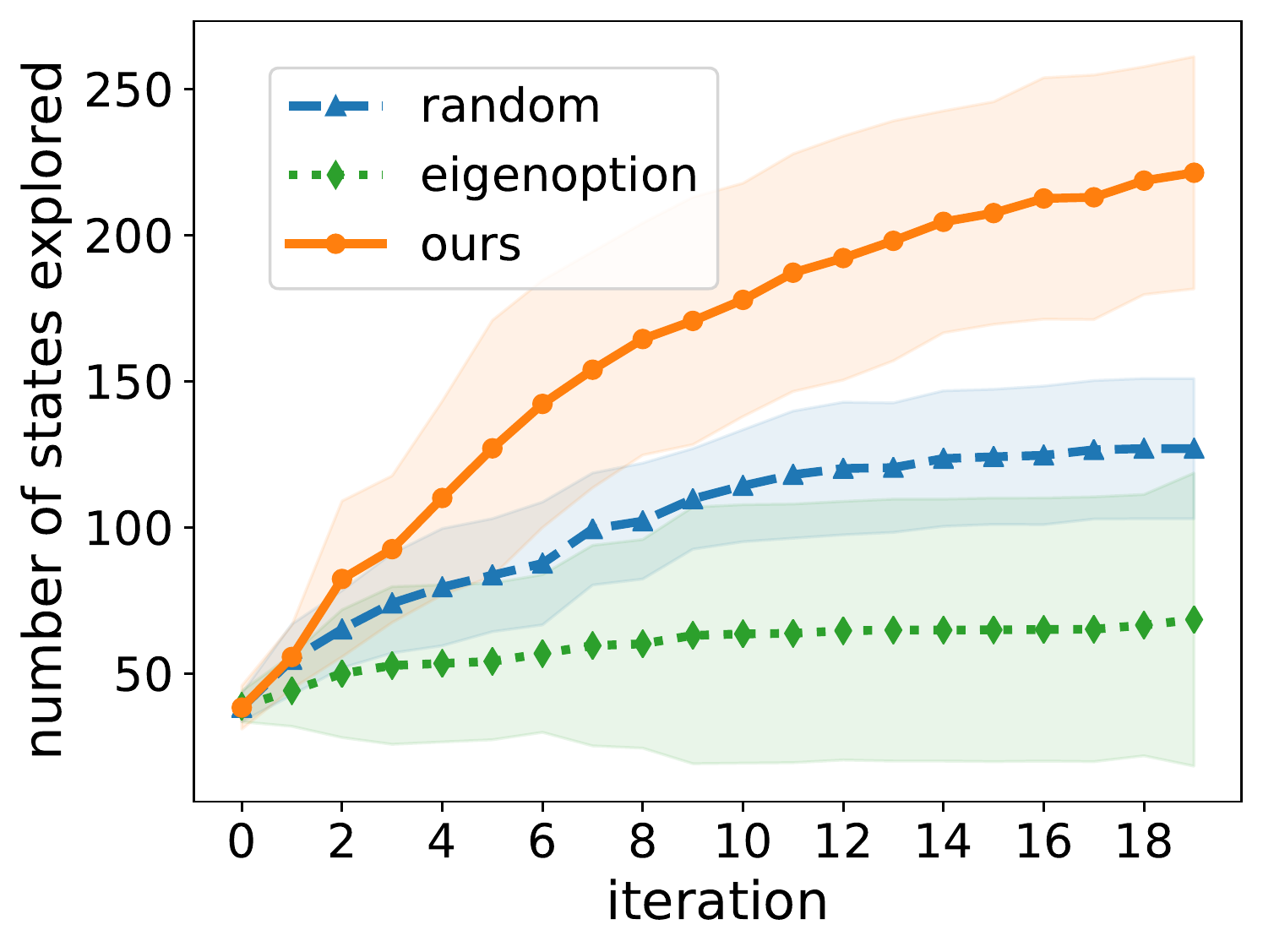}}%
	\hfill
	\subfloat[number of doors opened]{\includegraphics[width=0.23\linewidth, height=0.22\linewidth]{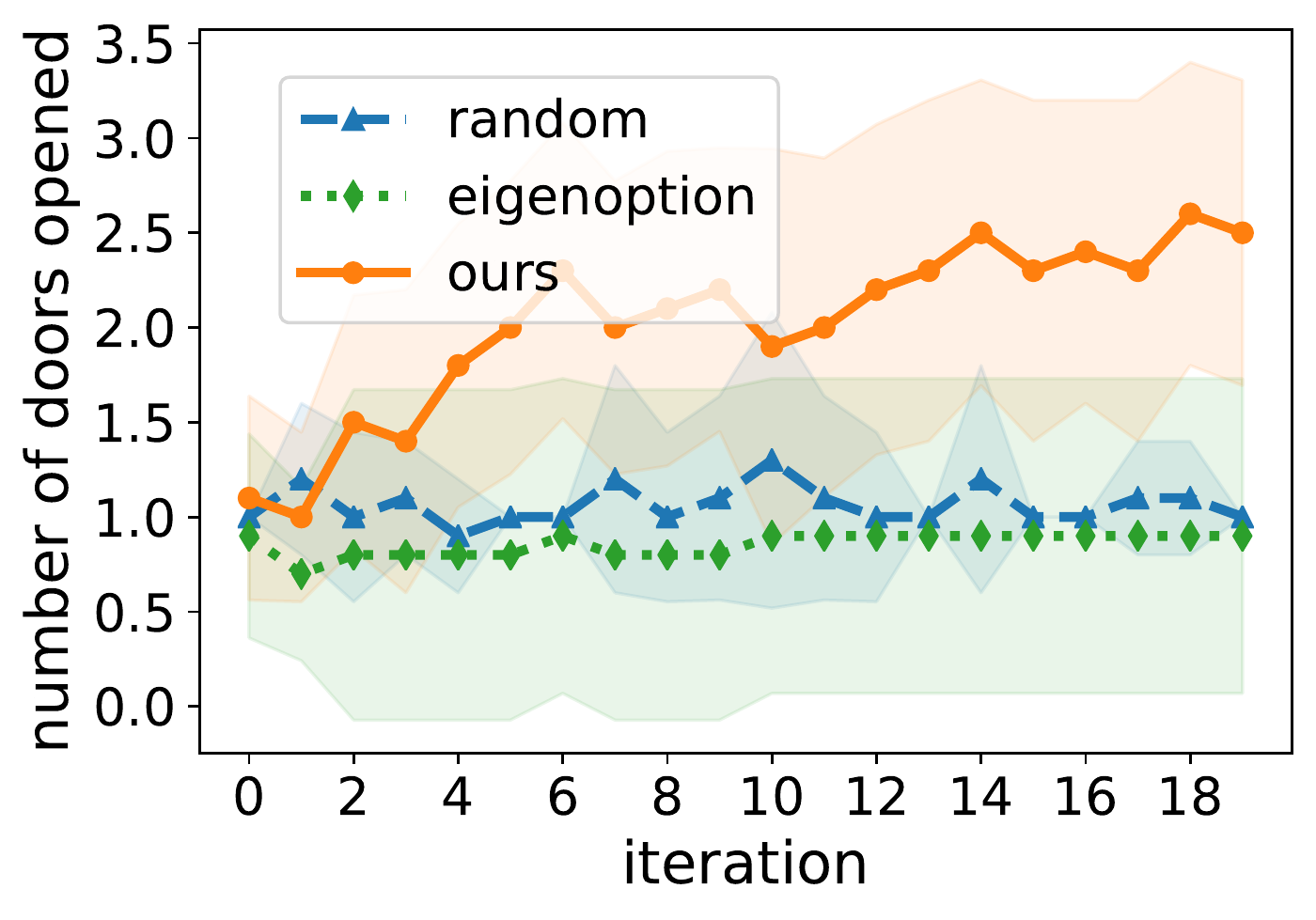}}%
	\hfill
	\subfloat[distance from start]{\includegraphics[width=0.23\linewidth, height=0.22\linewidth]{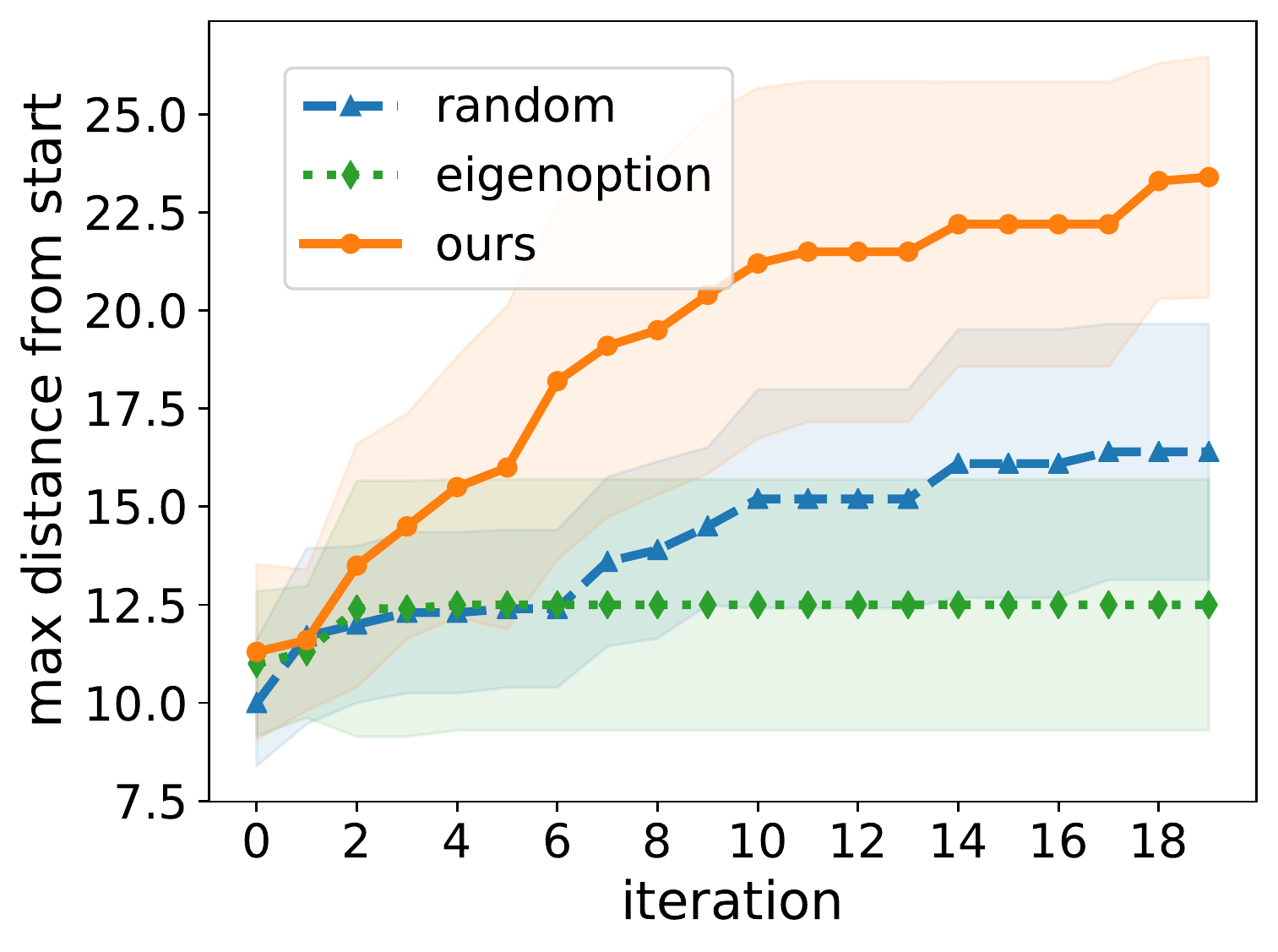}}%
	\hfill
	\caption{(Minecraft Door-Rooms) Exploring and grounding options in unknown environments. 4 rooms are connected by doors, the agent starts from the bottom, opens doors and enters the other rooms. The random agent takes random actions. Our agent starts with an iteration of random walk, then after each iteration, it computes the constructed MDP and fits the "go to and open door" option, (while the eigenoptions agent finds an eigenoption), and uses the options for exploration in the next iteration. (a) shows the state visitation frequency in iteration 20 of our agent (Algorithm~\ref{algo:lifting_batch}), and the MDP constructed throughout the 20 iterations. (b)-(d) compares our agent with the baselines on the average number of the explored states, doors opened and max distance traversed. The results are averaged over 10 seeds and shaded regions show the standard deviation across seeds. Our agent using abstract successor options explores on average twice as many states as the baselines, as well as quickly learns to open doors and navigating to new rooms.}
	\label{fig:minecraft_open_door}
	\vspace{-5mm}
\end{figure*}

\textbf{Minecraft Door-Rooms Experiments:}
As introduced in Section~\ref{sec:experiments}, to test our batched option grounding algorithm (Algorithm~\ref{algo:lifting_batch}) on new environments with unknown transition dynamics, we built two settings in the Malmo Minecraft environment: \emph{Bake-Rooms} and \emph{Door-Rooms}. The results on Bake-Rooms can be found in Figure~\ref{fig:minecraft_potato} in the main text. Here, we present the results on the Door-Rooms setting shown in Figure~\ref{fig:minecraft_open_door}.

\emph{Training and Results: } We compare our algorithm with the following two baselines: eigenoptions~\cite{machado2017eigenoption} and random walk. For this experiment, each agent runs for 20 iterations, with 200 steps per iteration as follows: In the first iteration, all agents execute randomly chosen actions. After each iteration, the agents construct an MDP graph based on collected transitions from all prior iterations. The eigenoption agent computes $k=1$ eigenoption of the second smallest eigenvalue (Fiedler vector) using the normalized graph Laplacian, while our algorithm grounds the $k=1$ abstract option: \emph{open door and go to door}. In the next iteration, the agents perform random walks with both the primitive actions and the acquired options, update the MDP graphs, compute new options, $\ldots$

Figure~\ref{fig:minecraft_open_door} shows our obtained results. Figure \ref{fig:minecraft_open_door}(a) shows the state visitation frequencies of our algorithm in the 20th iteration and the constructed MDP graph. The agent starts from the bottom room (R1), and learns to navigate towards the door, open the door and enter the next rooms. 
Figures \ref{fig:minecraft_open_door}(b)-(d) compare the agents in terms of the total number of states explored, number of doors opened and the maximum distance from the starting location. Note that the door layouts are different from the Bake-Rooms environment. Our agent explores on average more than twice as many states as the two baselines, quickly learns to open the doors and navigate to new rooms, while the baselines on average only learn to open the first door and mostly stay in the starting room.

\begin{figure*}
    \centering
	\subfloat[iteration 0]{\includegraphics[width=0.25\linewidth]{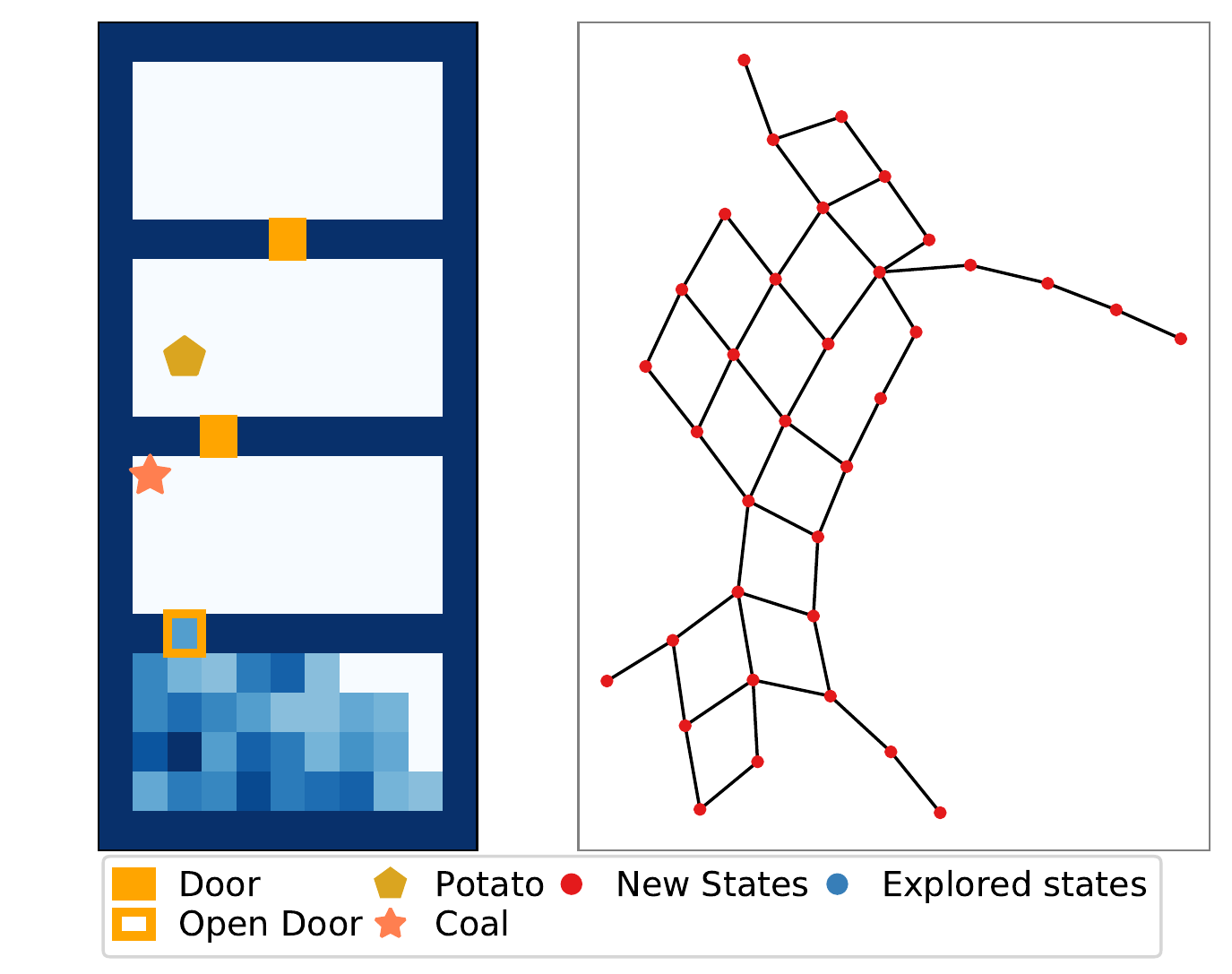}}%
	\hfill
	\subfloat[iteration 1]{\includegraphics[width=0.25\linewidth]{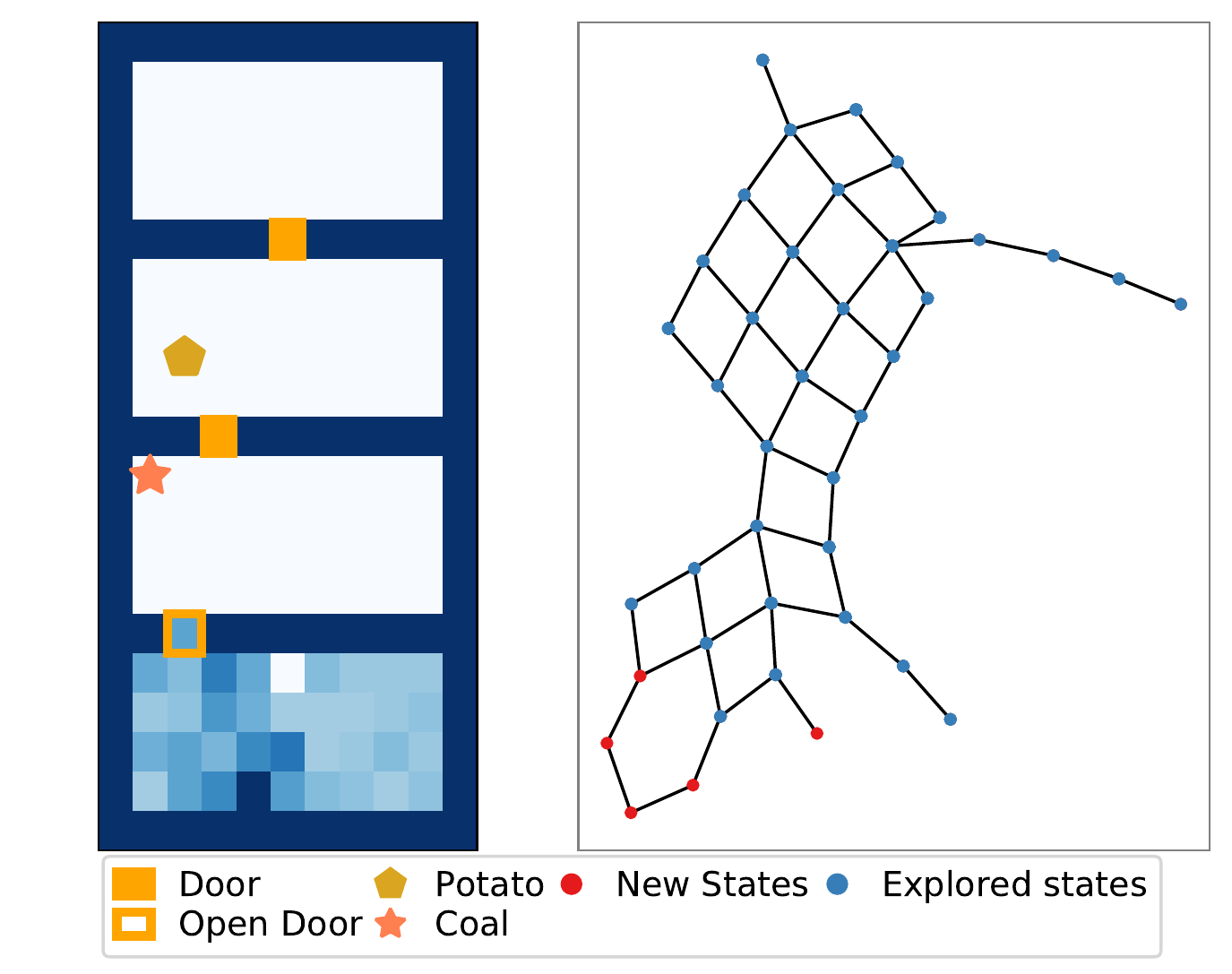}}%
	\hfill
	\subfloat[iteration 2]{\includegraphics[width=0.25\linewidth]{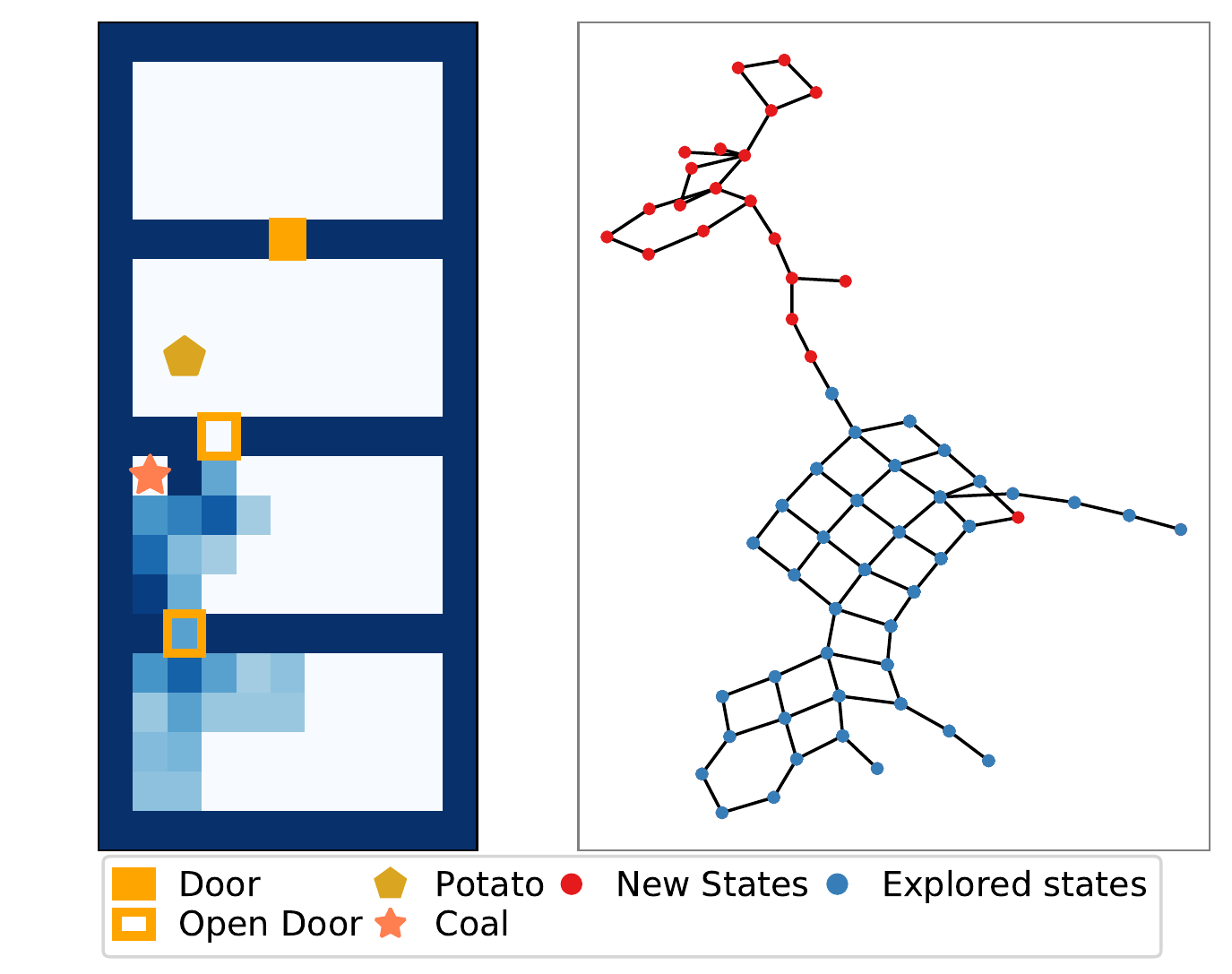}}%
	\hfill
	\subfloat[iteration 3]{\includegraphics[width=0.25\linewidth]{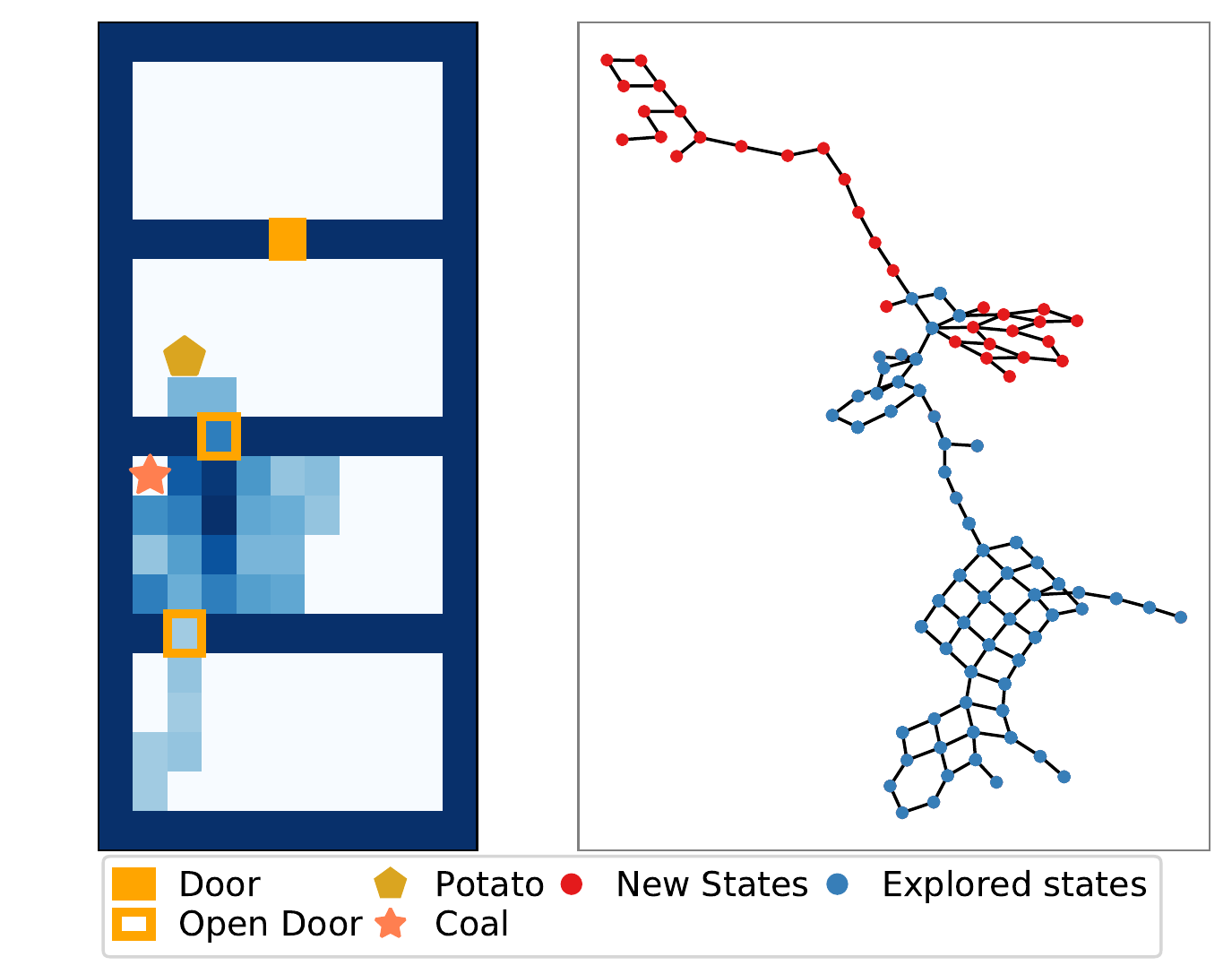}}%
	\vfill
	\subfloat[iteration 4]{\includegraphics[width=0.25\linewidth]{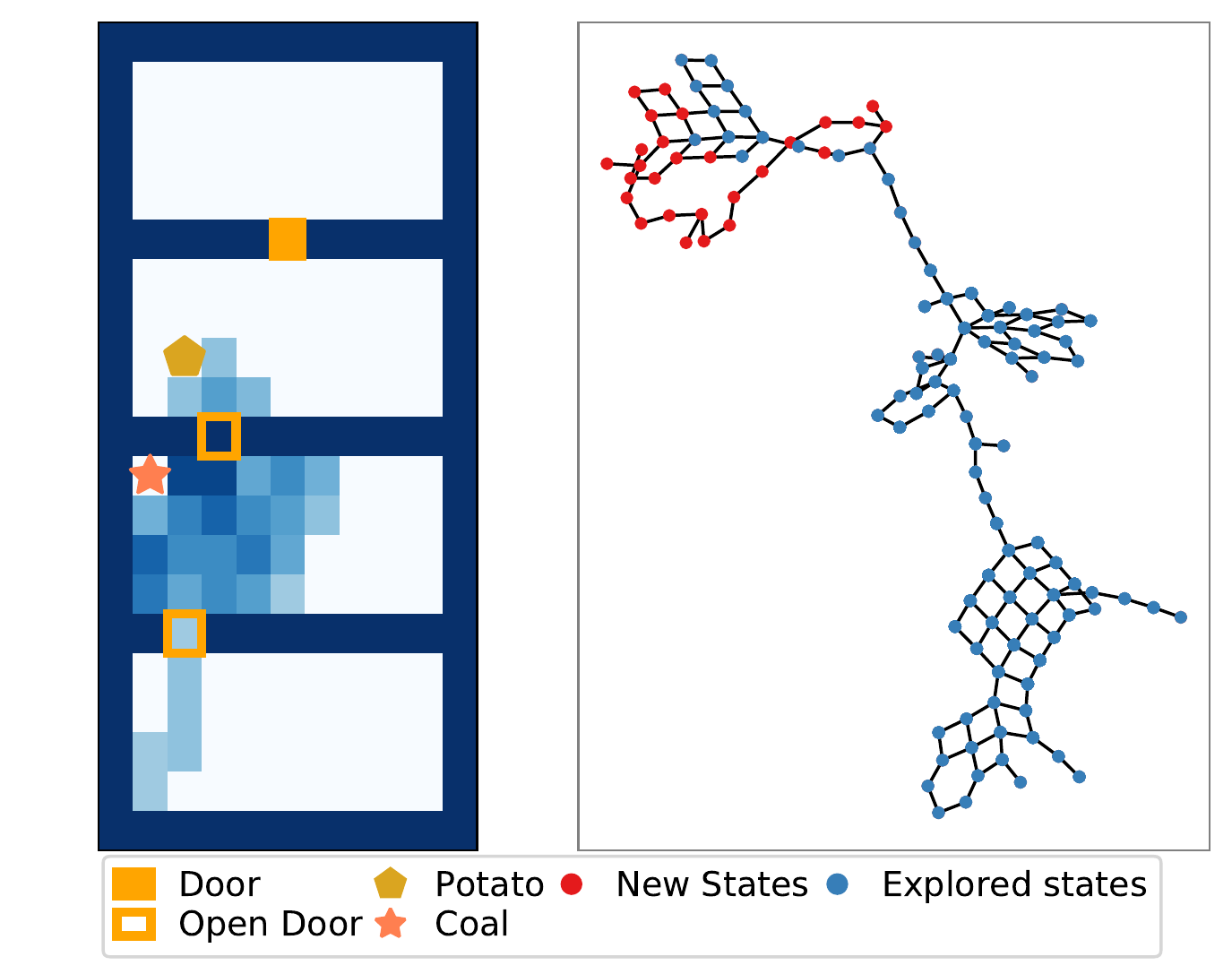}}%
	\hfill
	\subfloat[iteration 5]{\includegraphics[width=0.25\linewidth]{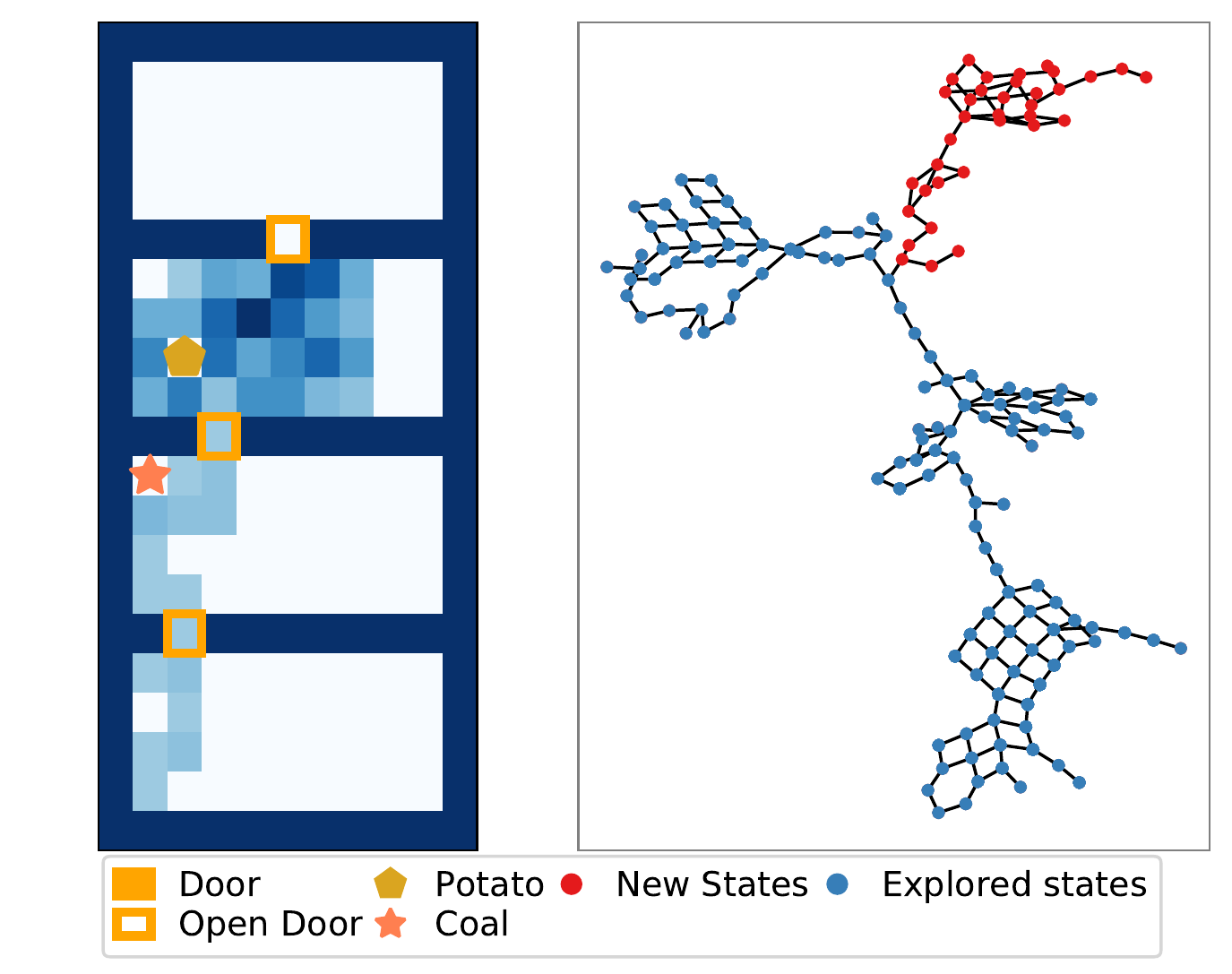}}%
	\hfill
	\subfloat[iteration 6]{\includegraphics[width=0.25\linewidth]{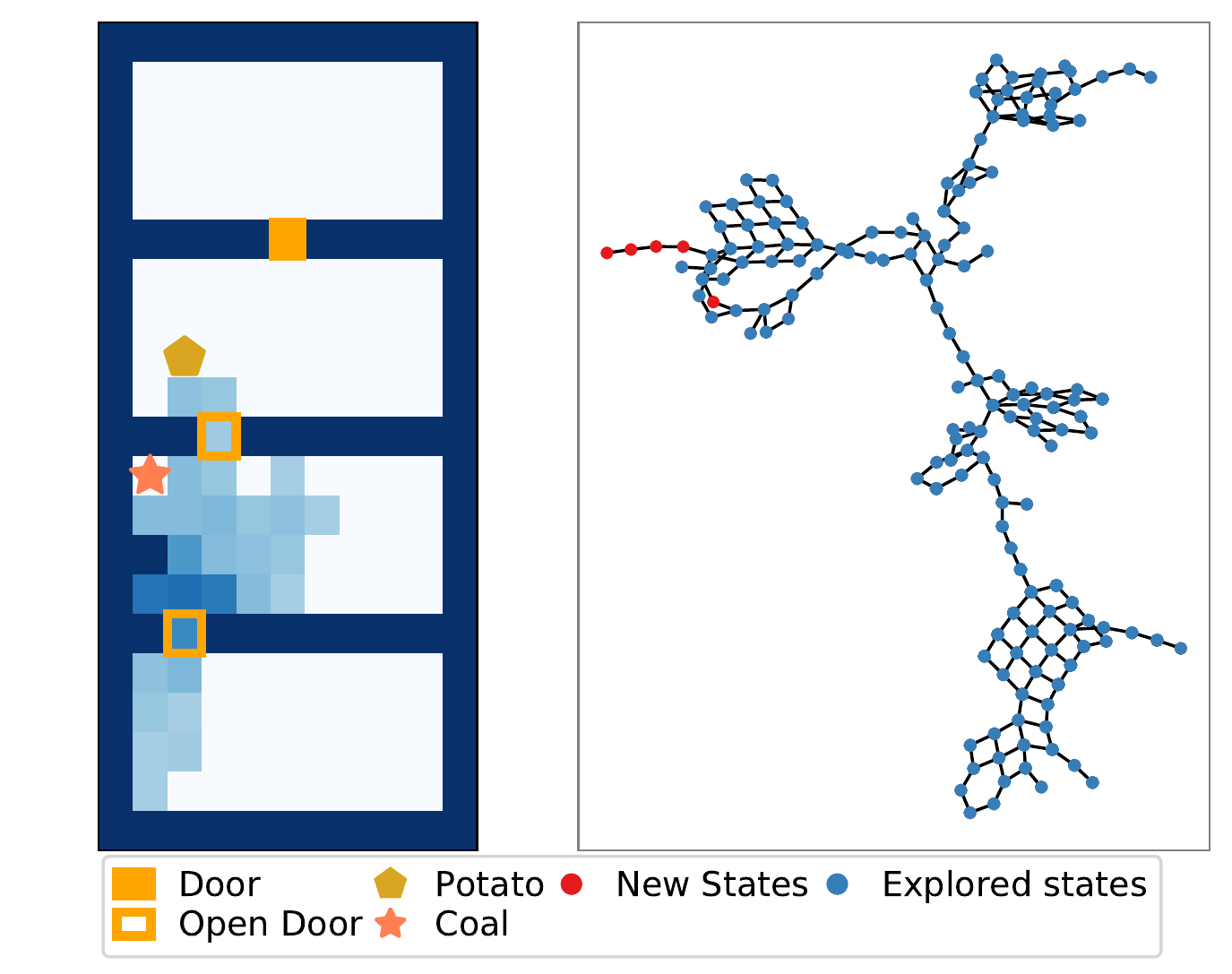}}%
	\hfill
	\subfloat[iteration 7]{\includegraphics[width=0.25\linewidth]{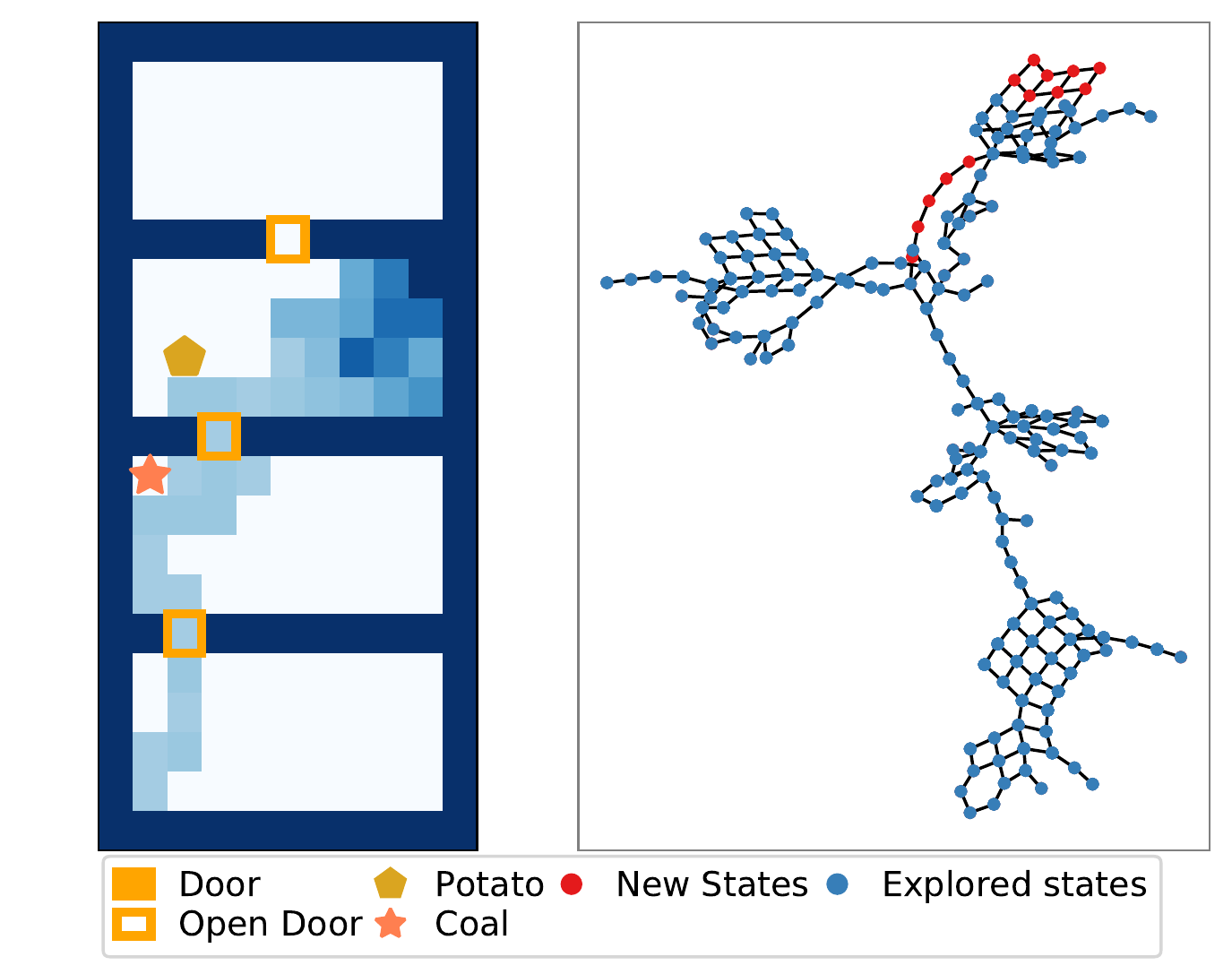}}%
	\vfill
	\subfloat[iteration 8]{\includegraphics[width=0.25\linewidth]{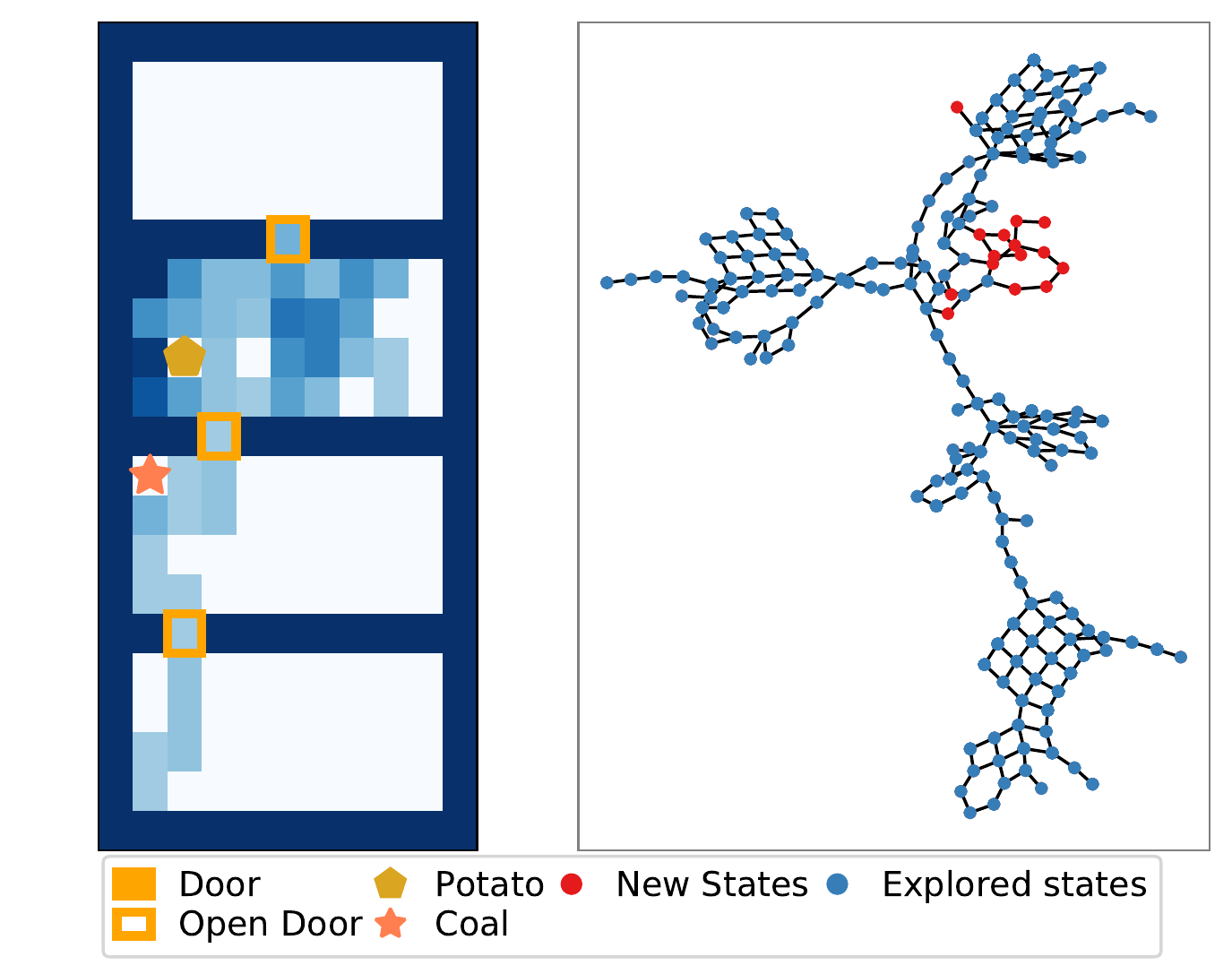}}%
	\hfill
	\subfloat[iteration 9]{\includegraphics[width=0.25\linewidth]{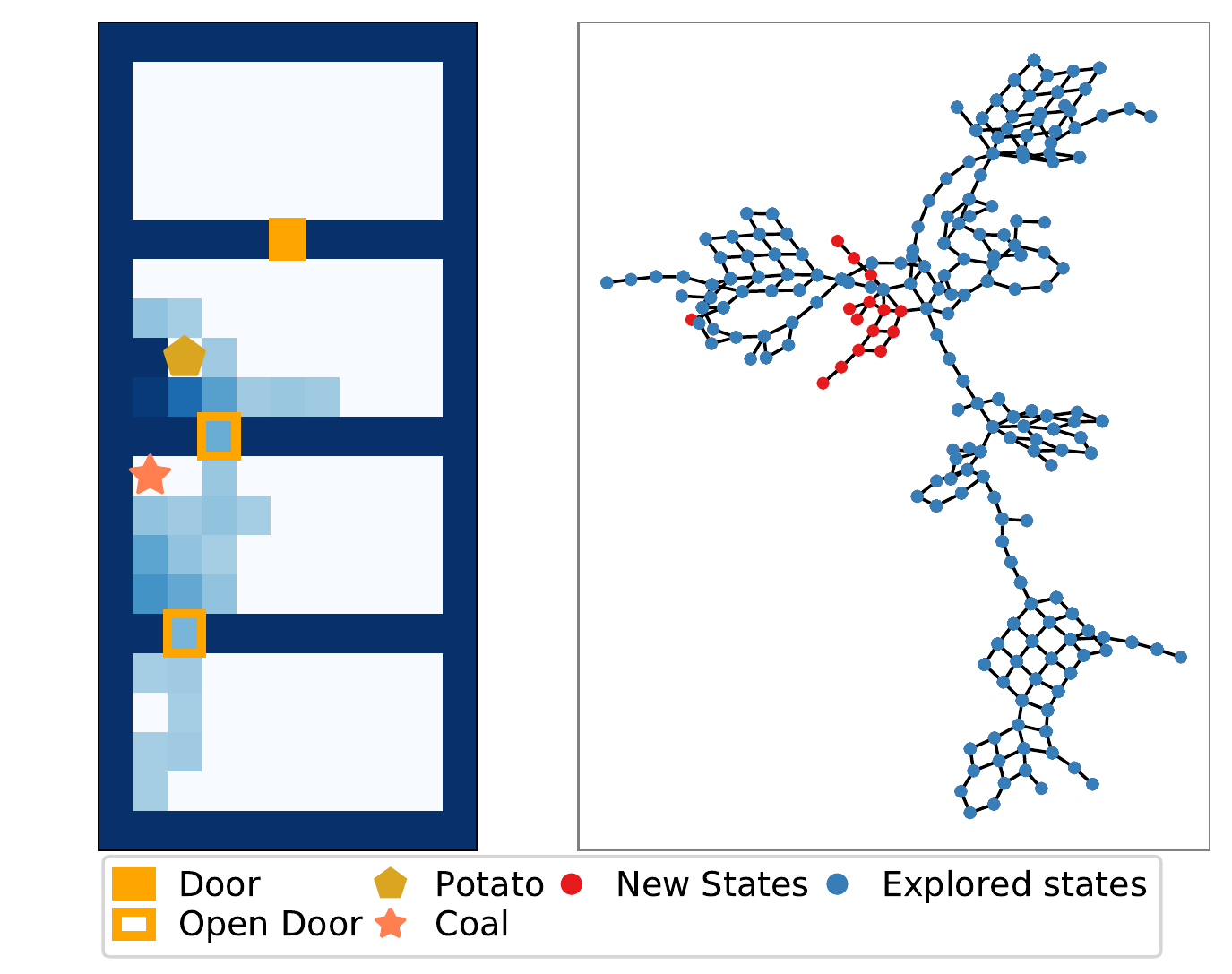}}%
	\hfill
	\subfloat[iteration 10]{\includegraphics[width=0.25\linewidth]{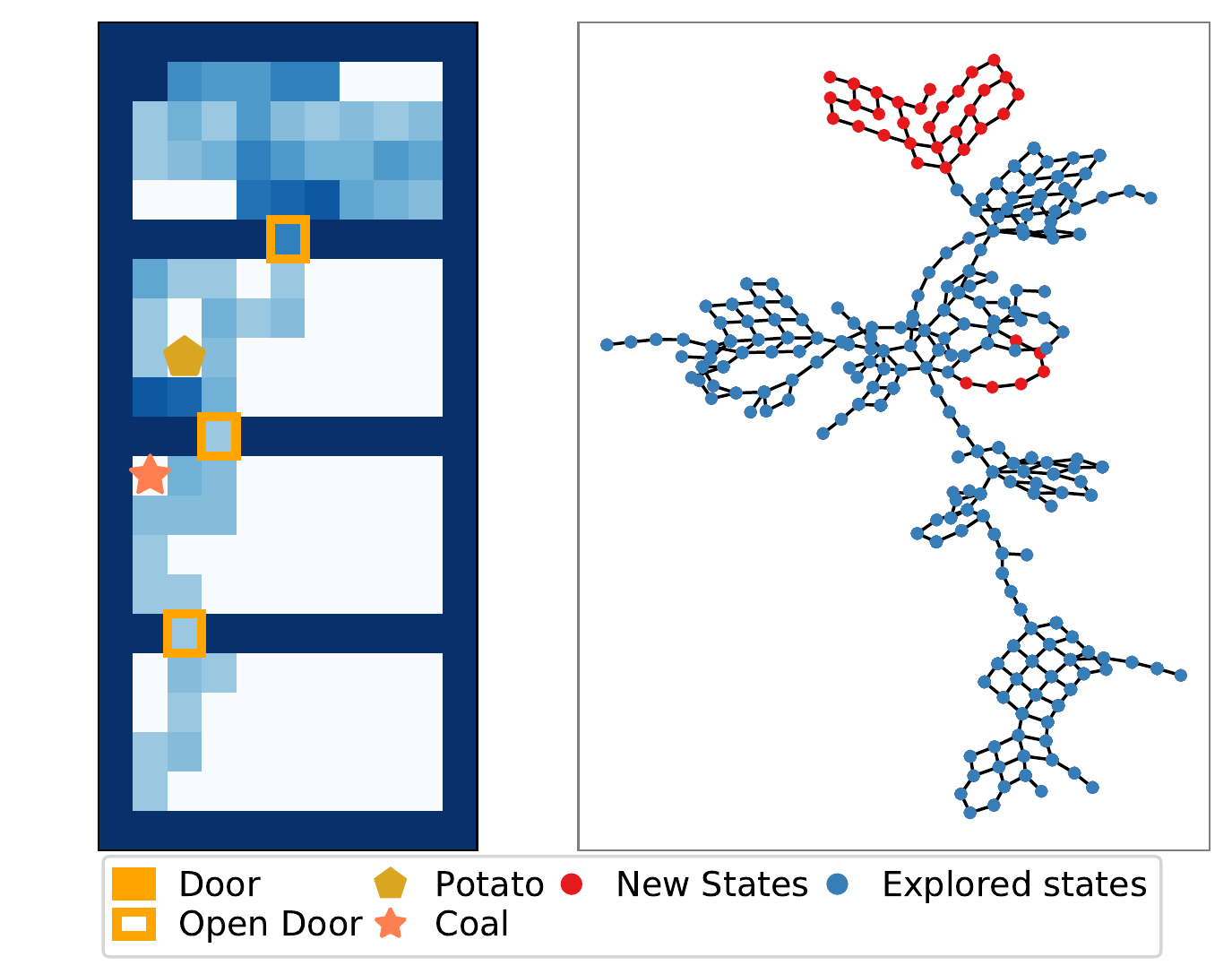}}%
	\hfill
	\subfloat[iteration 11]{\includegraphics[width=0.25\linewidth]{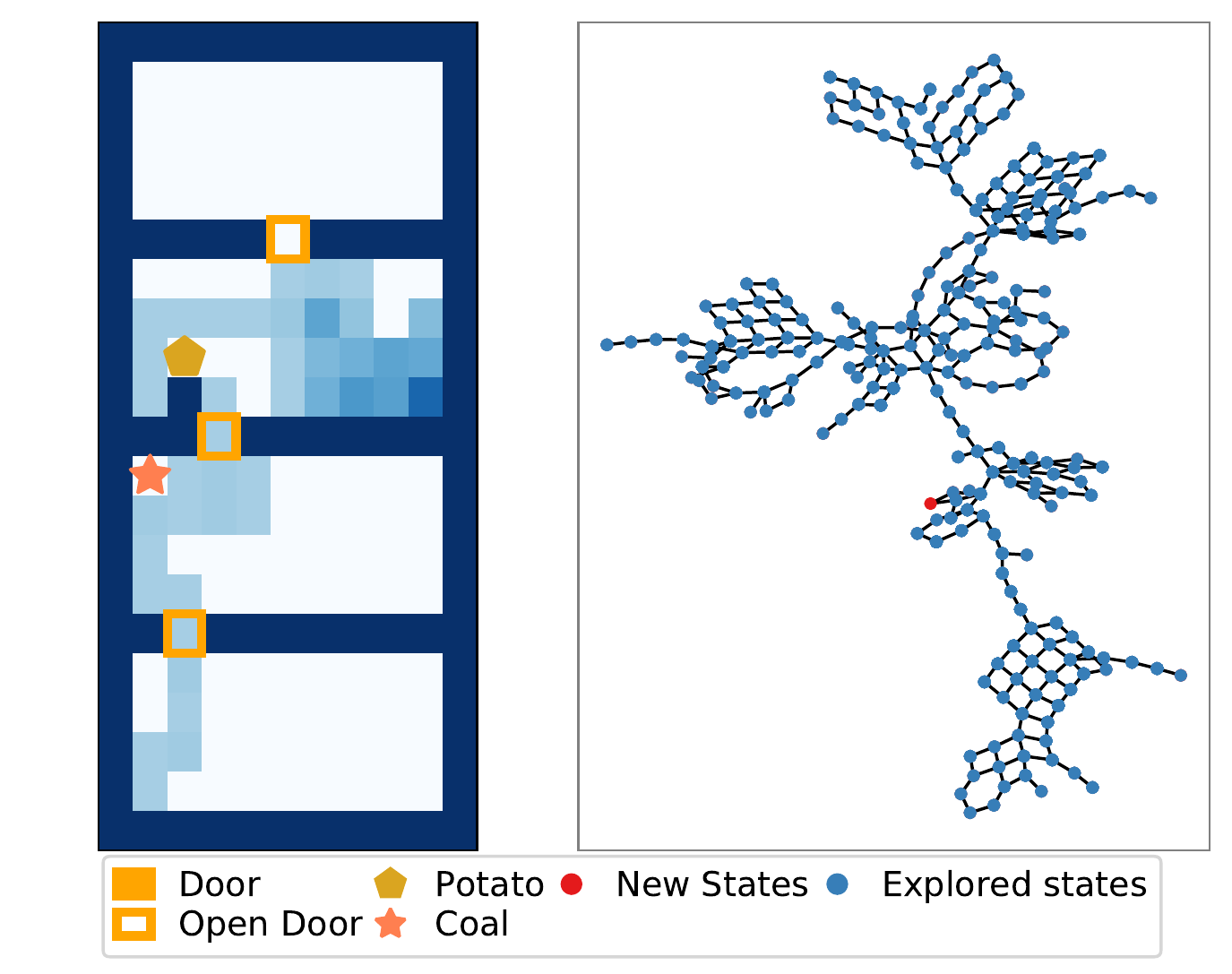}}%
	\vfill
	\subfloat[iteration 12]{\includegraphics[width=0.25\linewidth]{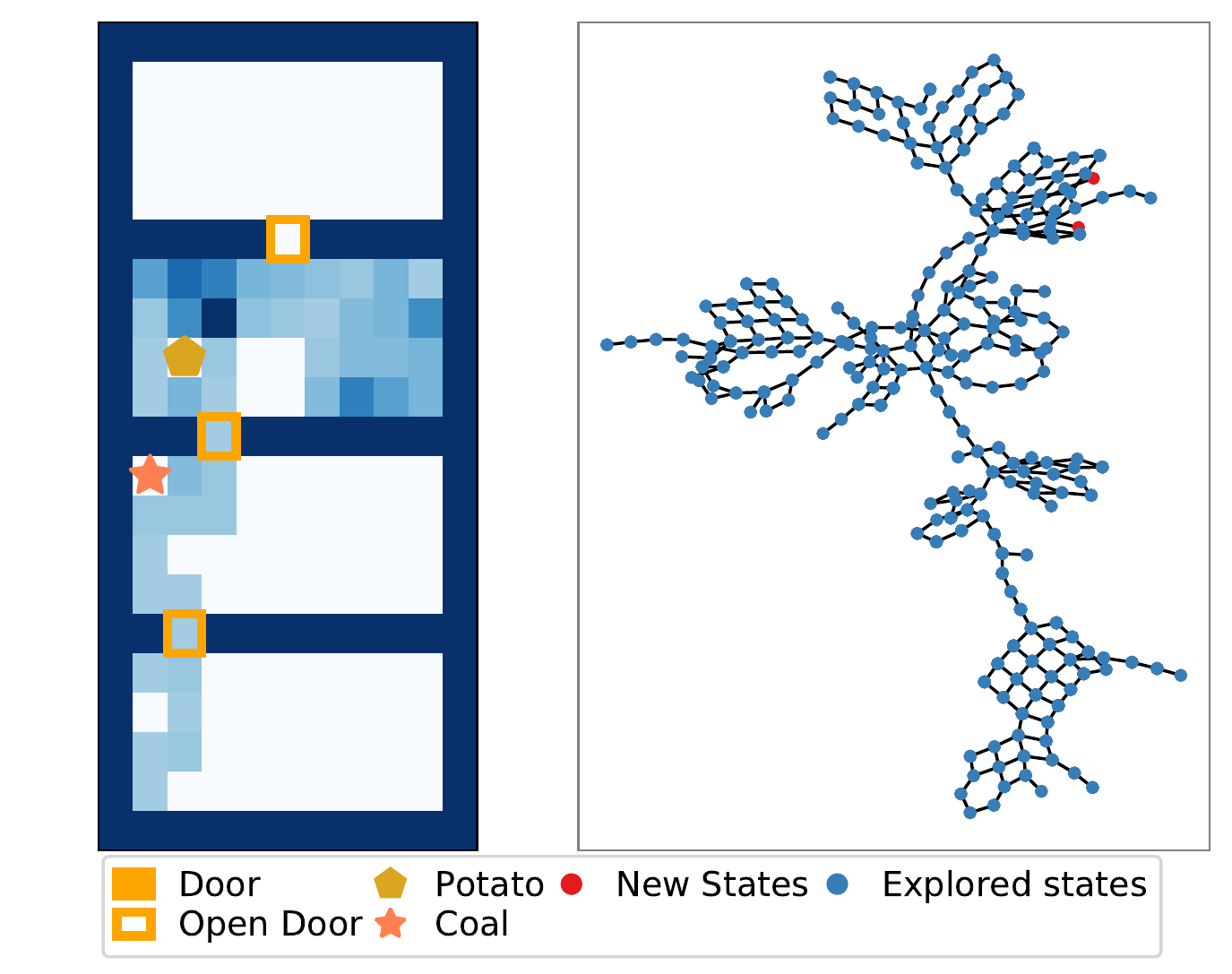}}%
	\hfill
	\subfloat[iteration 13]{\includegraphics[width=0.25\linewidth]{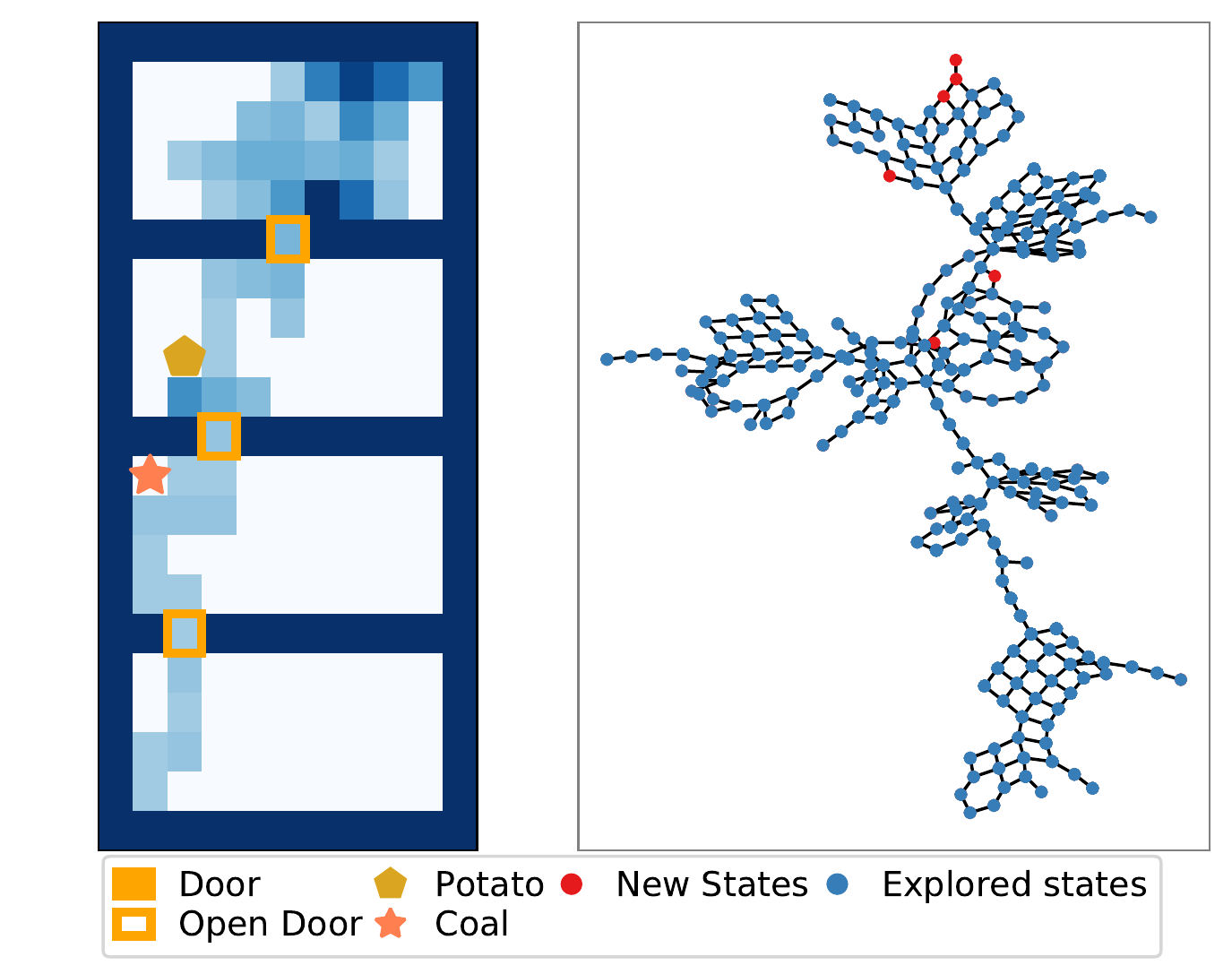}}%
	\hfill
	\subfloat[iteration 14]{\includegraphics[width=0.25\linewidth]{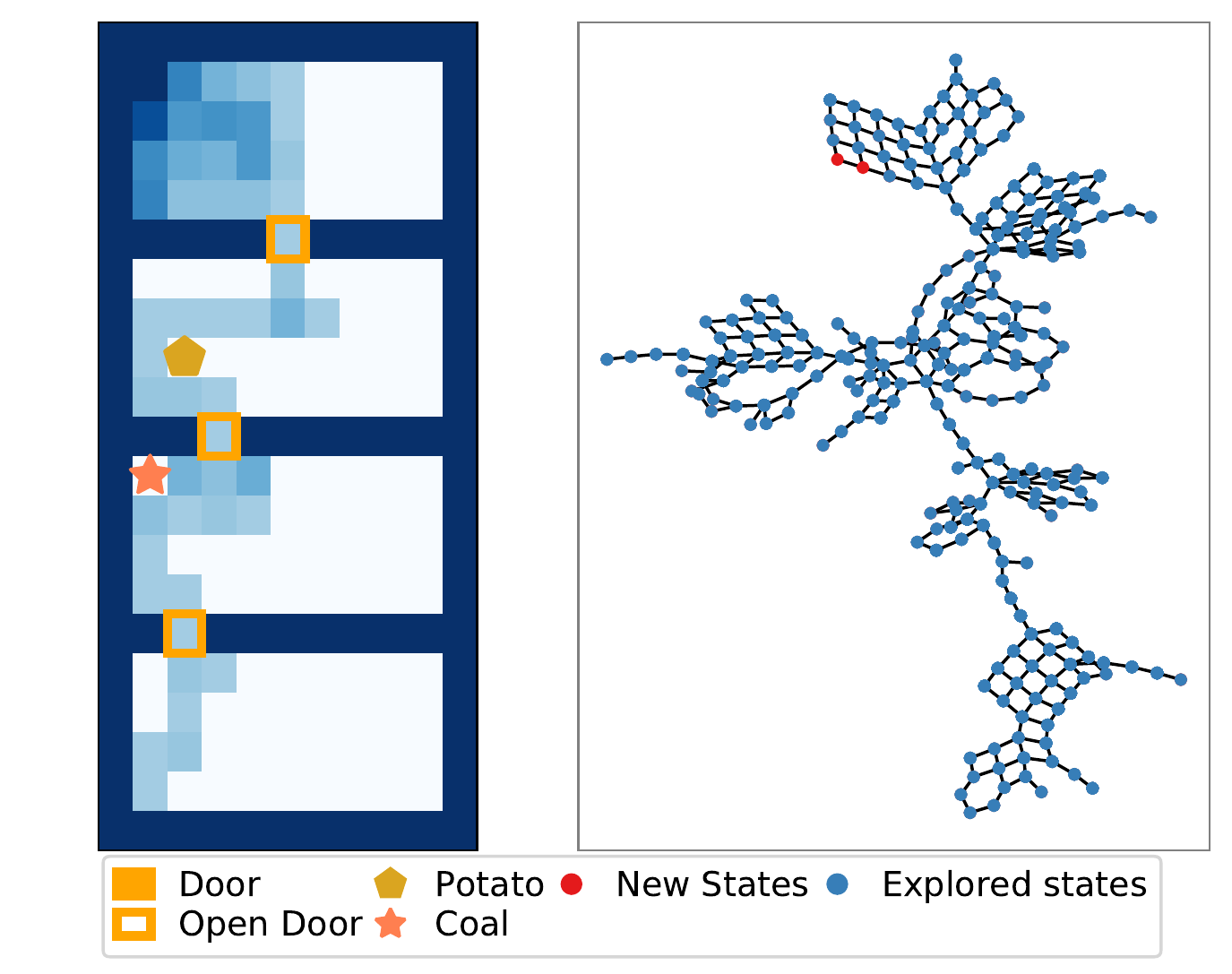}}%
	\hfill
	\subfloat[iteration 15]{\includegraphics[width=0.25\linewidth]{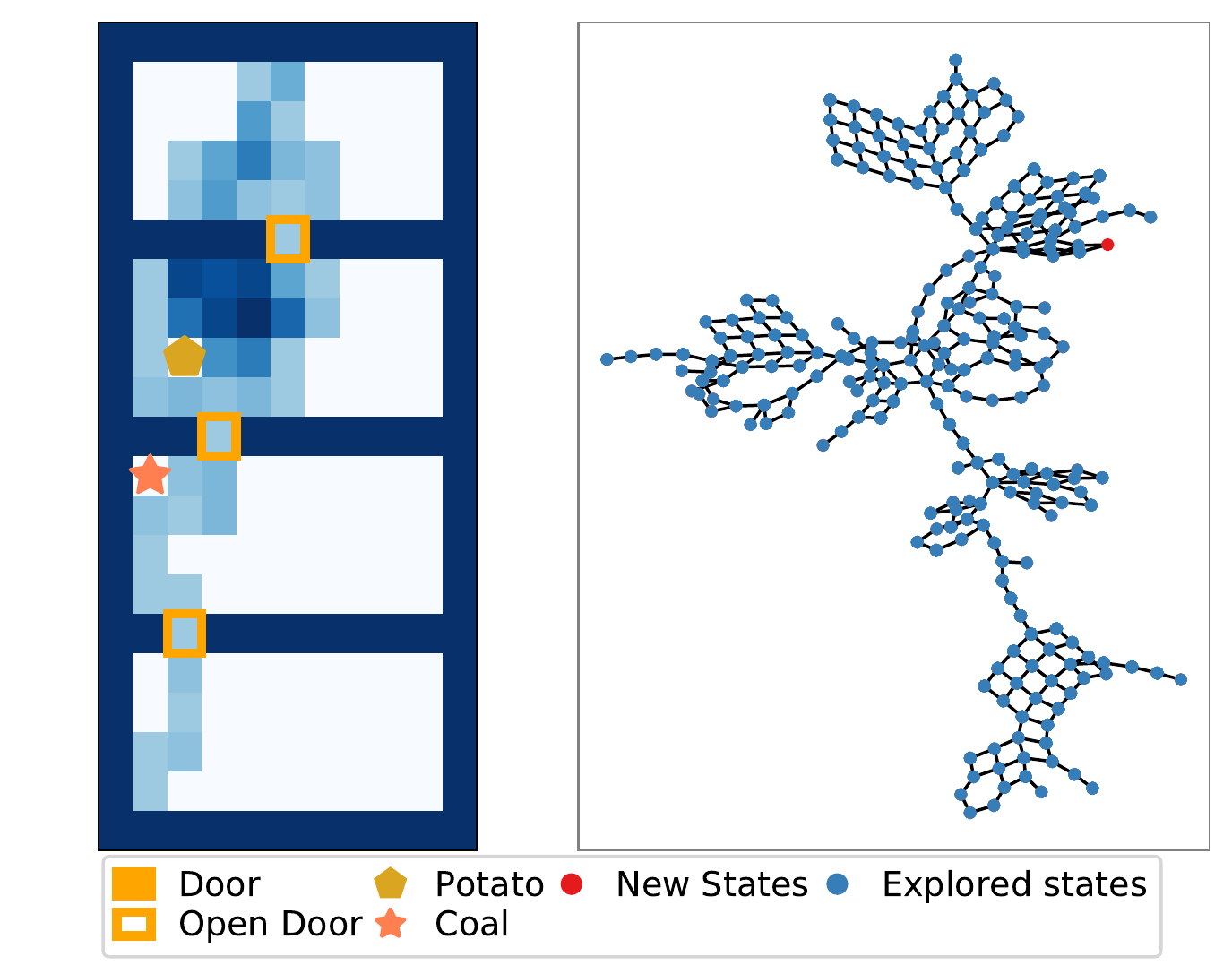}}%
	\vfill
	\subfloat[iteration 16]{\includegraphics[width=0.25\linewidth]{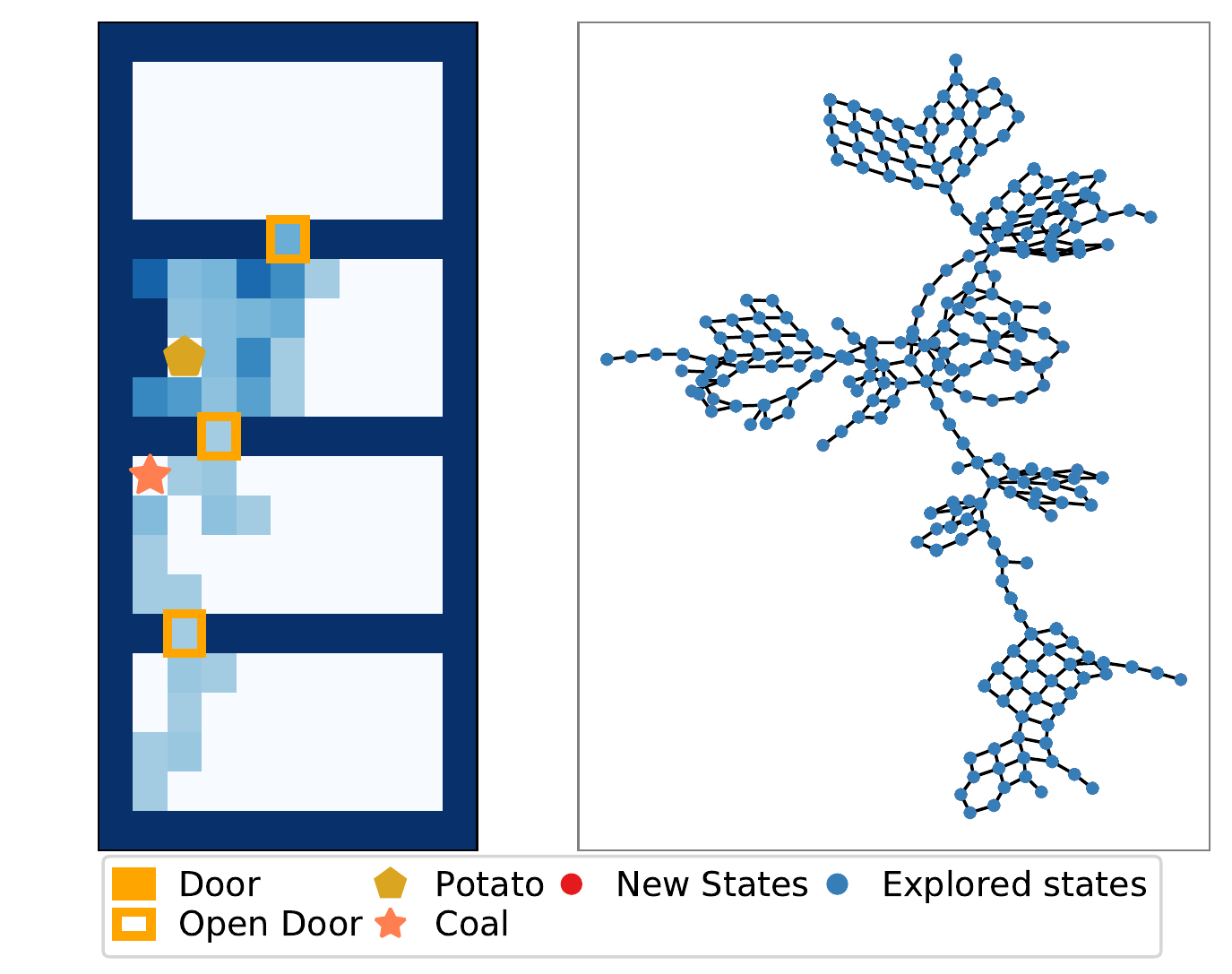}}%
	\hfill
	\subfloat[iteration 17]{\includegraphics[width=0.25\linewidth]{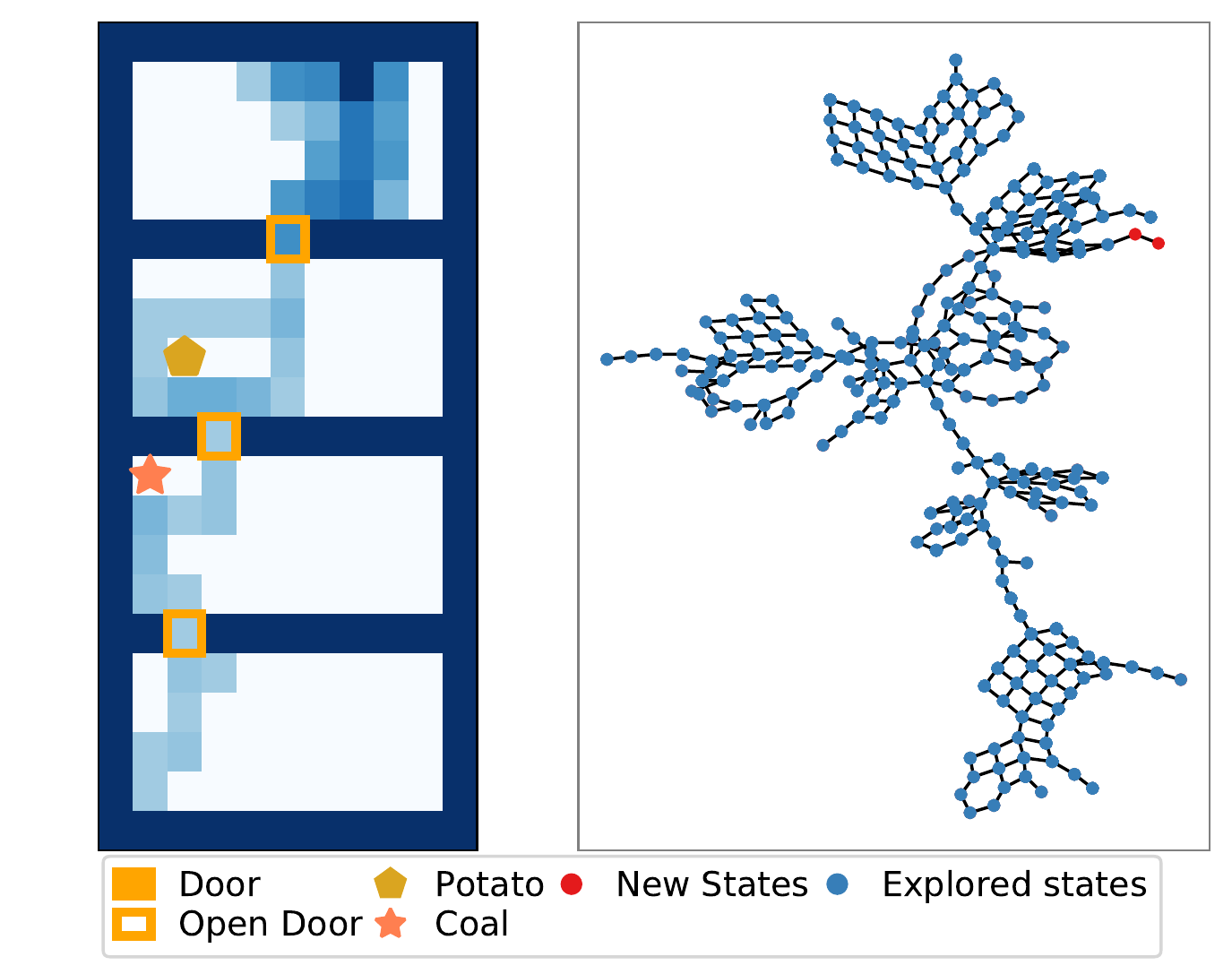}}%
	\hfill
	\subfloat[iteration 18]{\includegraphics[width=0.25\linewidth]{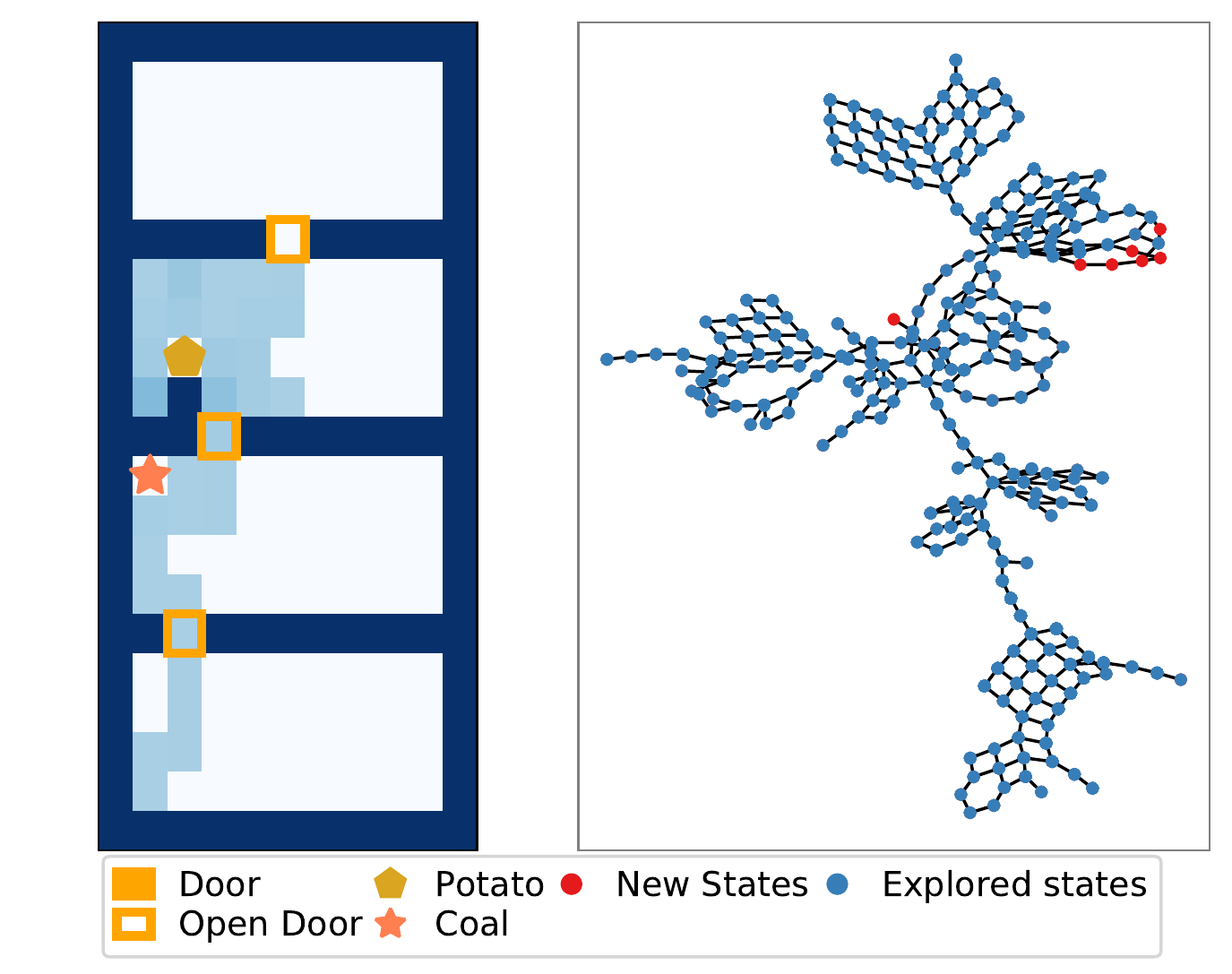}}%
	\hfill
	\subfloat[iteration 19]{\includegraphics[width=0.25\linewidth]{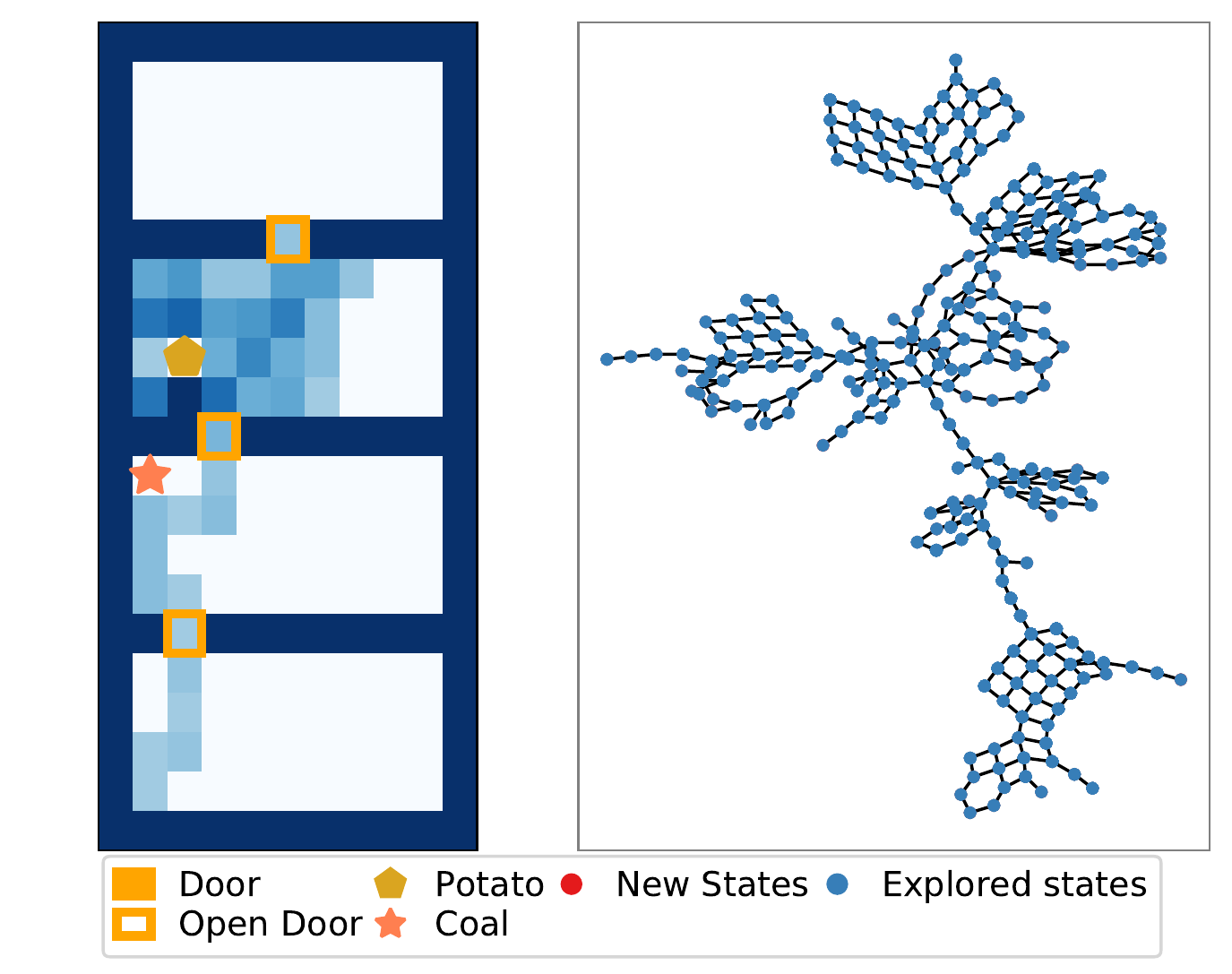}}%
	\hfill
    \caption{(Our Agent -  Algorithm~\ref{algo:lifting_batch}) State visitation frequencies and constructed graph per iteration in Minecraft Bake-Rooms}
    \label{fig:state-visitation-bake-rooms-ours}
\end{figure*}

\begin{figure*}
    \centering
	\subfloat[iteration 0]{\includegraphics[width=0.25\linewidth]{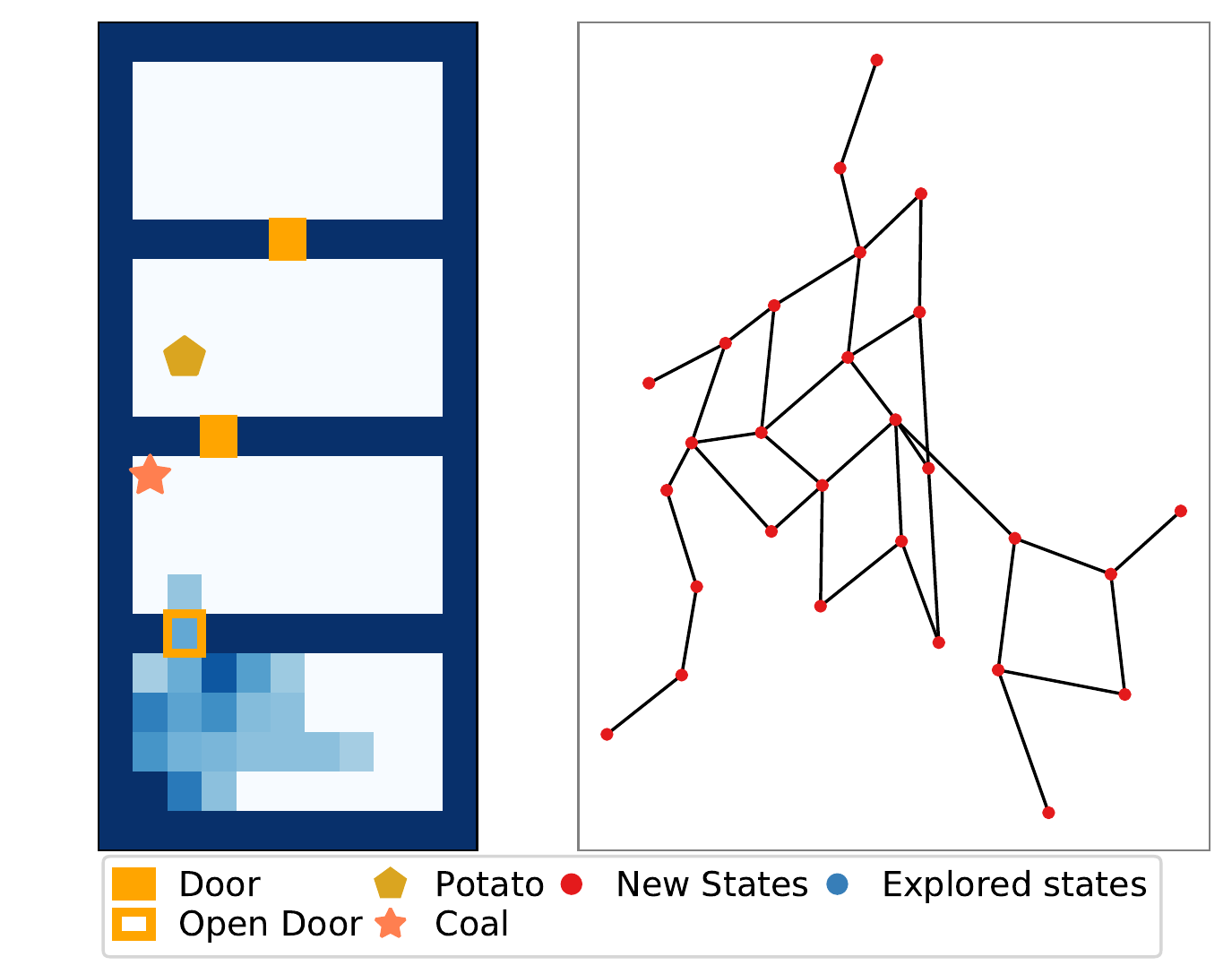}}%
	\hfill
	\subfloat[iteration 1]{\includegraphics[width=0.25\linewidth]{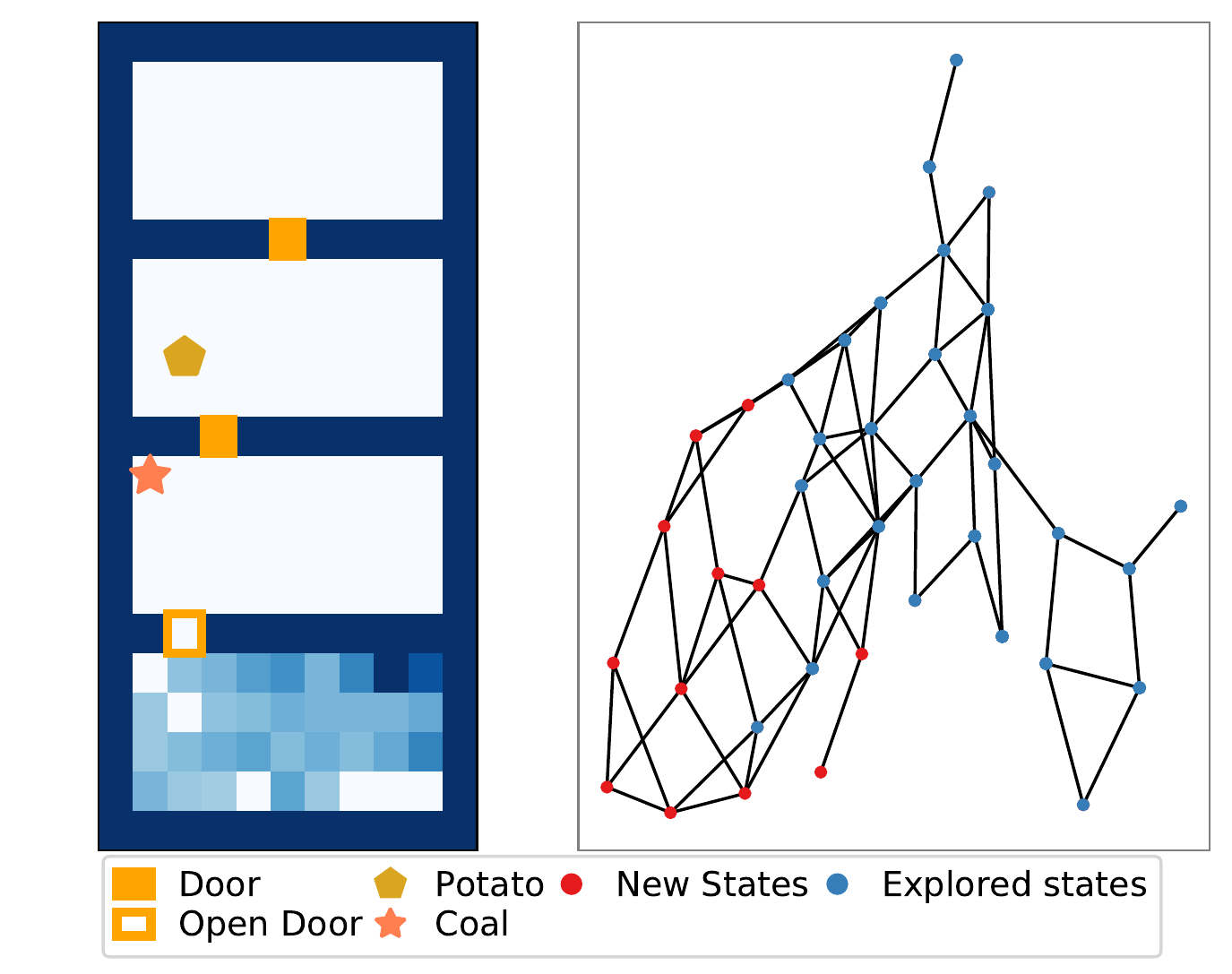}}%
	\hfill
	\subfloat[iteration 2]{\includegraphics[width=0.25\linewidth]{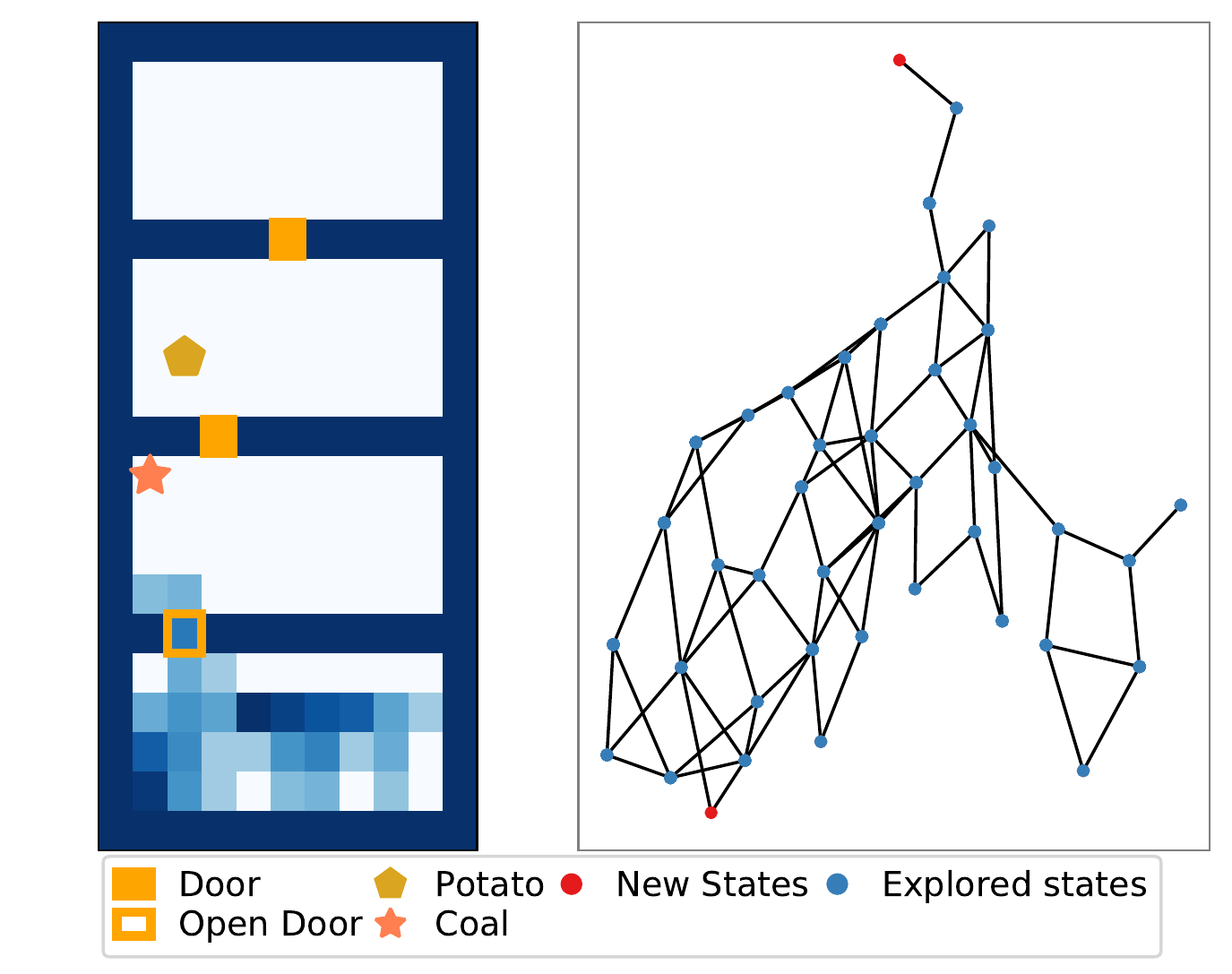}}%
	\hfill
	\subfloat[iteration 3]{\includegraphics[width=0.25\linewidth]{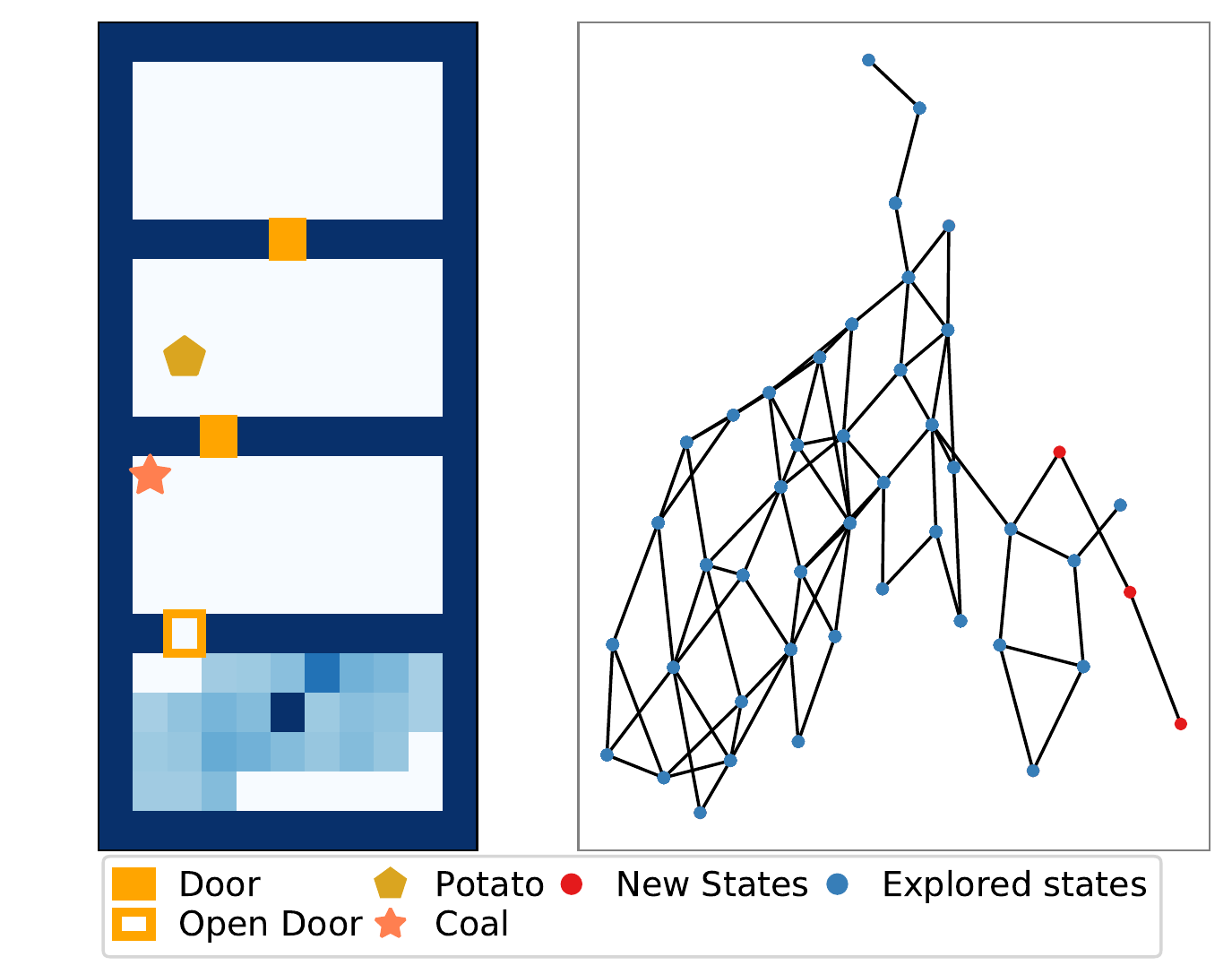}}%
	\vfill
	\subfloat[iteration 4]{\includegraphics[width=0.25\linewidth]{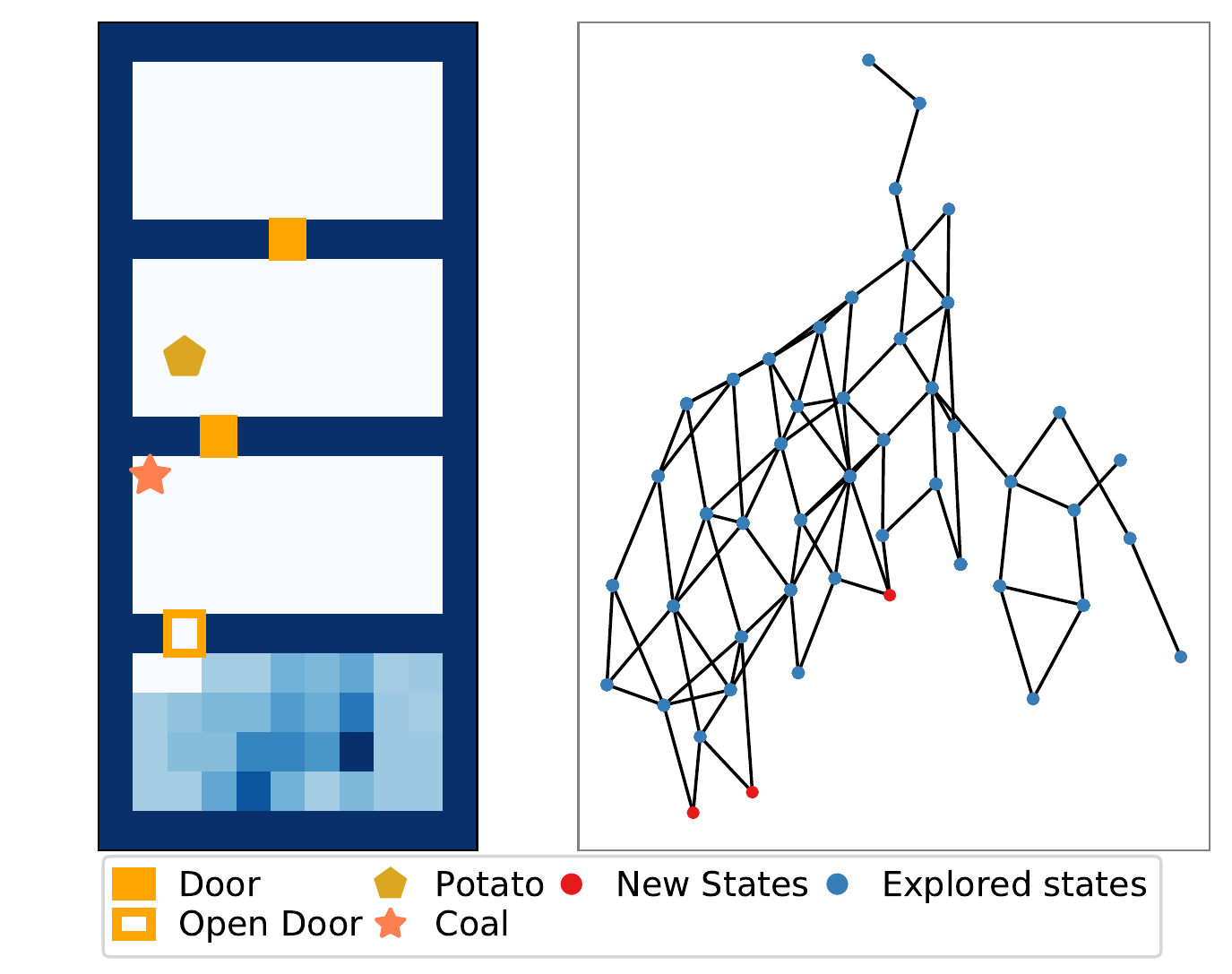}}%
	\hfill
	\subfloat[iteration 5]{\includegraphics[width=0.25\linewidth]{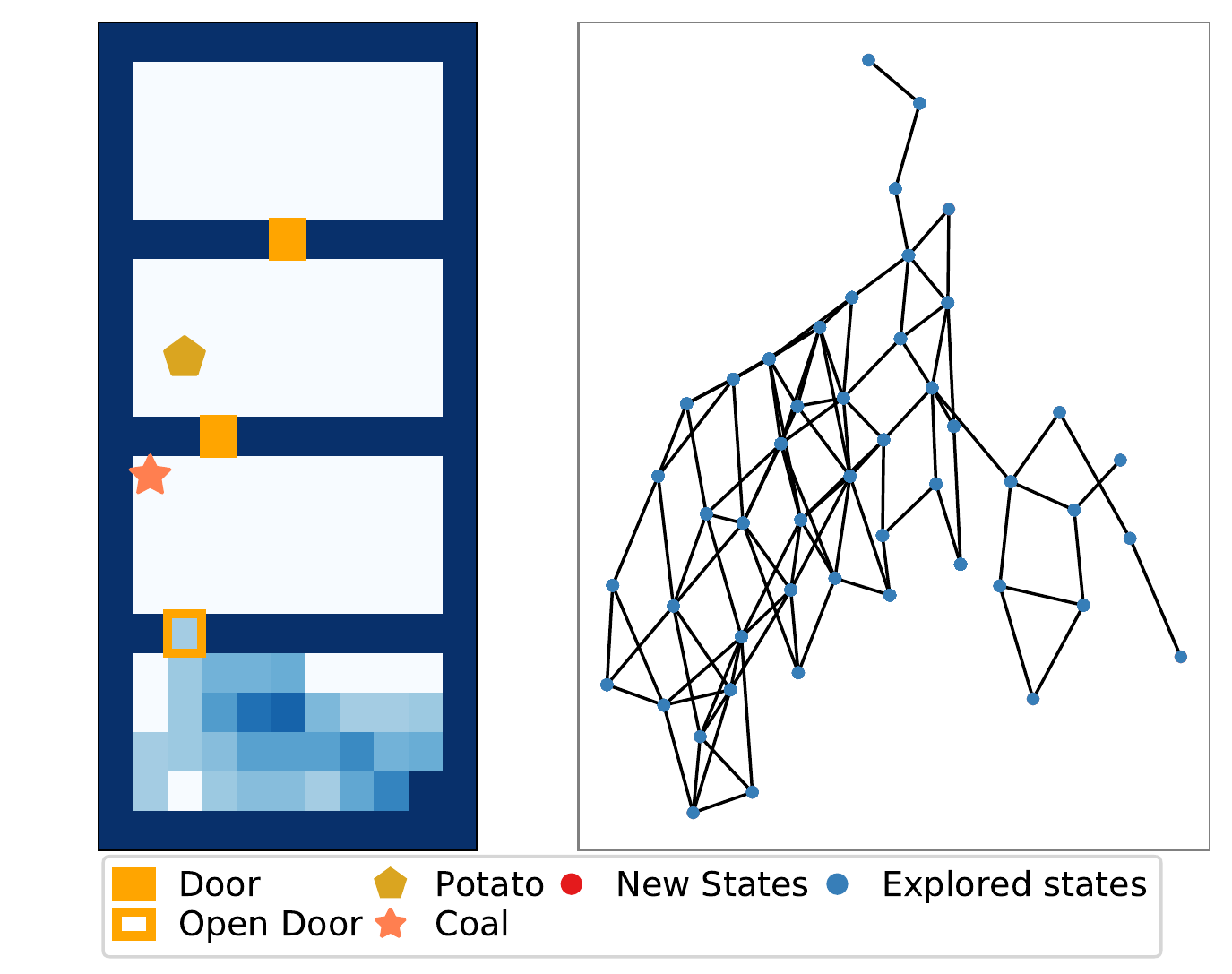}}%
	\hfill
	\subfloat[iteration 6]{\includegraphics[width=0.25\linewidth]{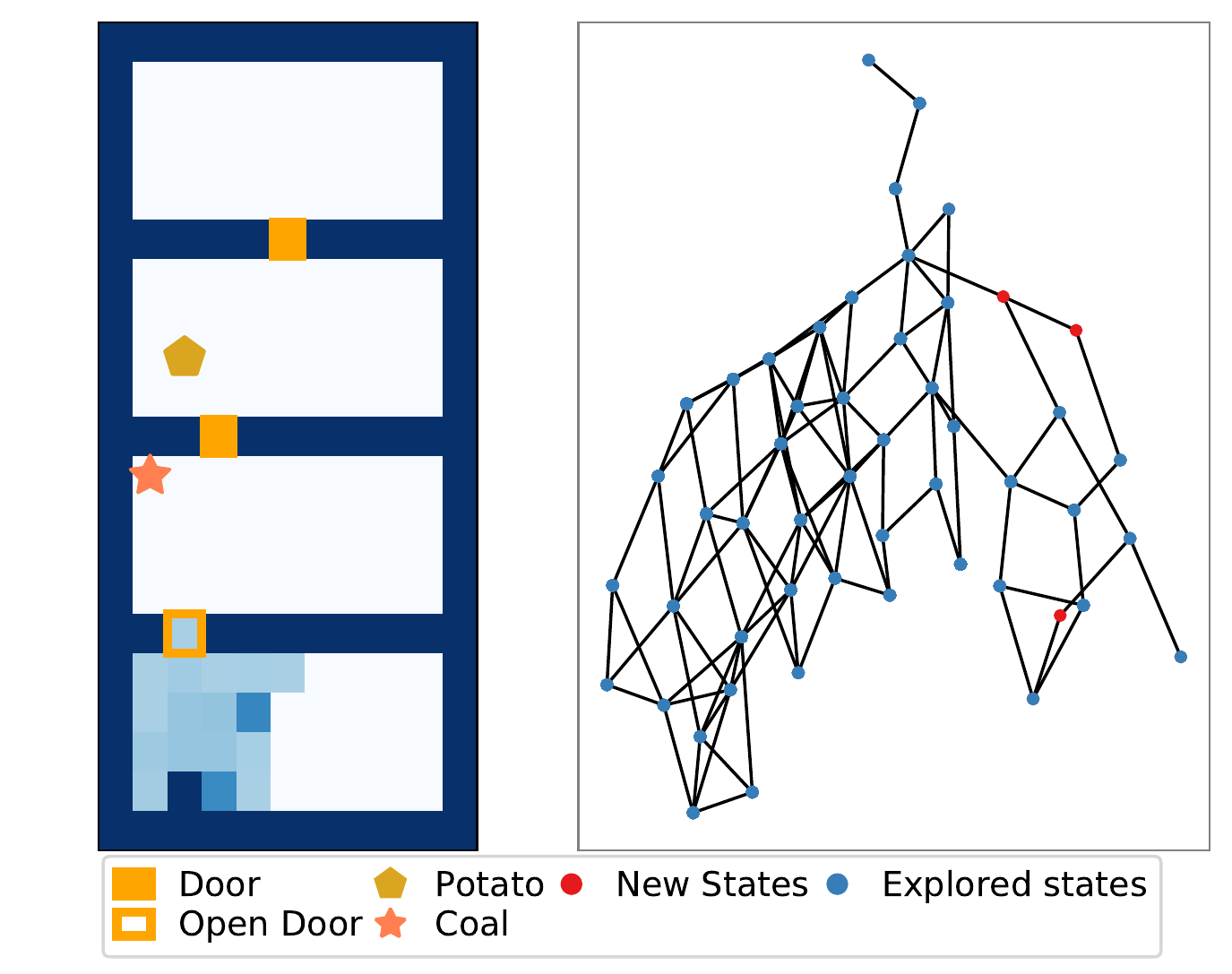}}%
	\hfill
	\subfloat[iteration 7]{\includegraphics[width=0.25\linewidth]{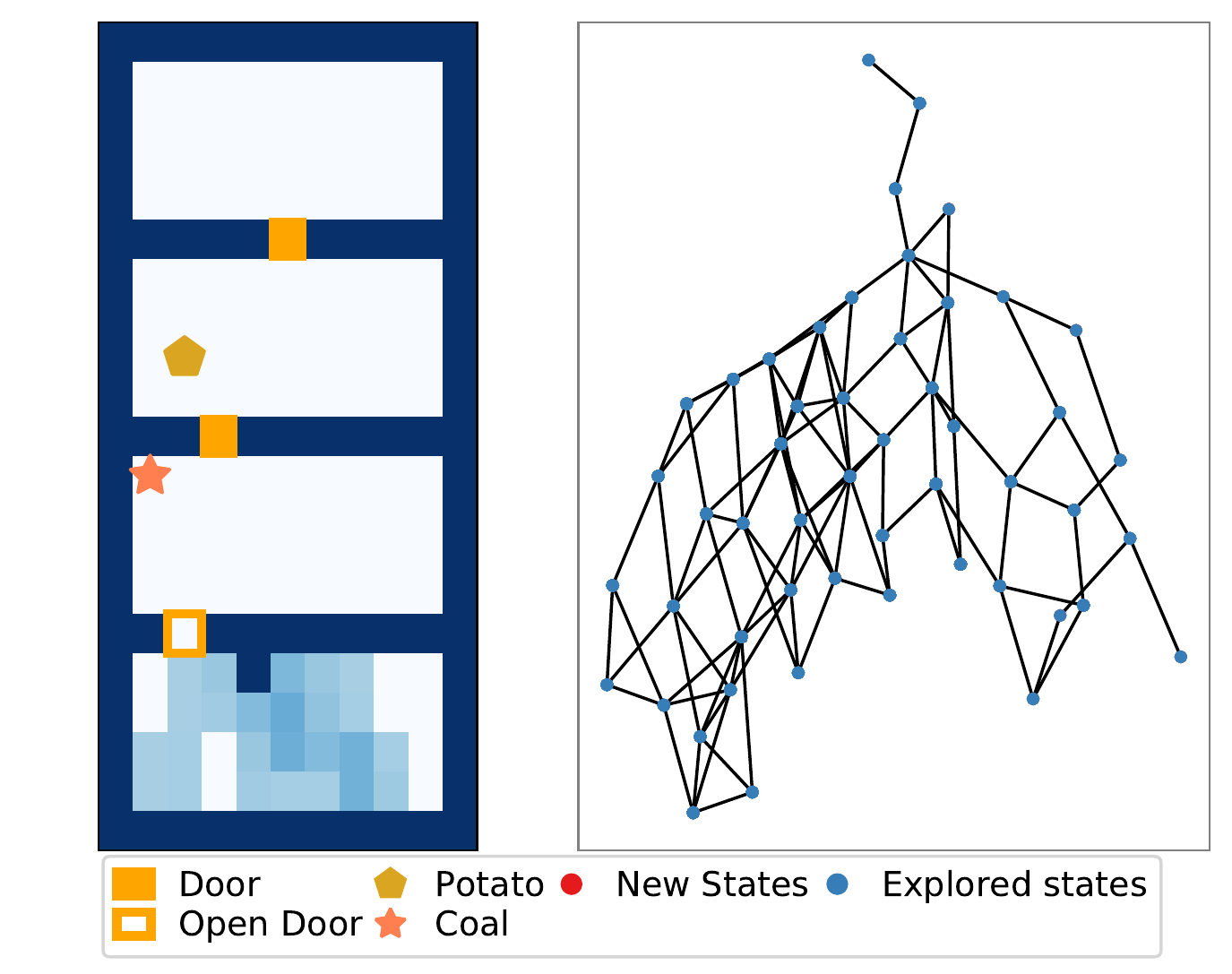}}%
	\vfill
	\subfloat[iteration 8]{\includegraphics[width=0.25\linewidth]{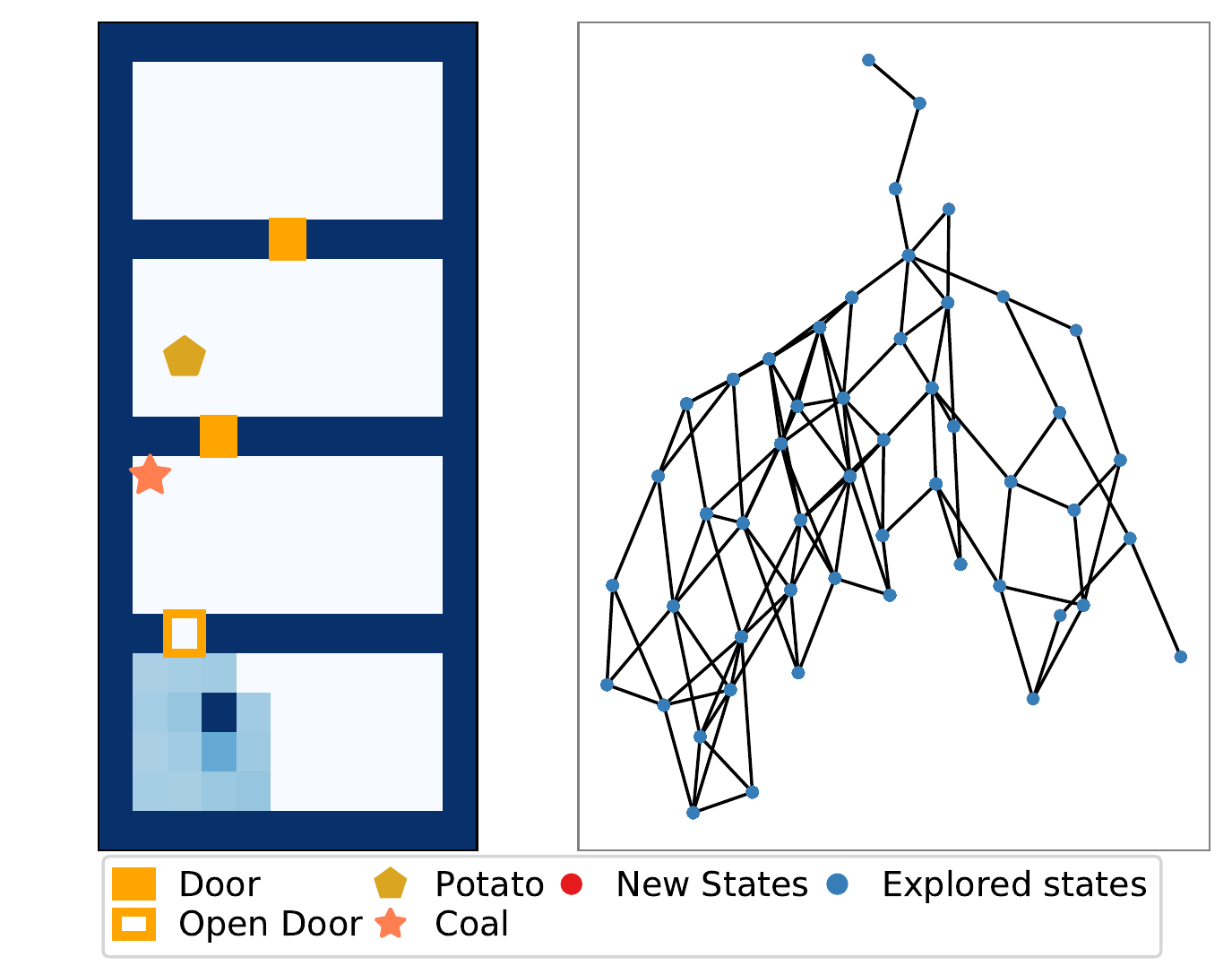}}%
	\hfill
	\subfloat[iteration 9]{\includegraphics[width=0.25\linewidth]{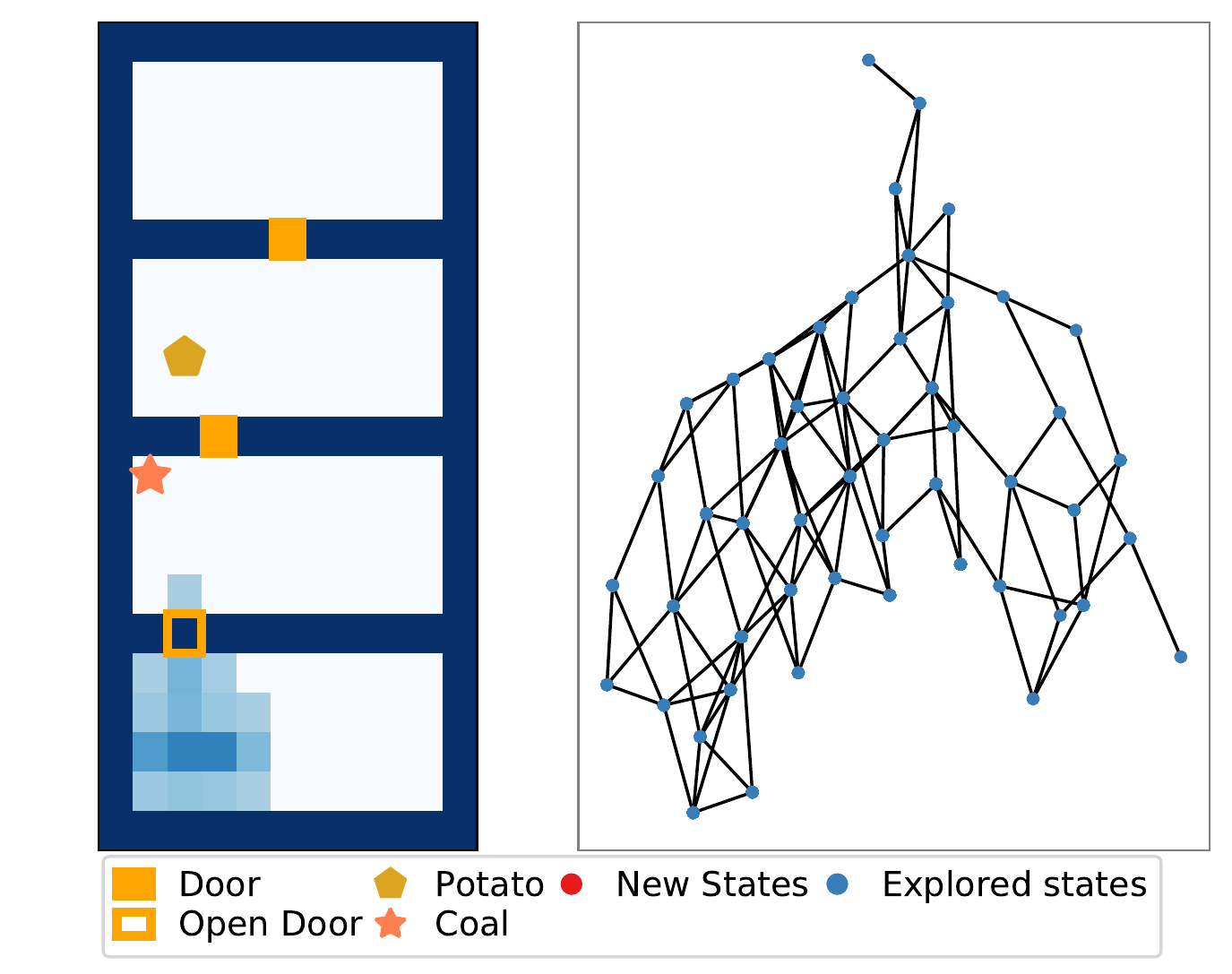}}%
	\hfill
	\subfloat[iteration 10]{\includegraphics[width=0.25\linewidth]{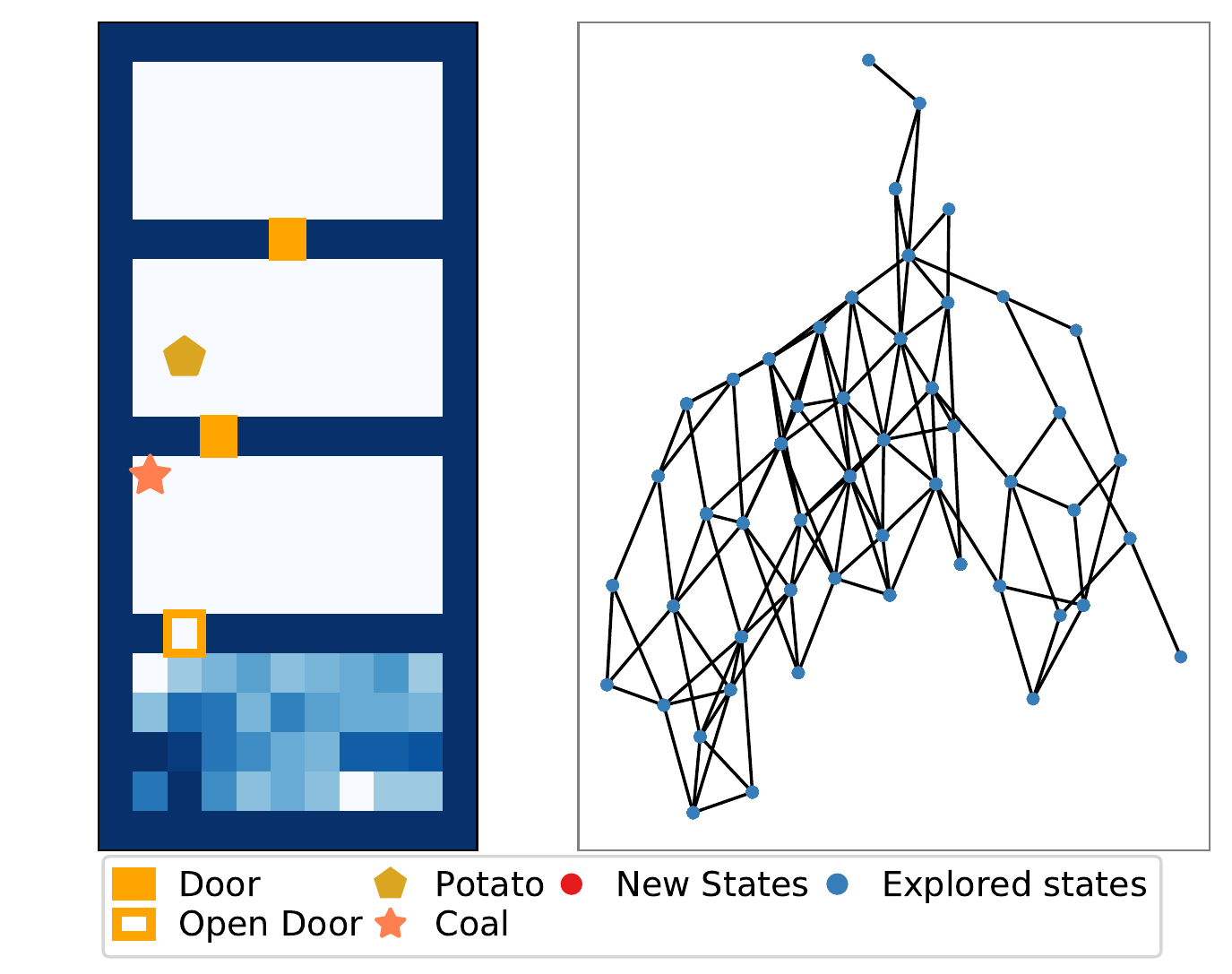}}%
	\hfill
	\subfloat[iteration 11]{\includegraphics[width=0.25\linewidth]{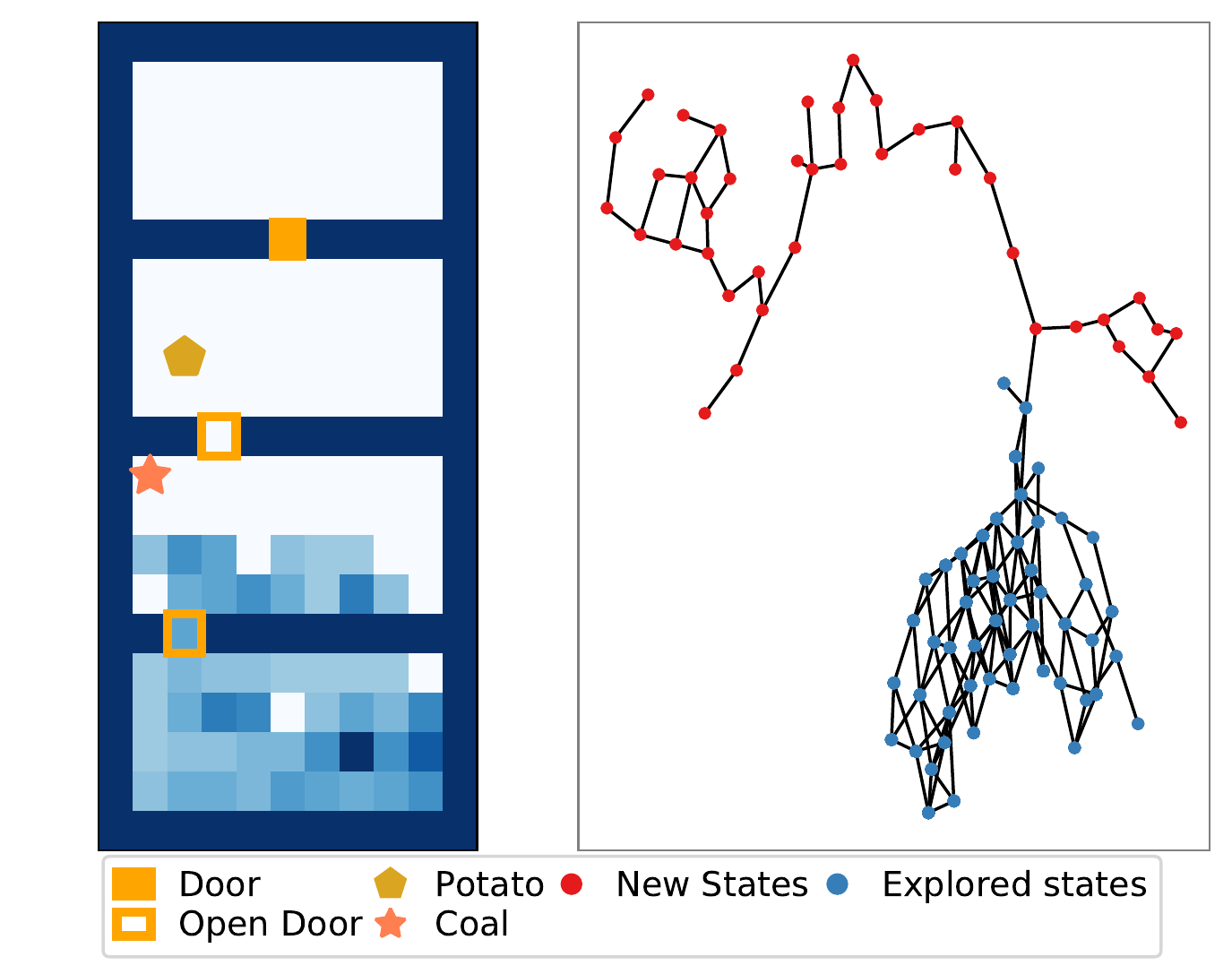}}%
	\vfill
	\subfloat[iteration 12]{\includegraphics[width=0.25\linewidth]{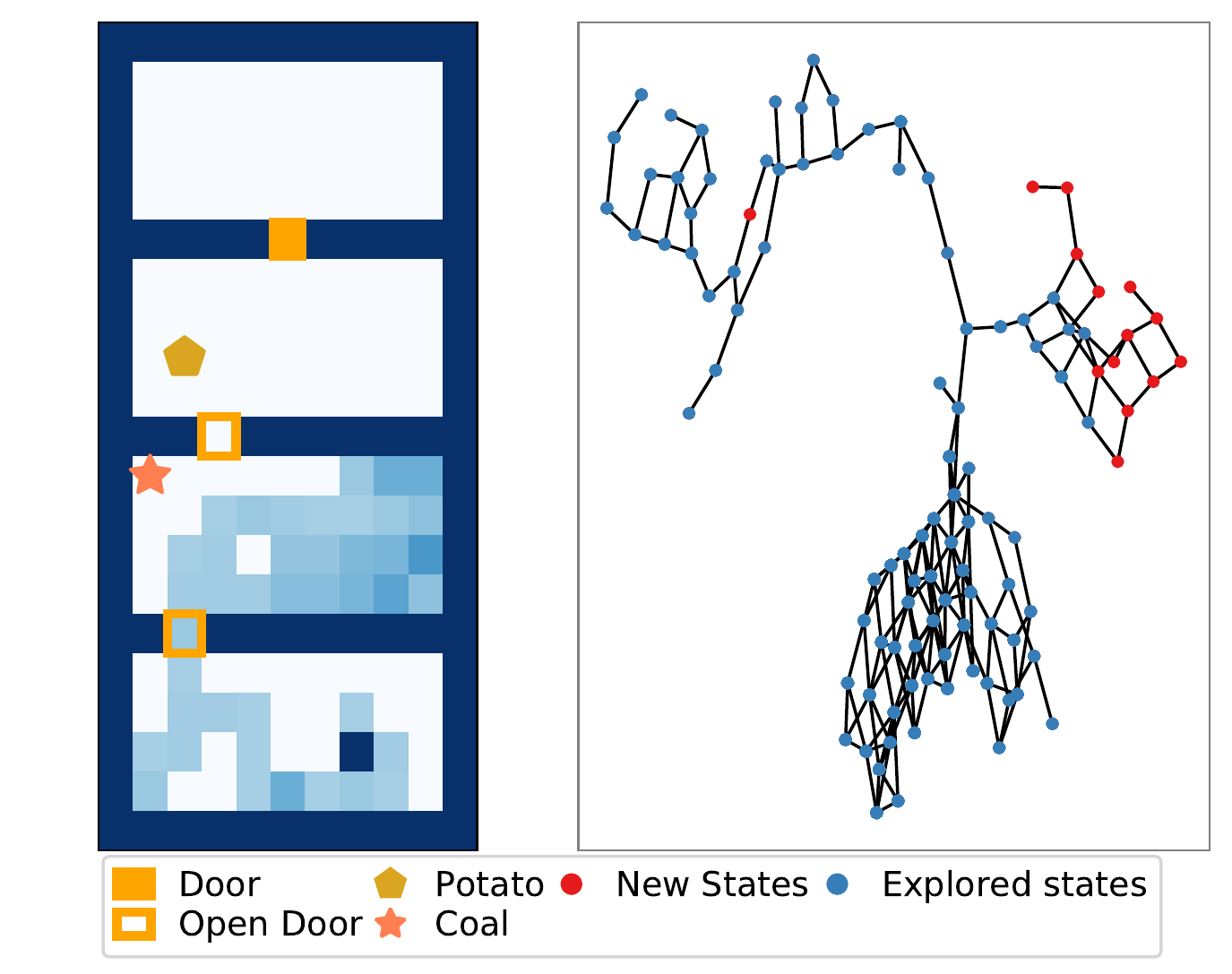}}%
	\hfill
	\subfloat[iteration 13]{\includegraphics[width=0.25\linewidth]{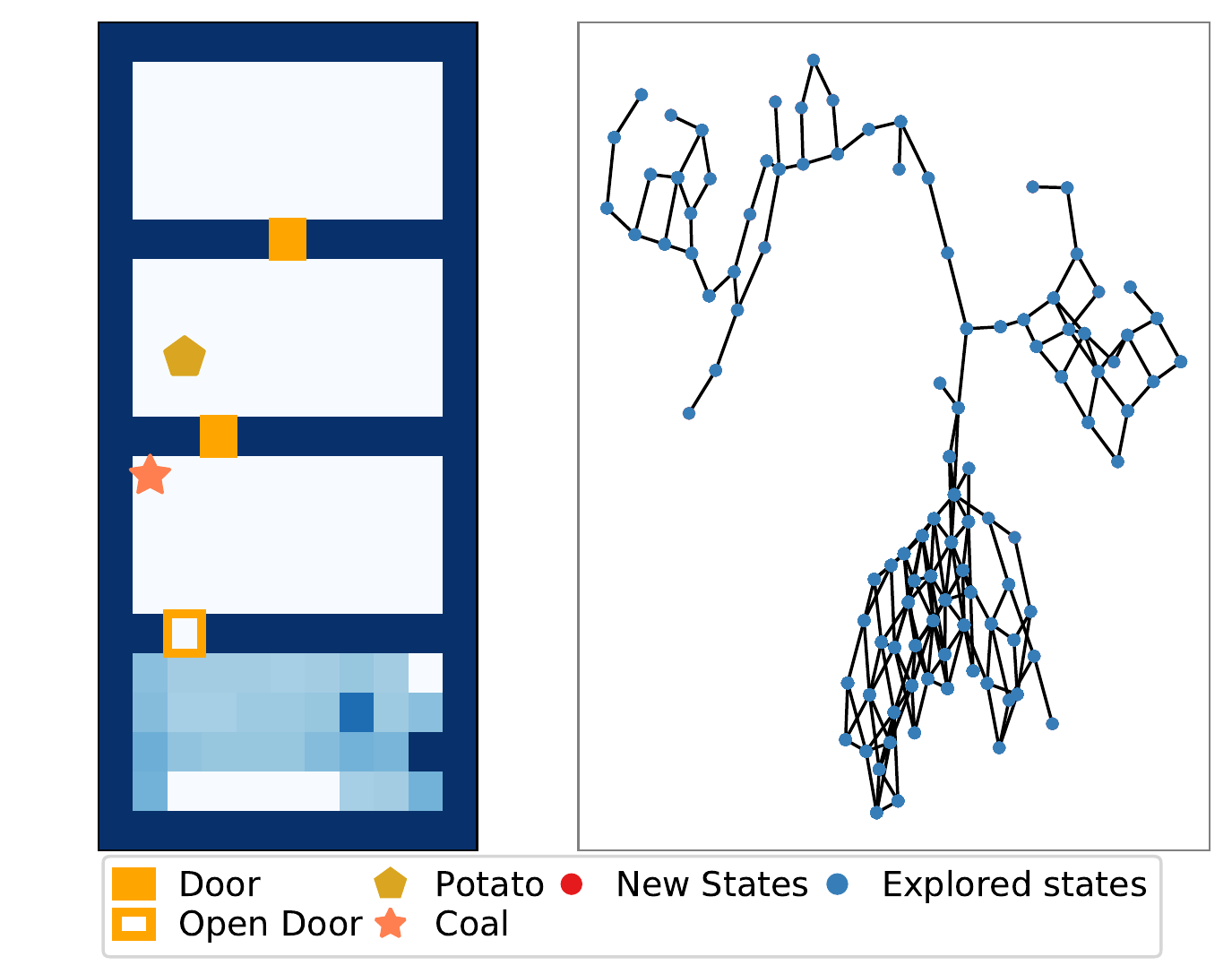}}%
	\hfill
	\subfloat[iteration 14]{\includegraphics[width=0.25\linewidth]{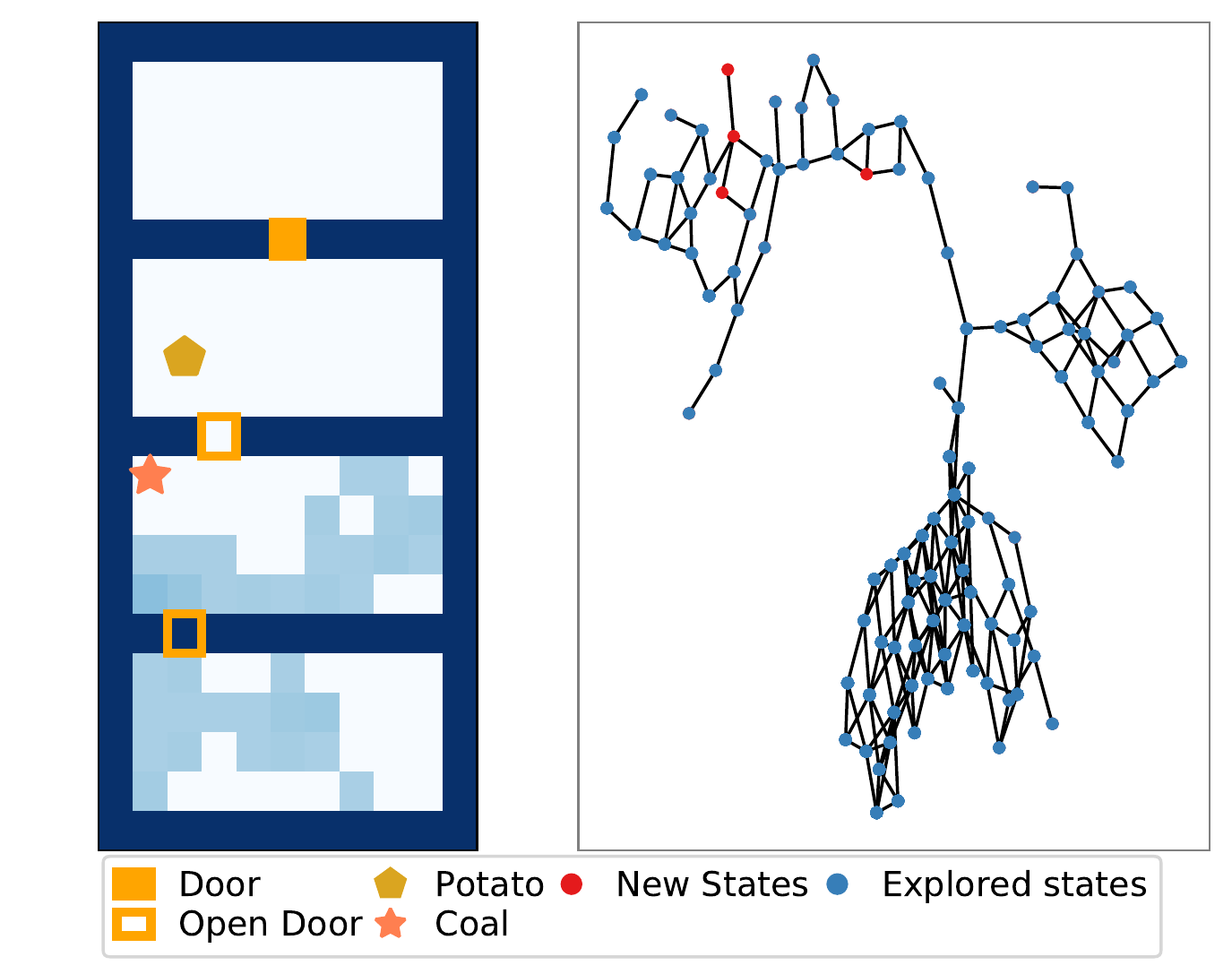}}%
	\hfill
	\subfloat[iteration 15]{\includegraphics[width=0.25\linewidth]{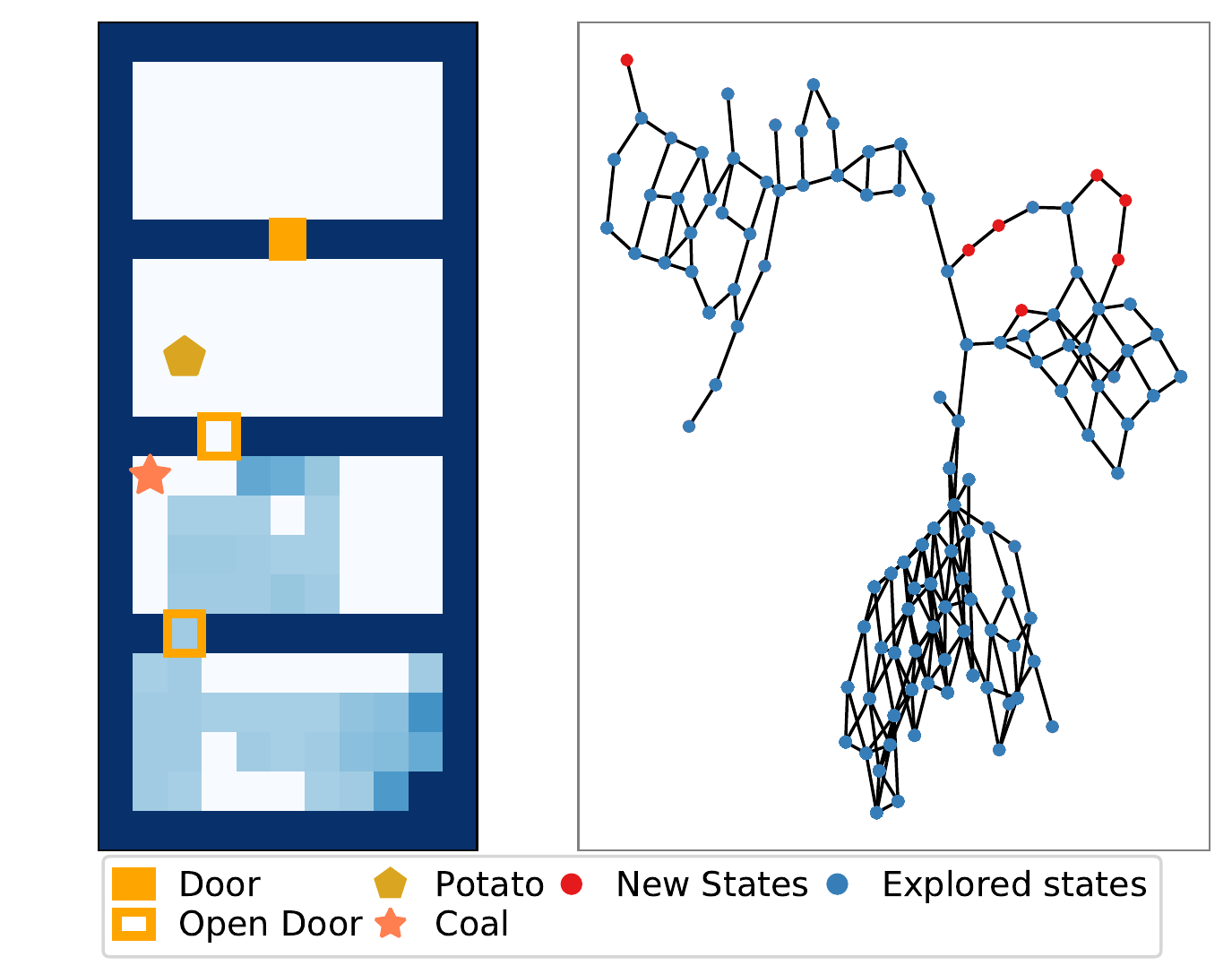}}%
	\vfill
	\subfloat[iteration 16]{\includegraphics[width=0.25\linewidth]{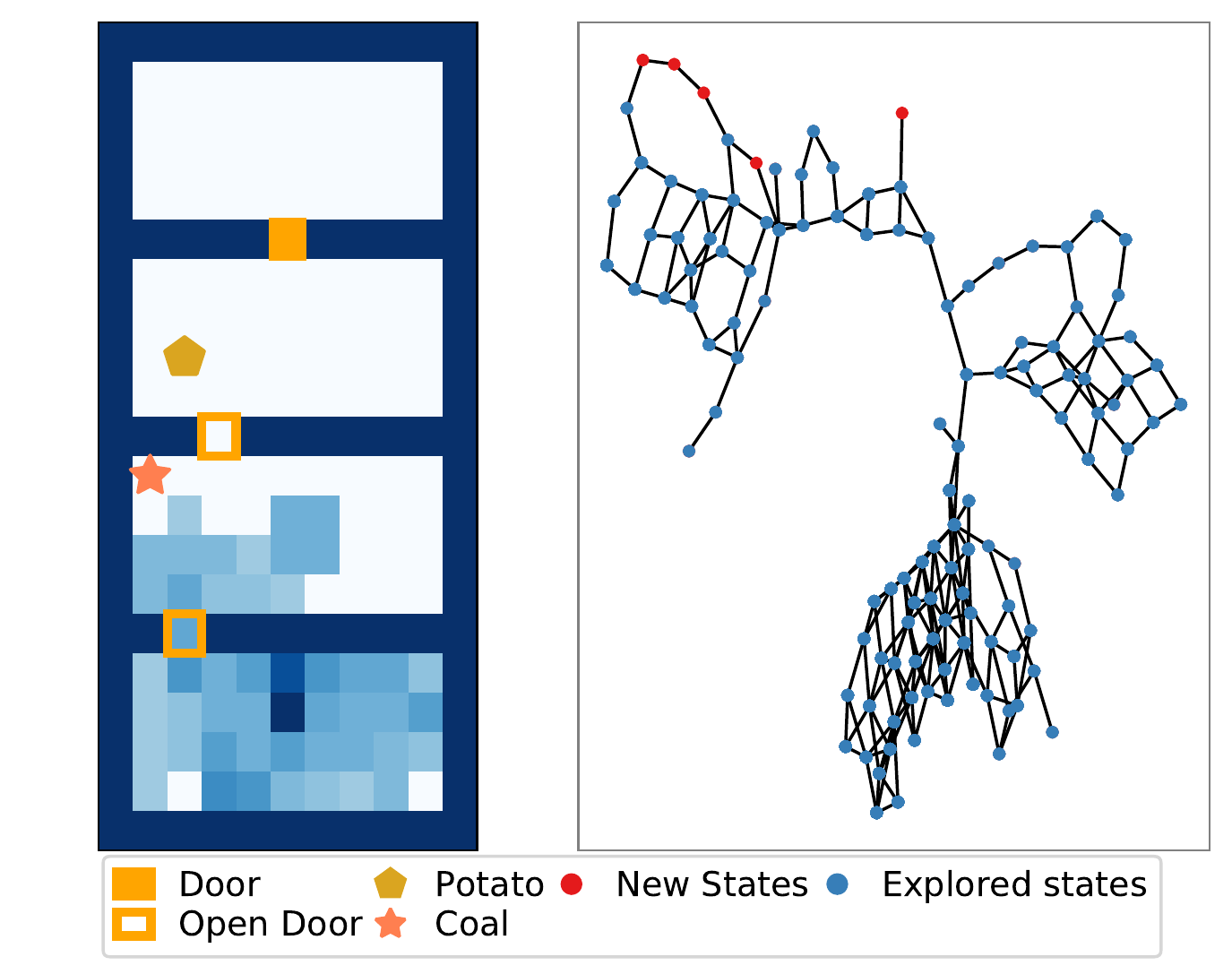}}%
	\hfill
	\subfloat[iteration 17]{\includegraphics[width=0.25\linewidth]{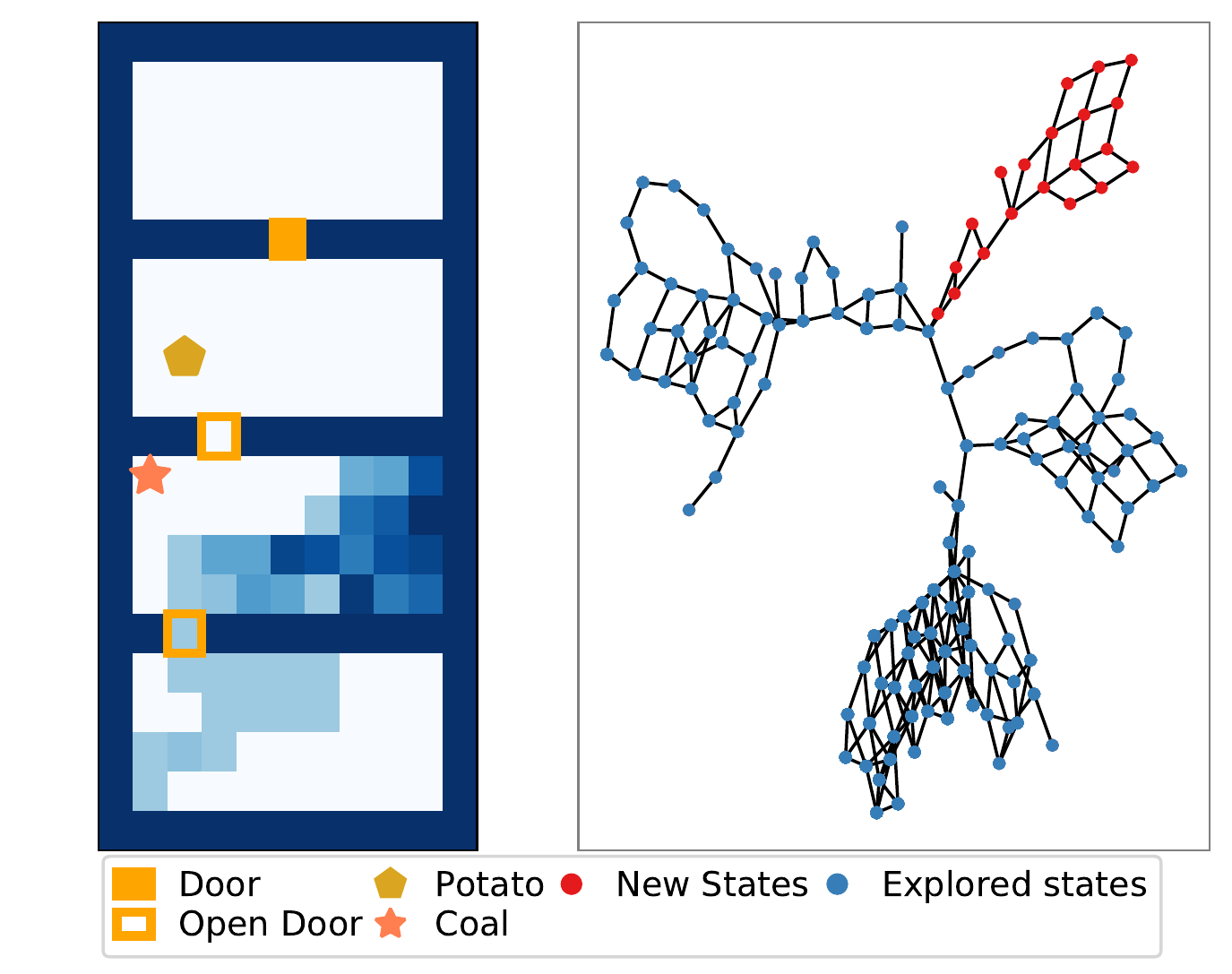}}%
	\hfill
	\subfloat[iteration 18]{\includegraphics[width=0.25\linewidth]{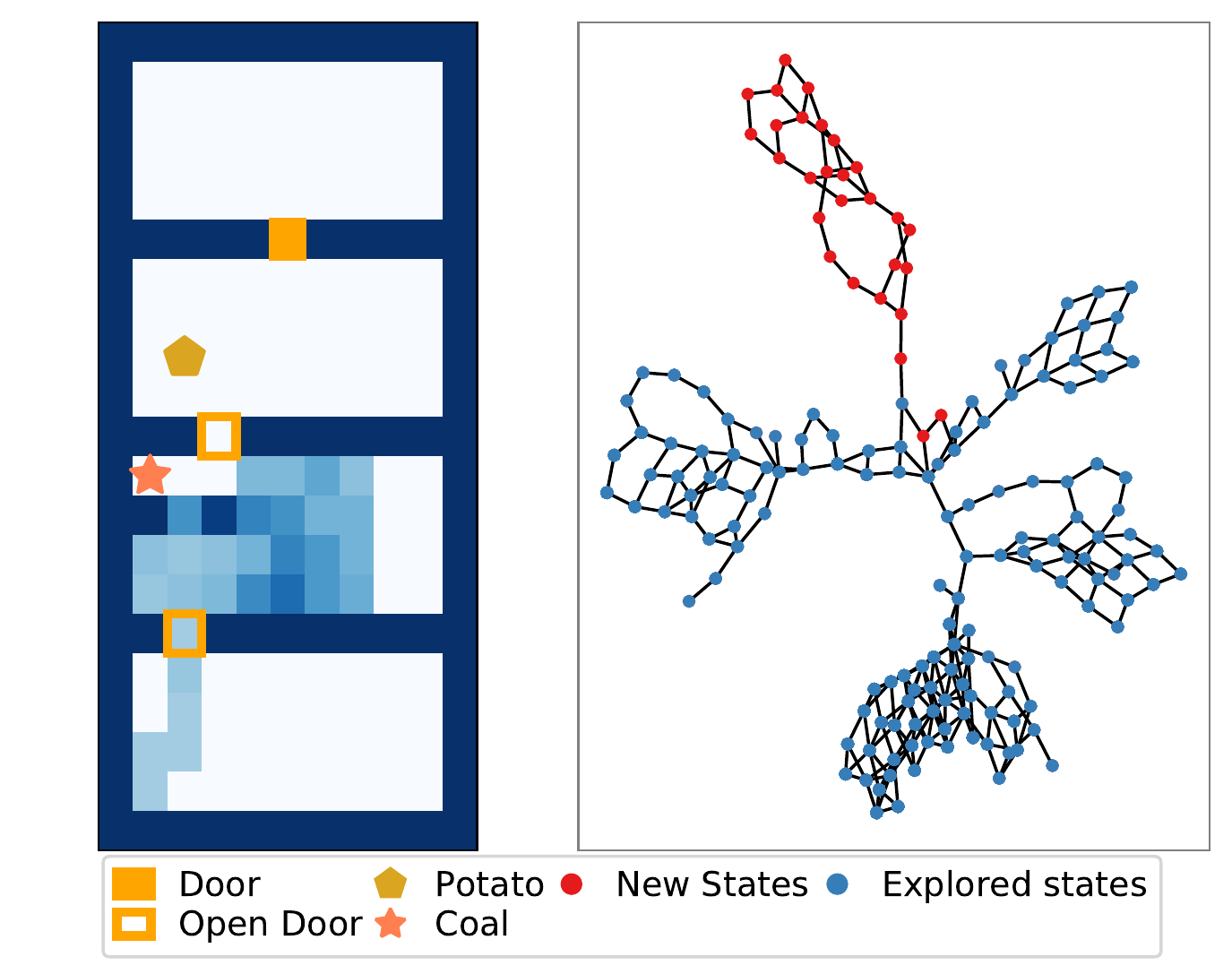}}%
	\hfill
	\subfloat[iteration 19]{\includegraphics[width=0.25\linewidth]{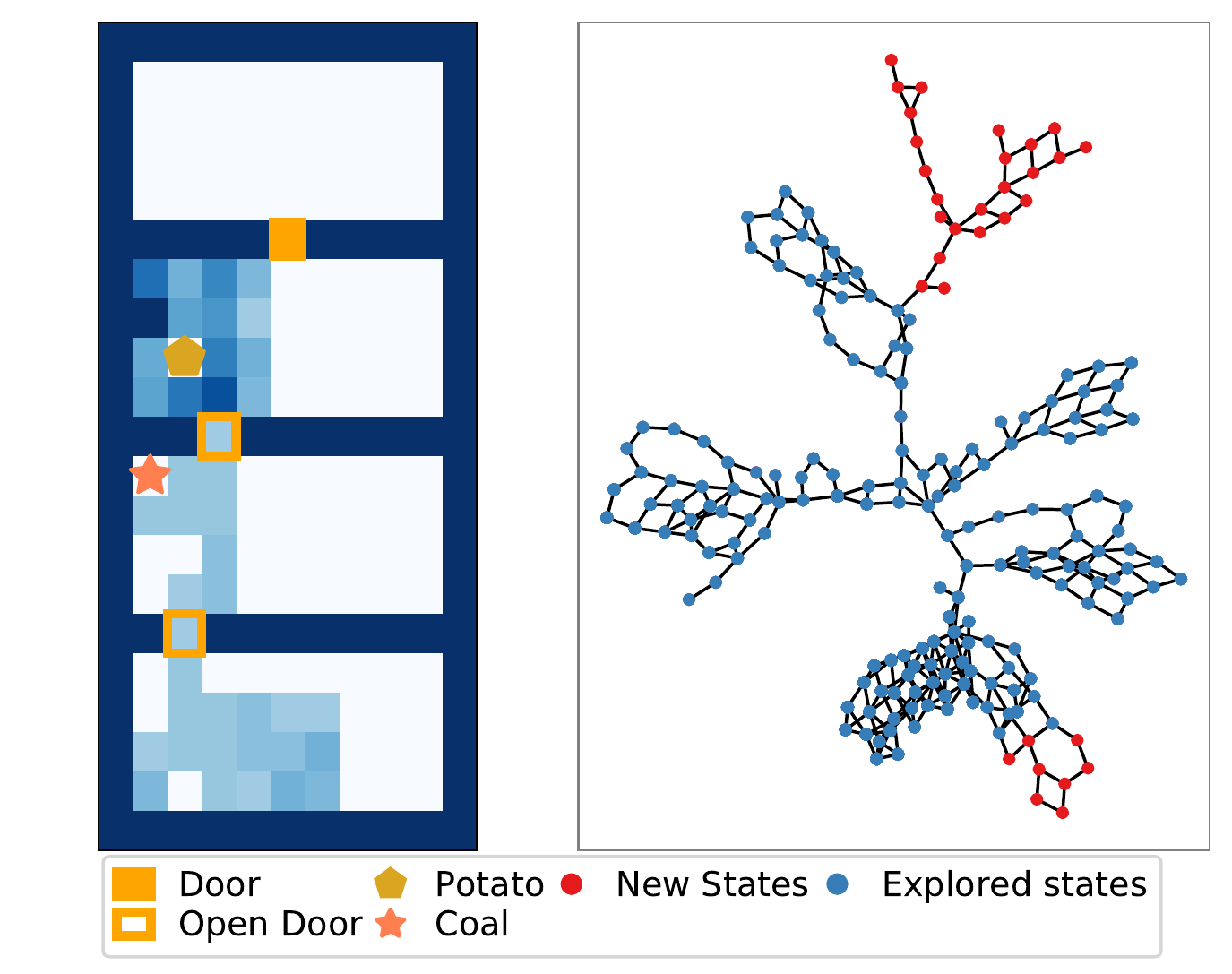}}%
	\hfill
    \caption{(Eigenoptions Agent) State visitation frequencies and constructed graph per iteration in Minecraft Bake-Rooms}
    \label{fig:state-visitation-bake-rooms-eigen}
\end{figure*}

\begin{figure*}
    \centering
	\subfloat[iteration 0]{\includegraphics[width=0.25\linewidth]{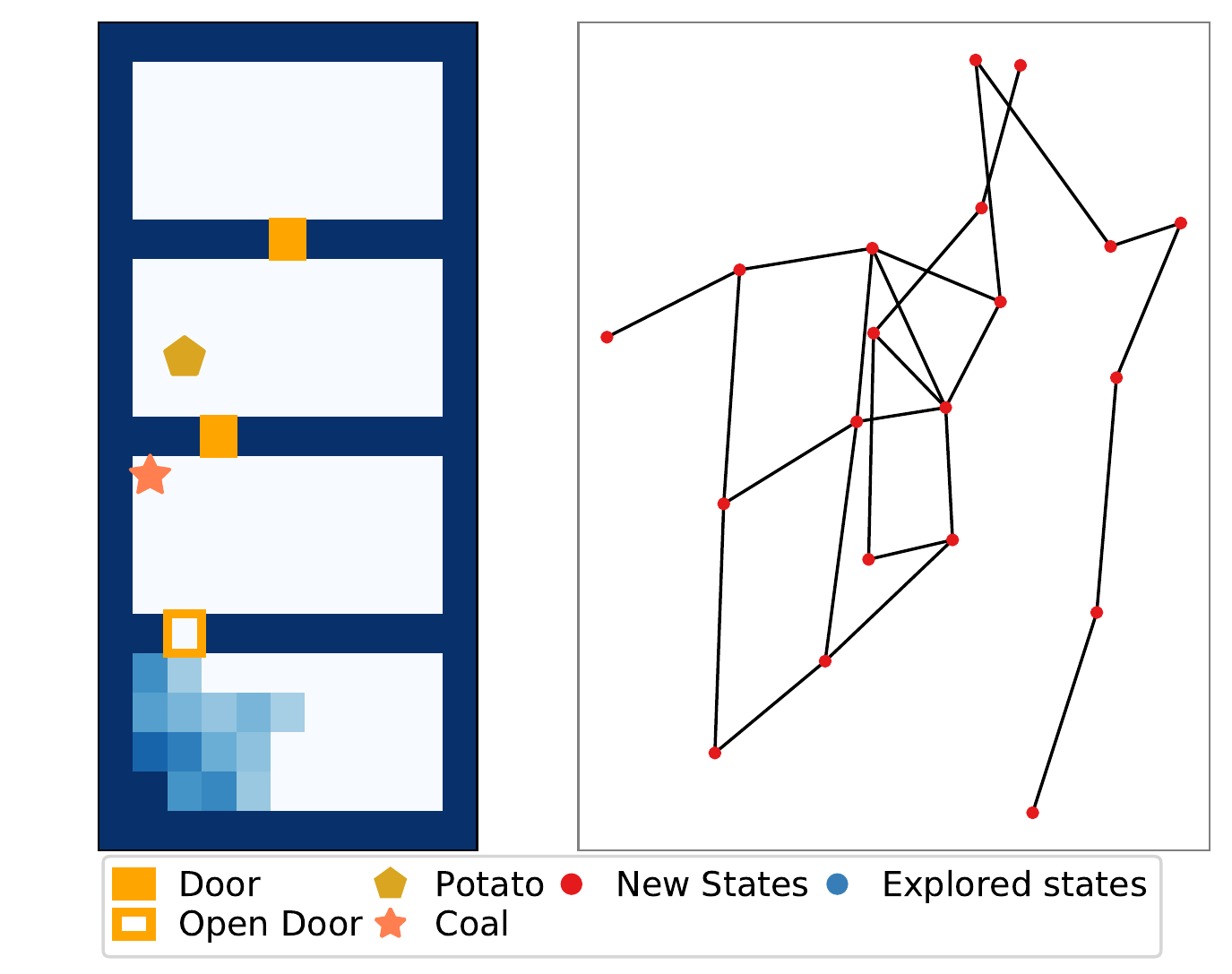}}%
	\hfill
	\subfloat[iteration 1]{\includegraphics[width=0.25\linewidth]{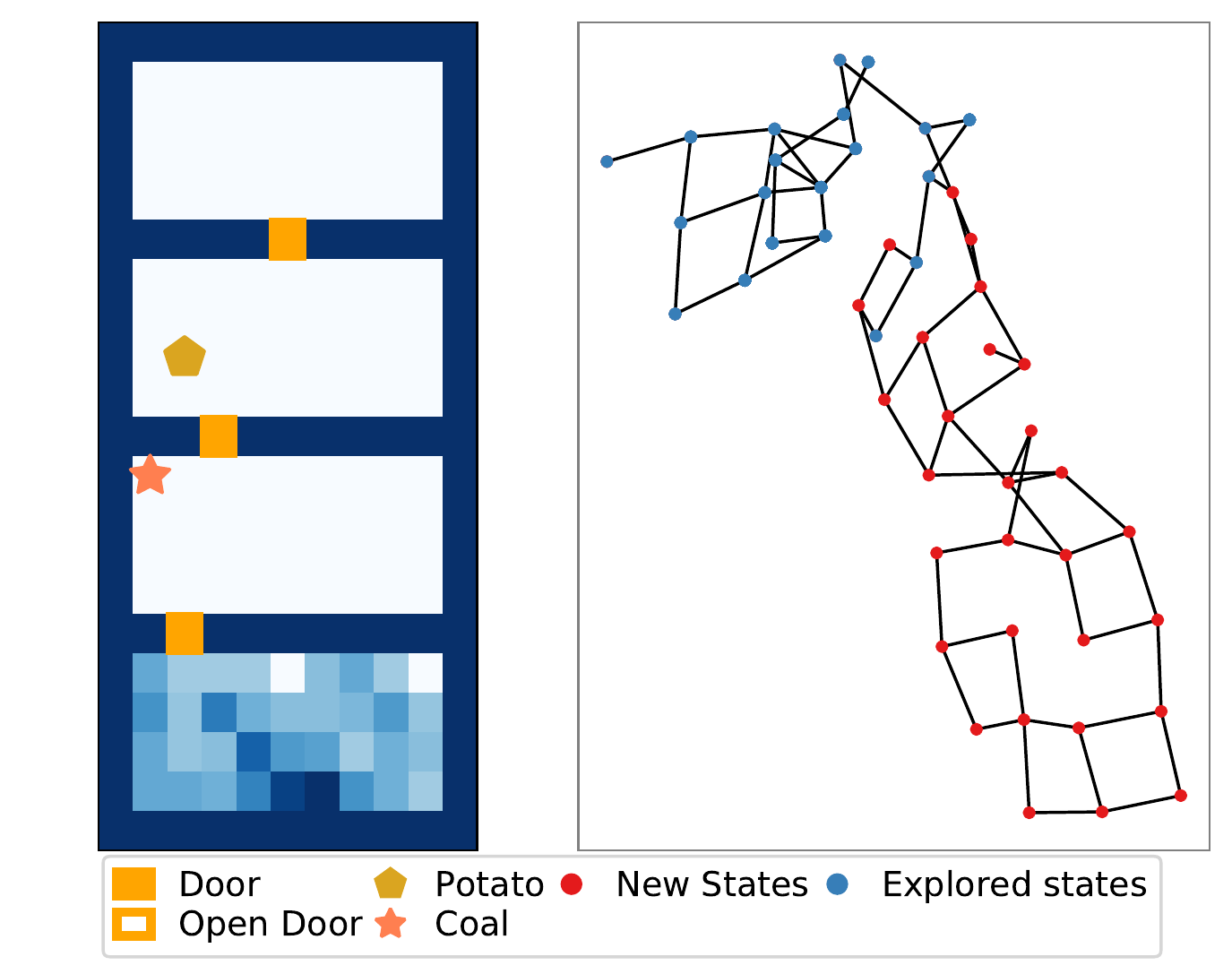}}%
	\hfill
	\subfloat[iteration 2]{\includegraphics[width=0.25\linewidth]{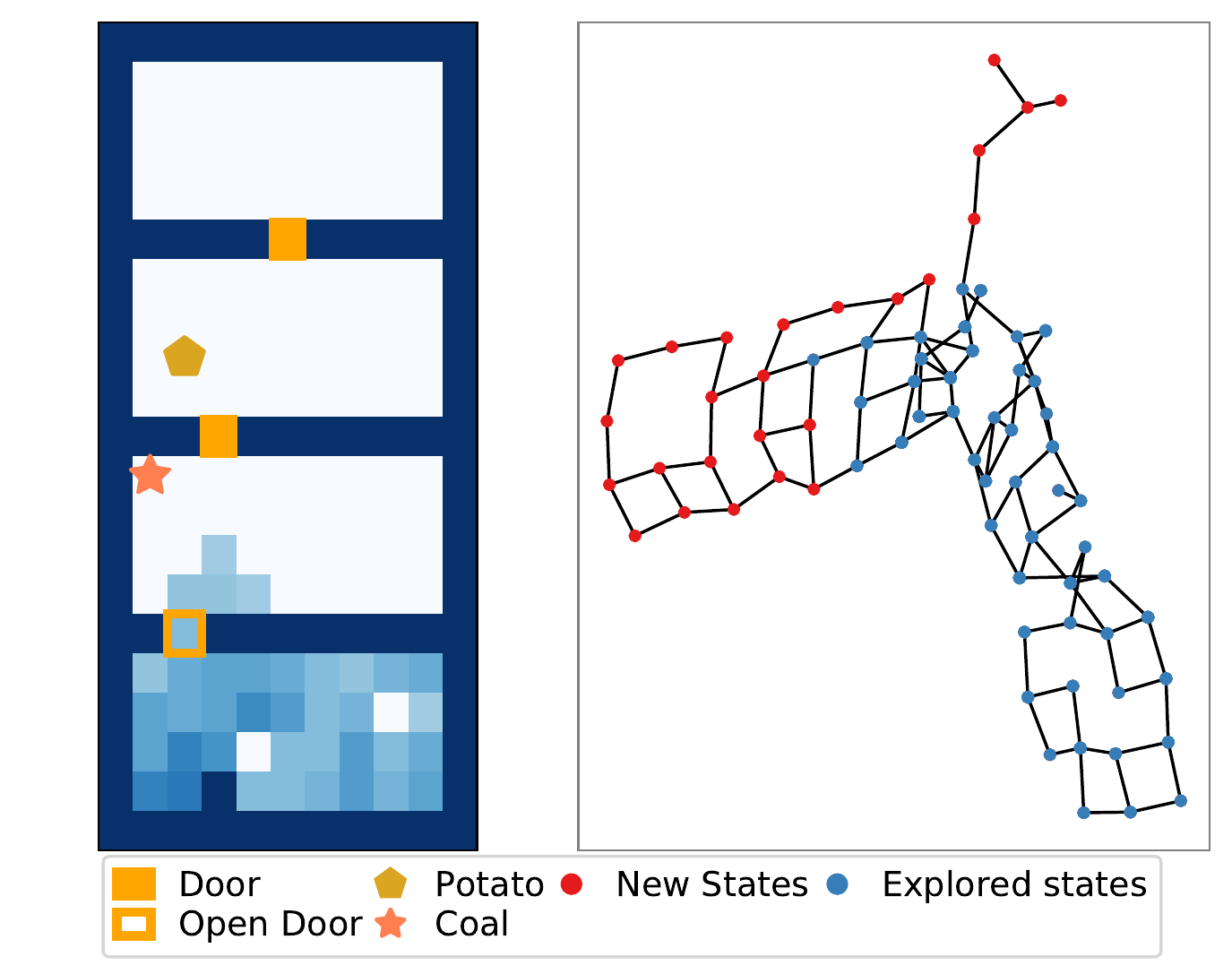}}%
	\hfill
	\subfloat[iteration 3]{\includegraphics[width=0.25\linewidth]{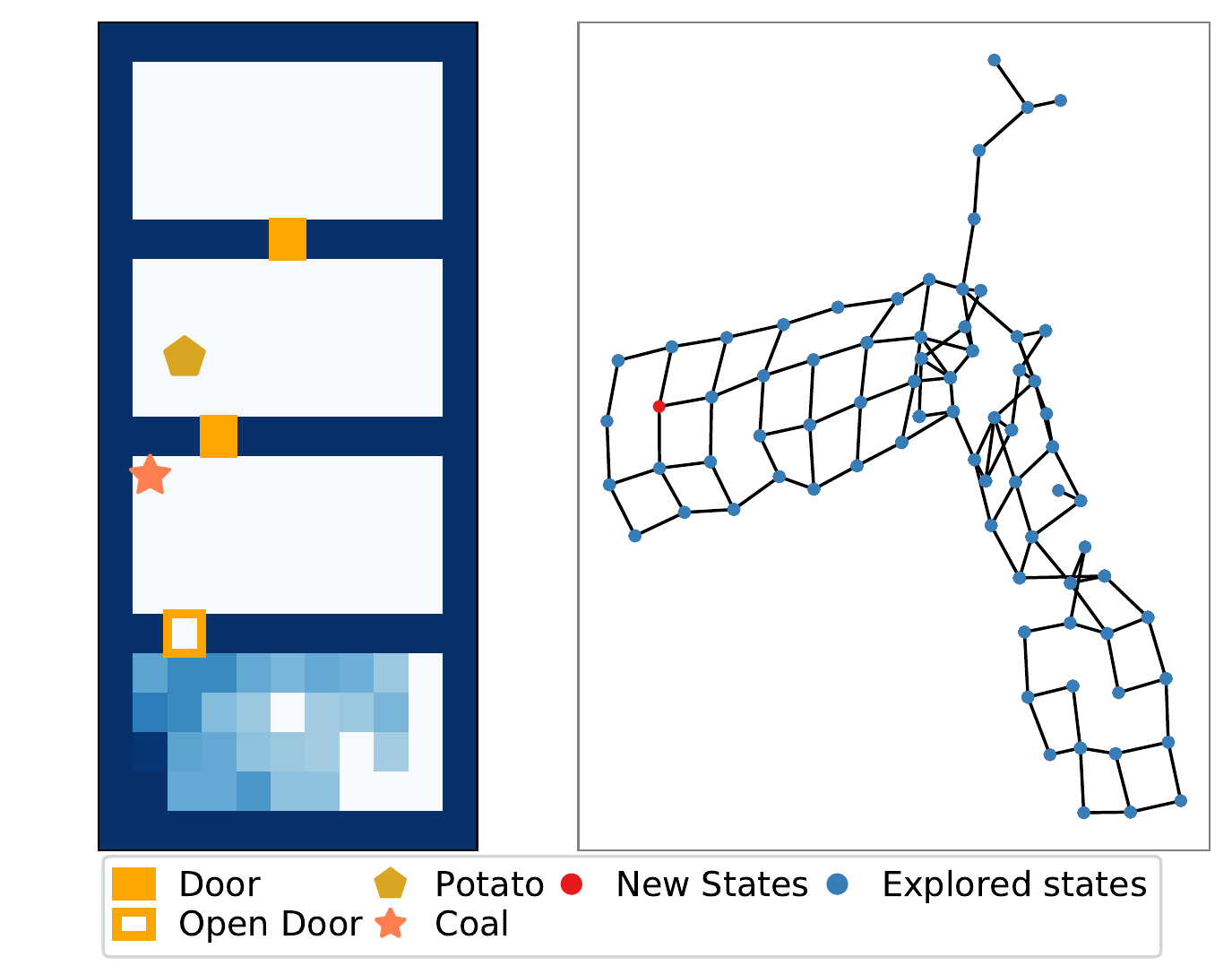}}%
	\vfill
	\subfloat[iteration 4]{\includegraphics[width=0.25\linewidth]{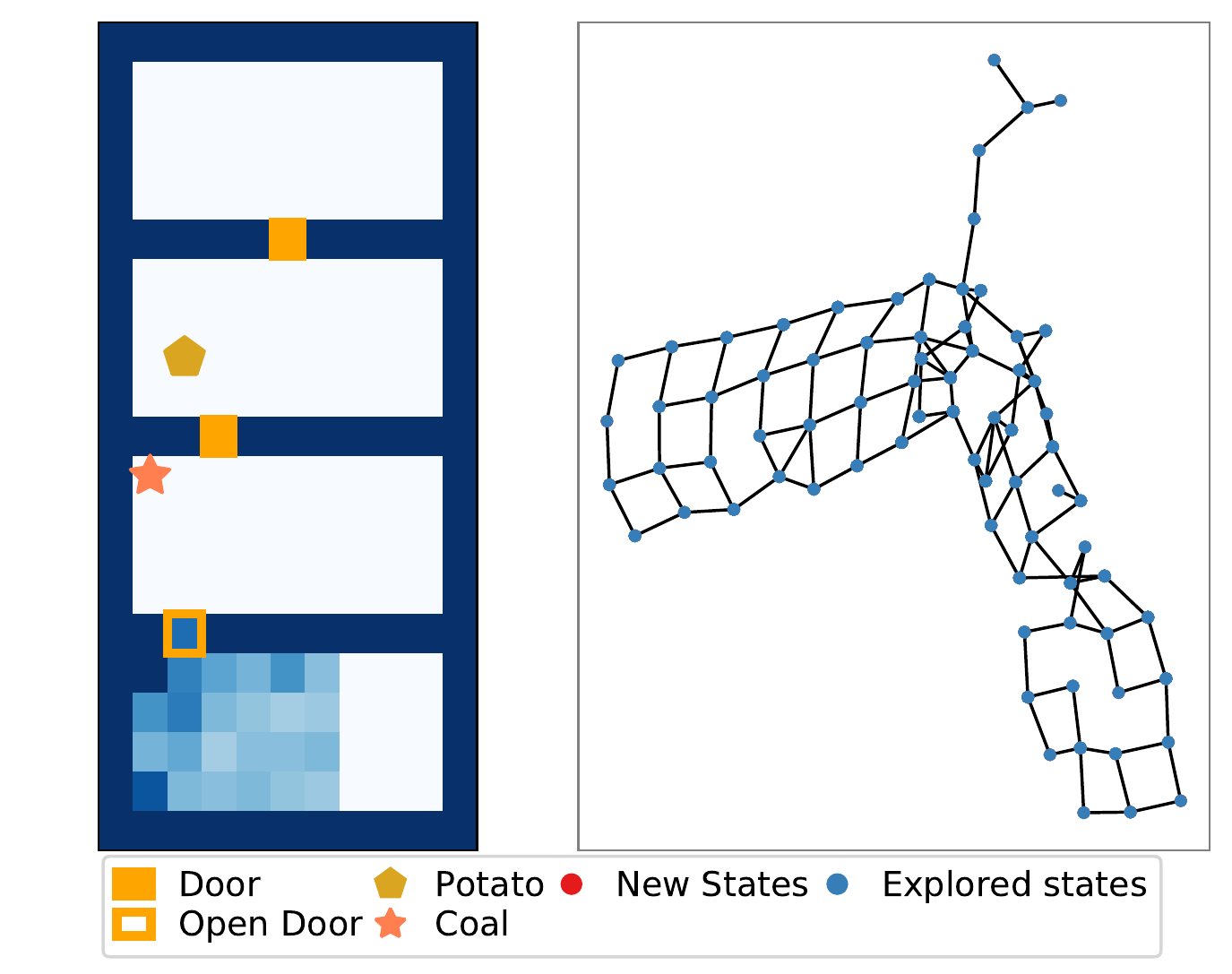}}%
	\hfill
	\subfloat[iteration 5]{\includegraphics[width=0.25\linewidth]{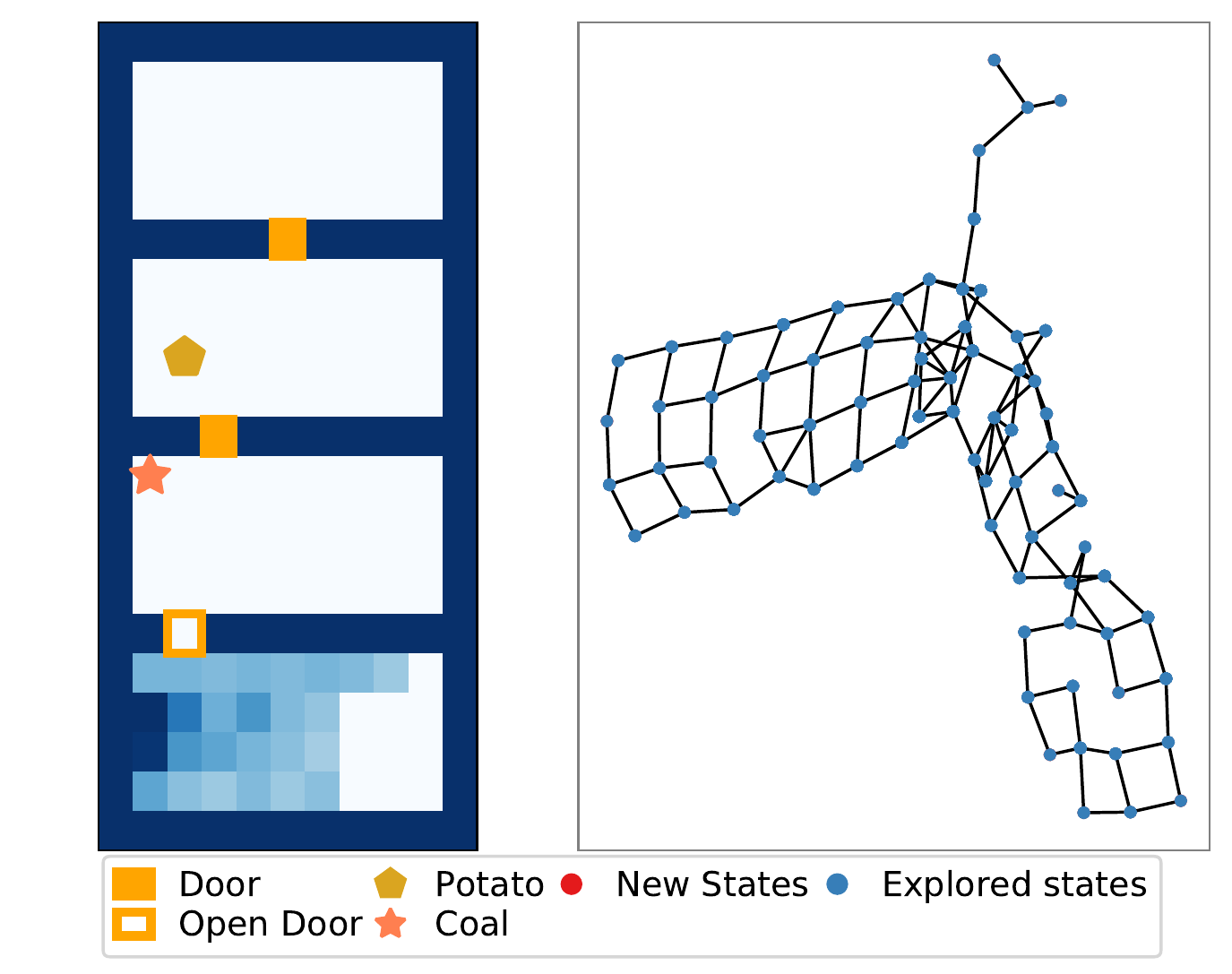}}%
	\hfill
	\subfloat[iteration 6]{\includegraphics[width=0.25\linewidth]{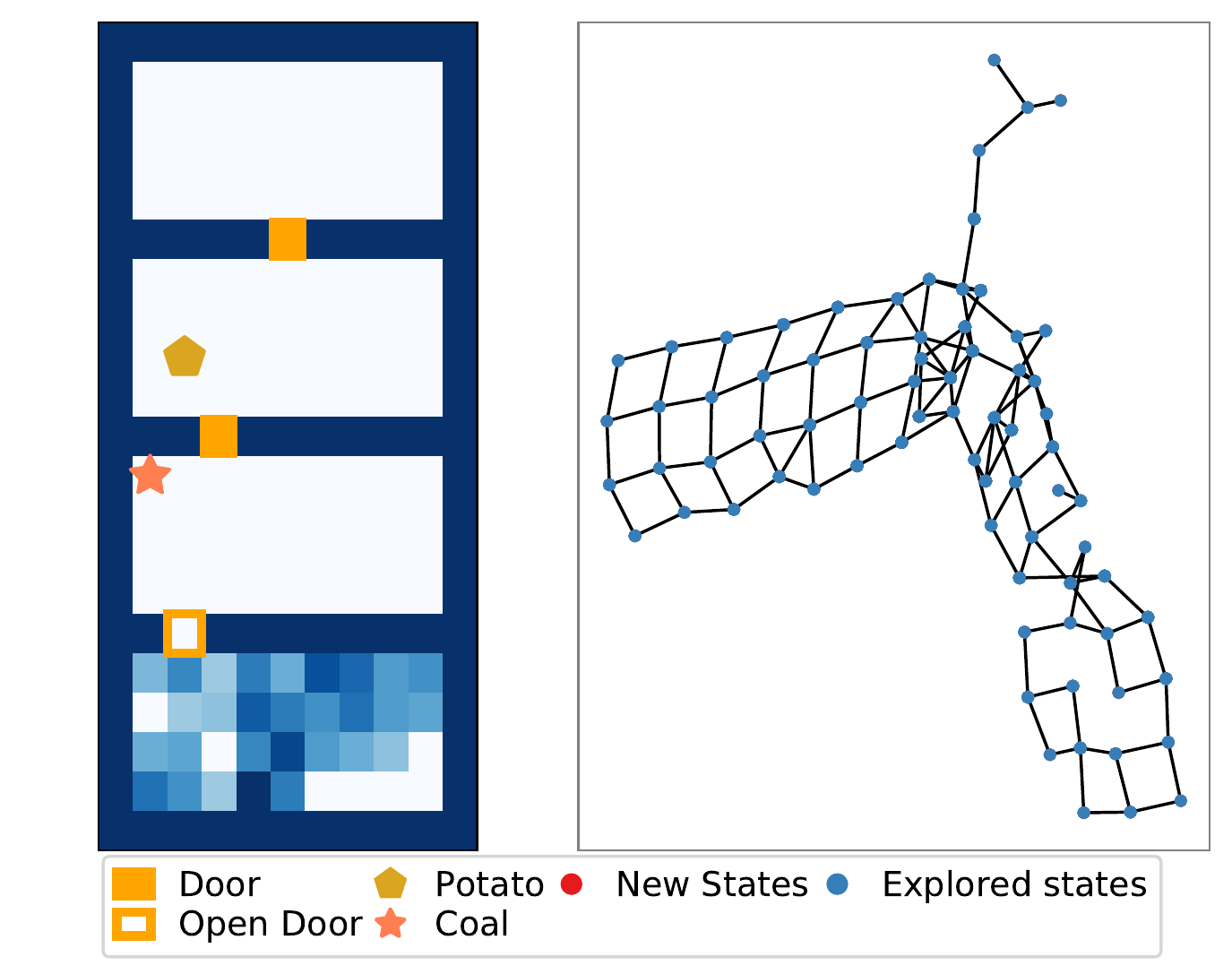}}%
	\hfill
	\subfloat[iteration 7]{\includegraphics[width=0.25\linewidth]{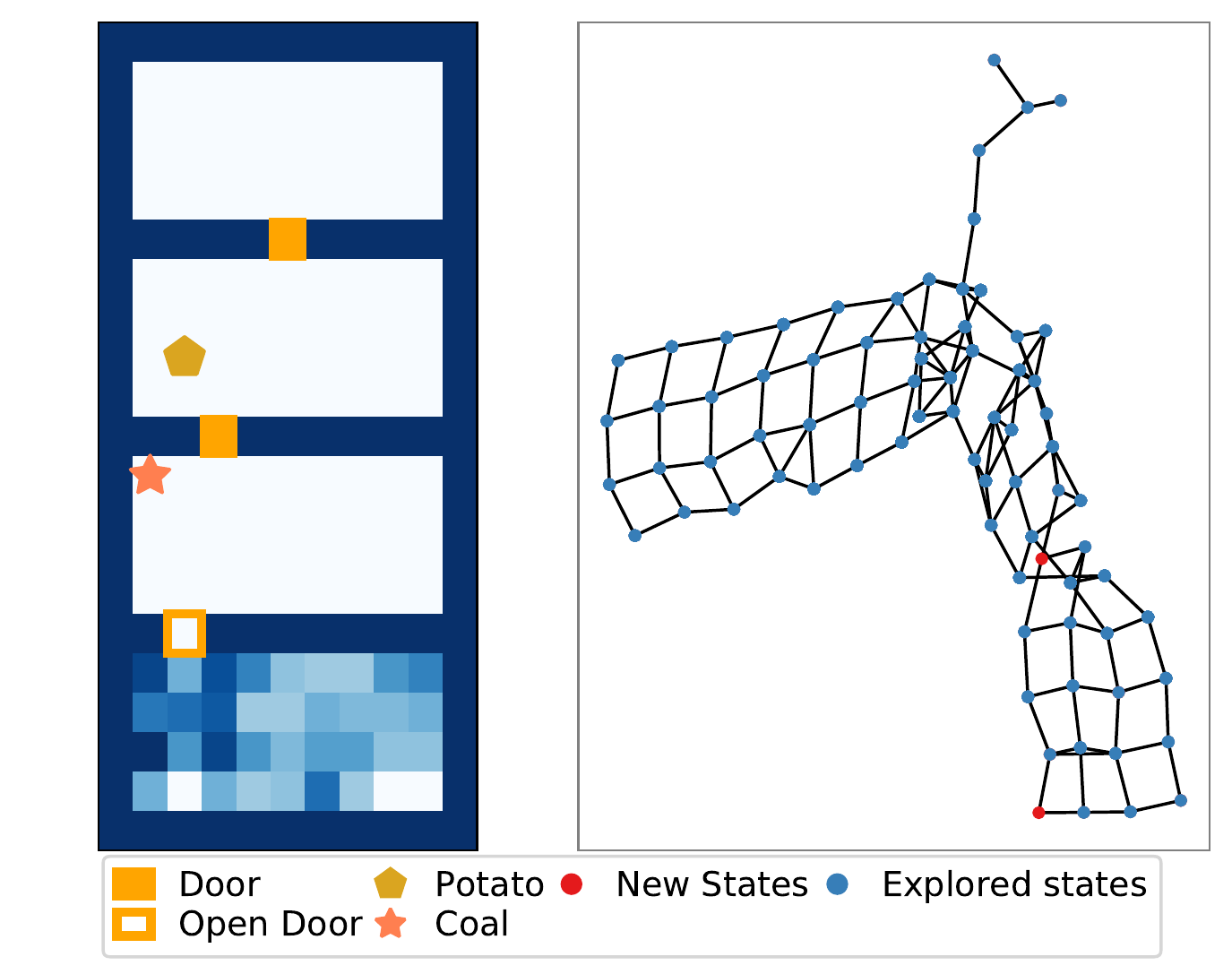}}%
	\vfill
	\subfloat[iteration 8]{\includegraphics[width=0.25\linewidth]{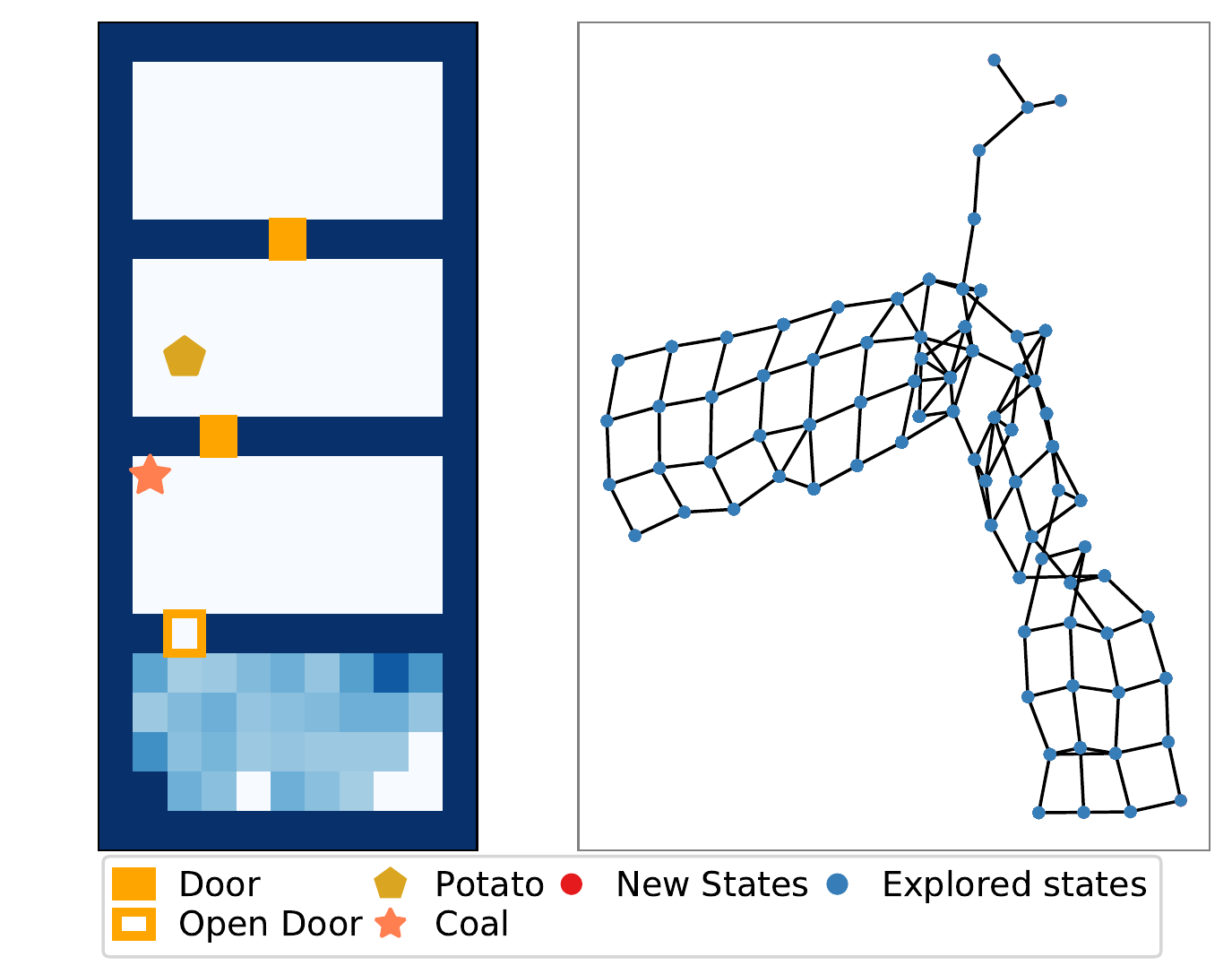}}%
	\hfill
	\subfloat[iteration 9]{\includegraphics[width=0.25\linewidth]{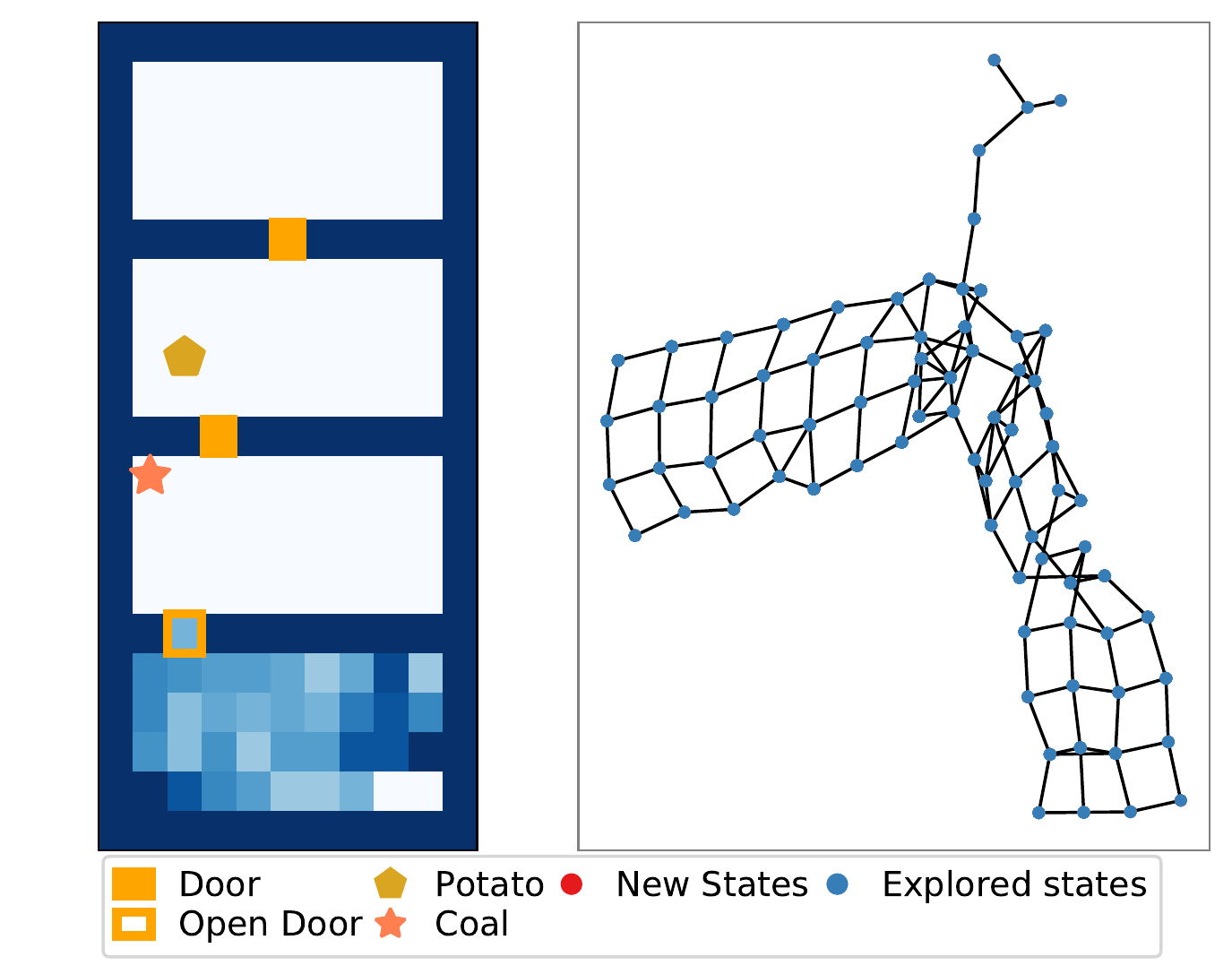}}%
	\hfill
	\subfloat[iteration 10]{\includegraphics[width=0.25\linewidth]{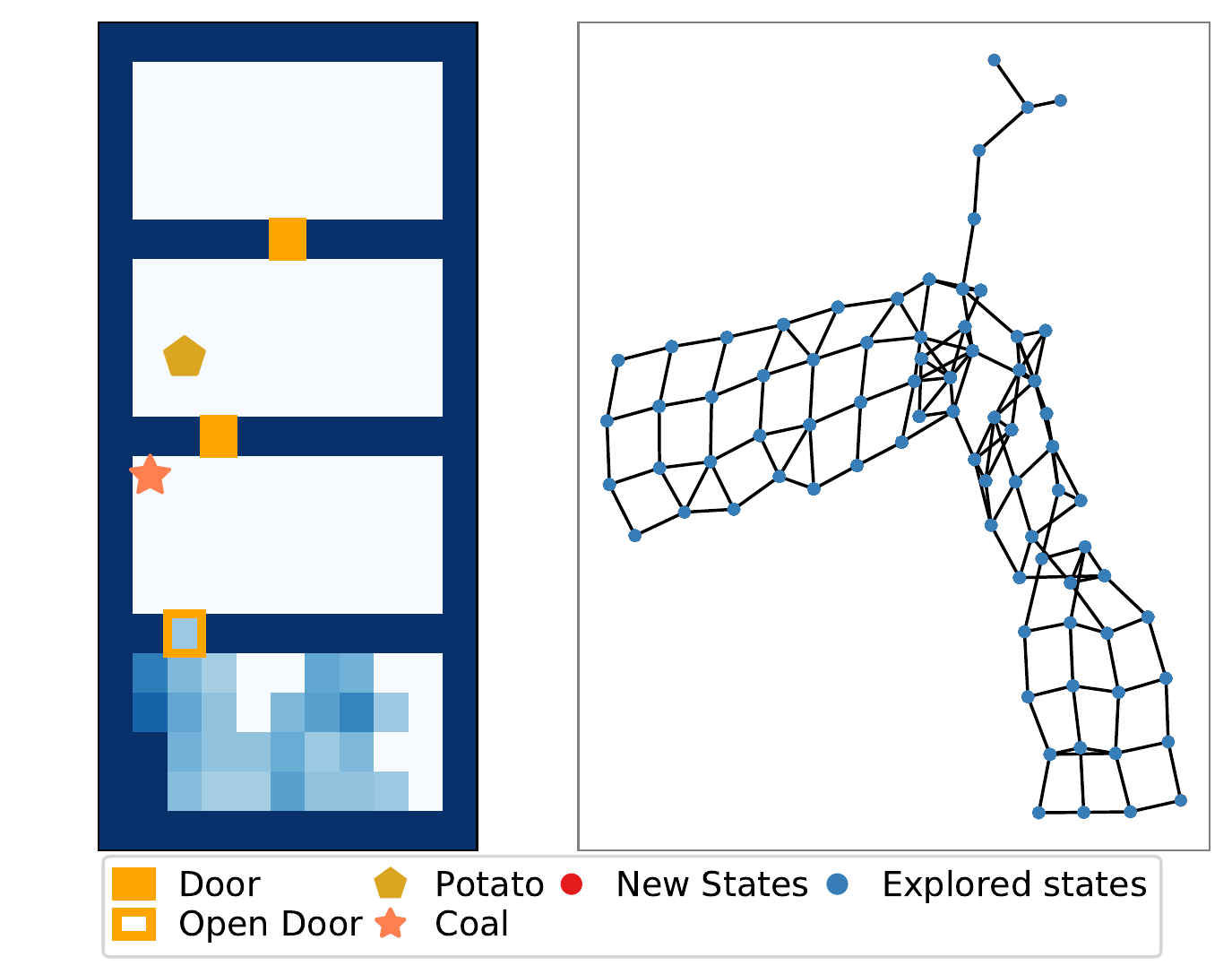}}%
	\hfill
	\subfloat[iteration 11]{\includegraphics[width=0.25\linewidth]{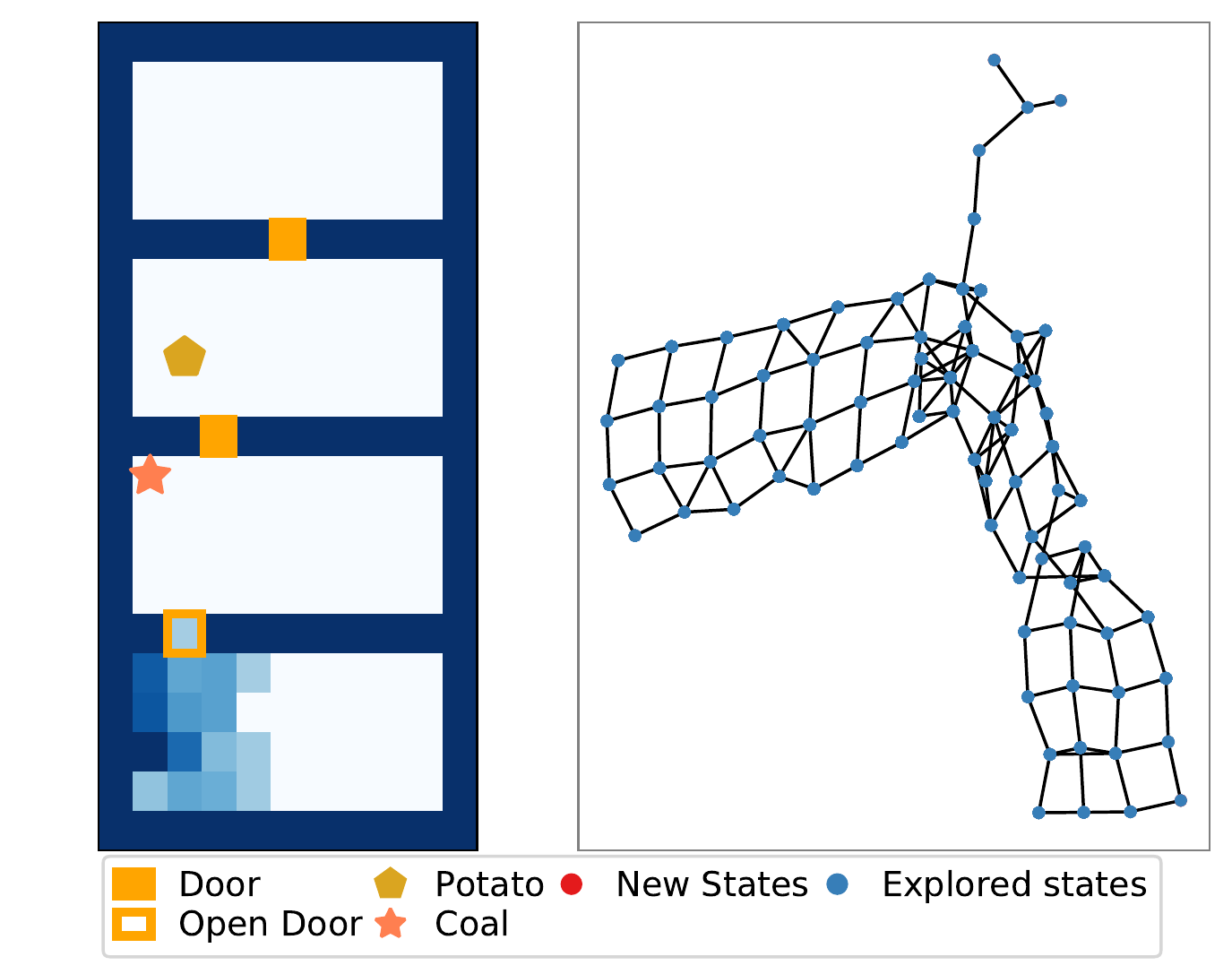}}%
	\vfill
	\subfloat[iteration 12]{\includegraphics[width=0.25\linewidth]{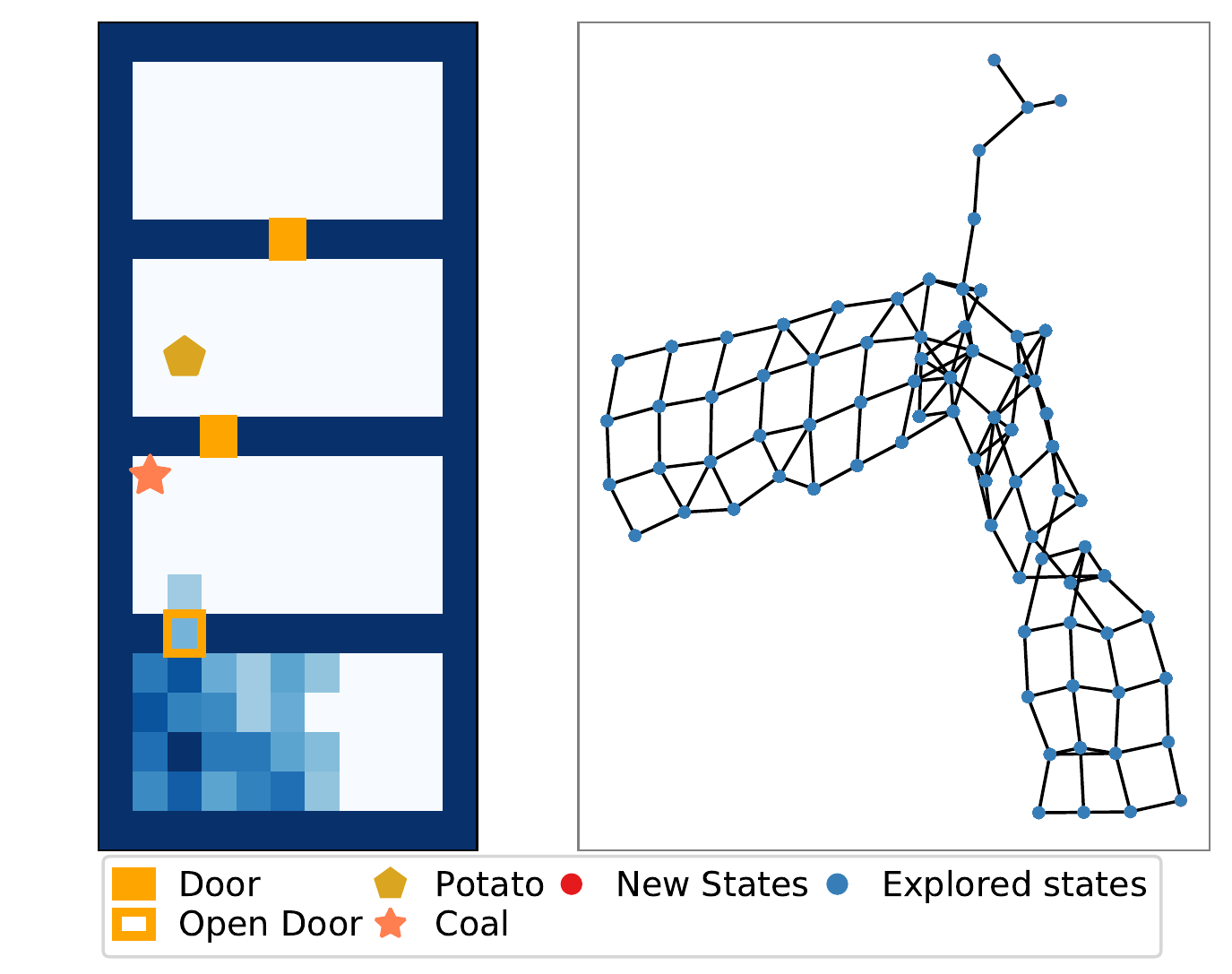}}%
	\hfill
	\subfloat[iteration 13]{\includegraphics[width=0.25\linewidth]{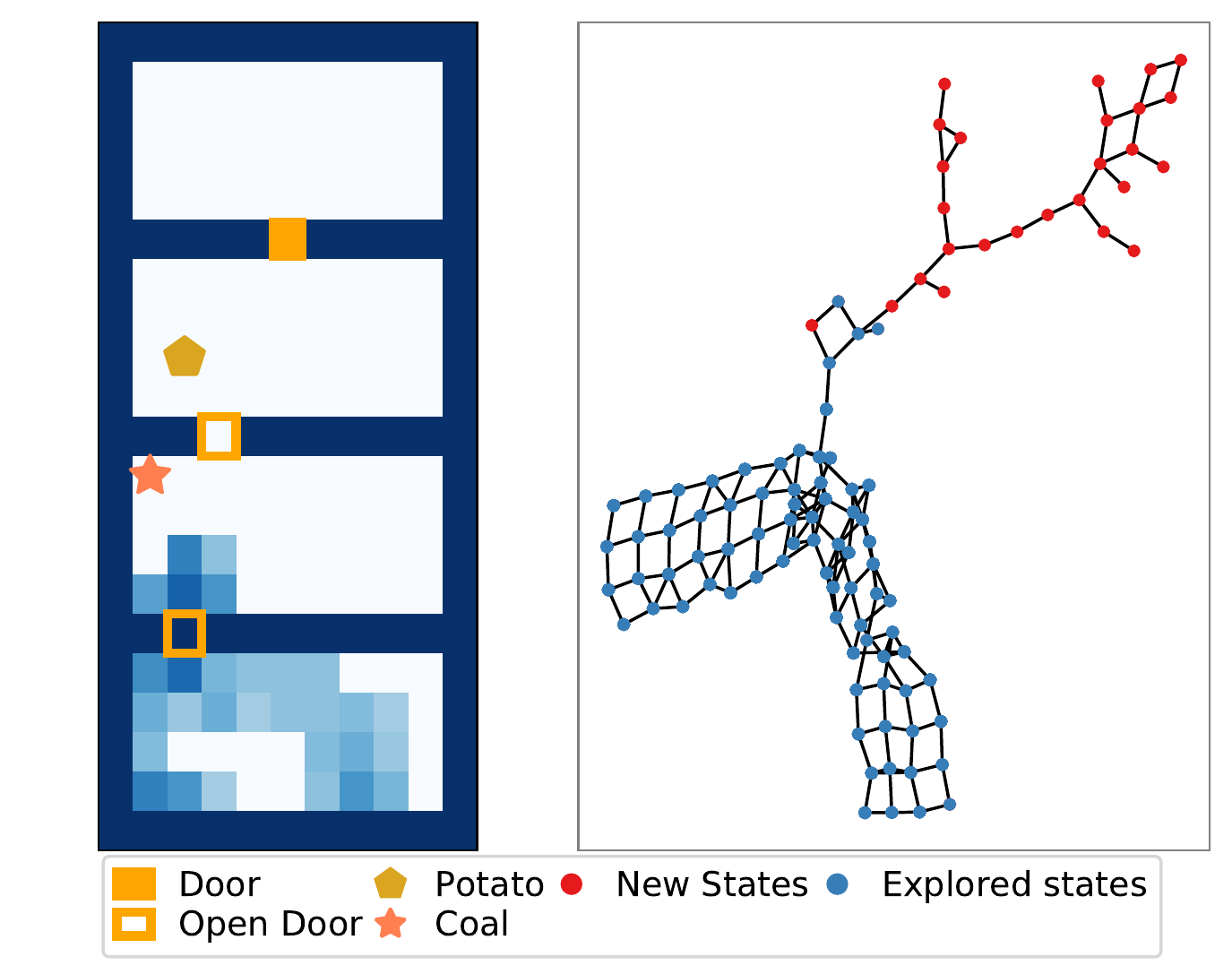}}%
	\hfill
	\subfloat[iteration 14]{\includegraphics[width=0.25\linewidth]{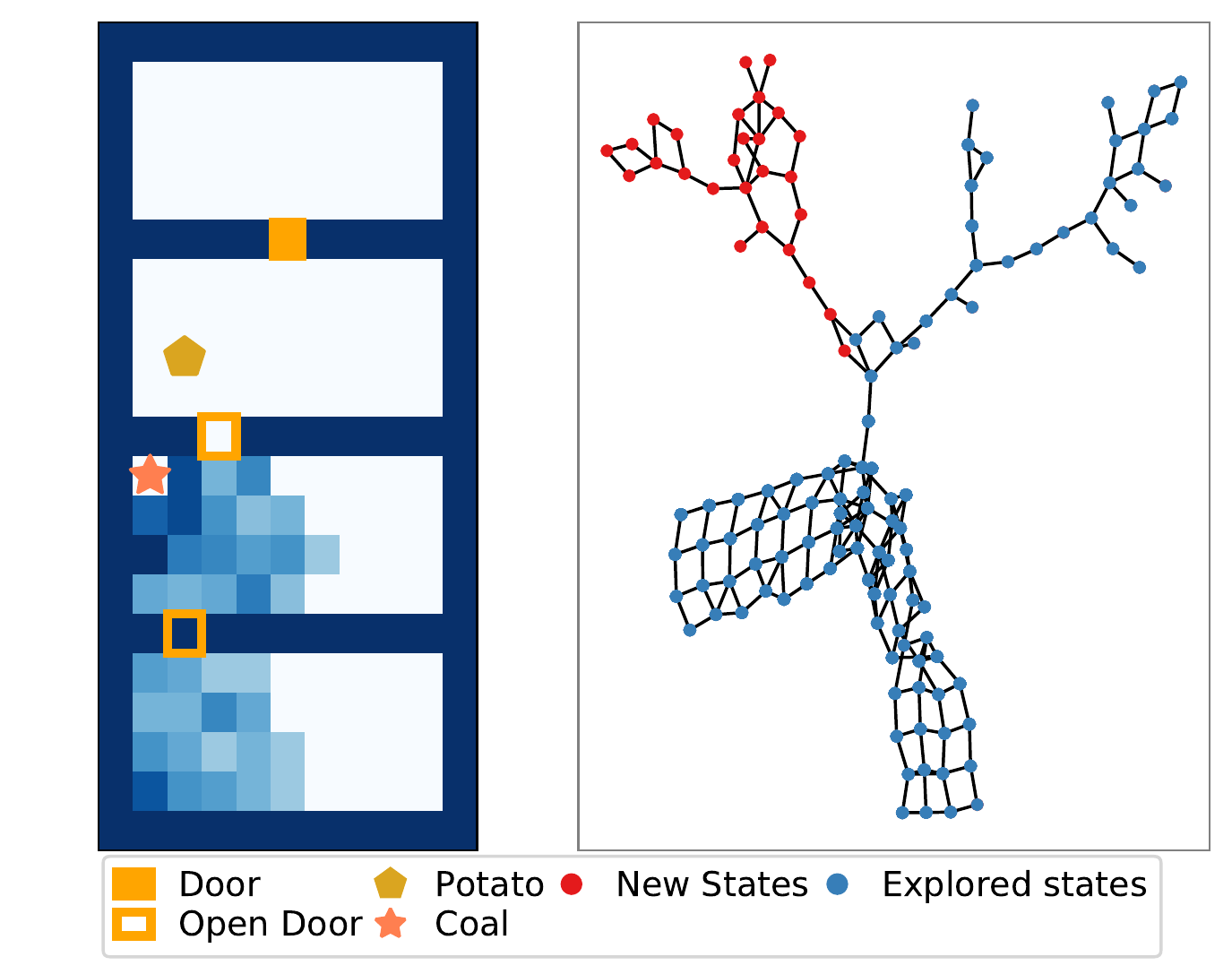}}%
	\hfill
	\subfloat[iteration 15]{\includegraphics[width=0.25\linewidth]{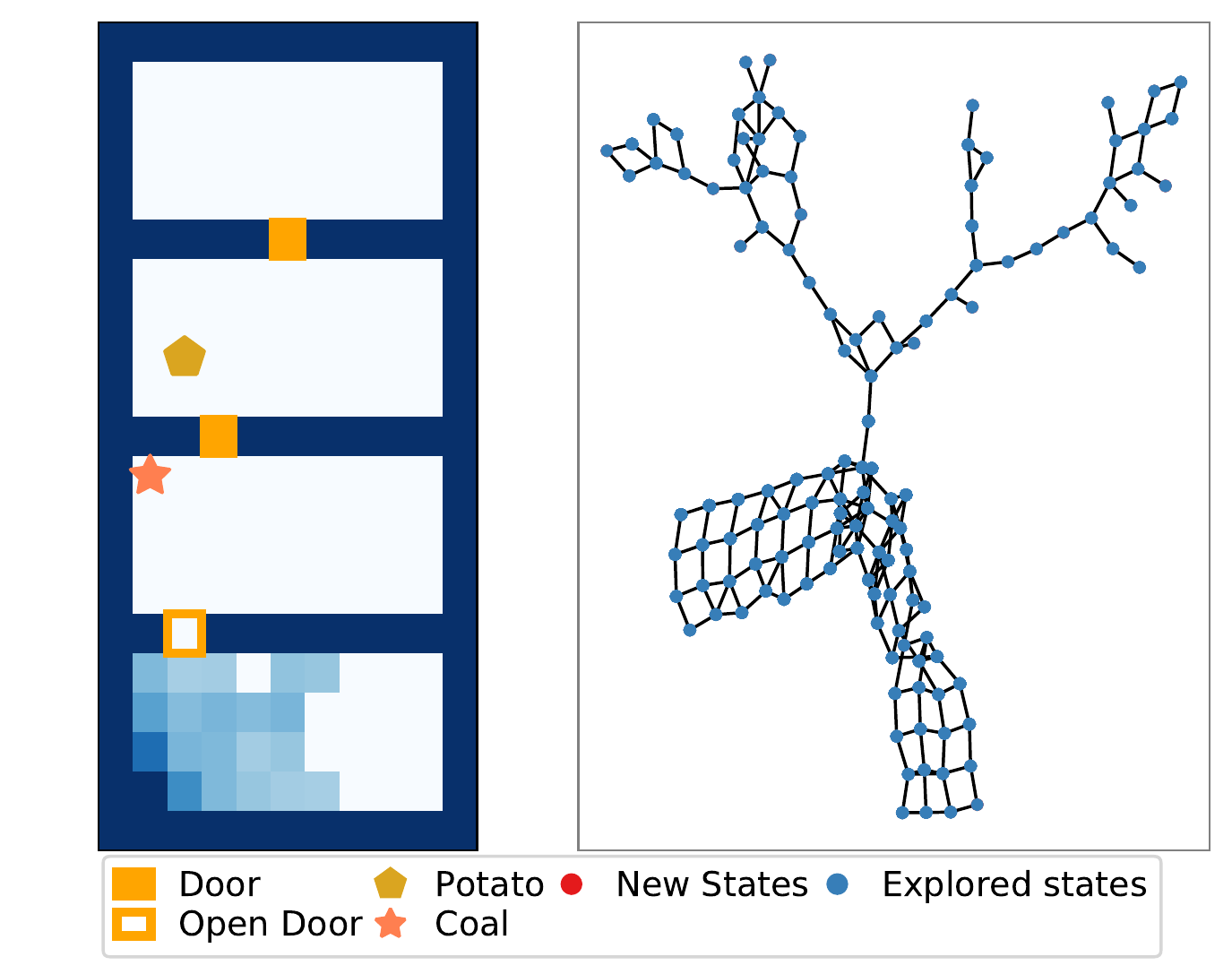}}%
	\vfill
	\subfloat[iteration 16]{\includegraphics[width=0.25\linewidth]{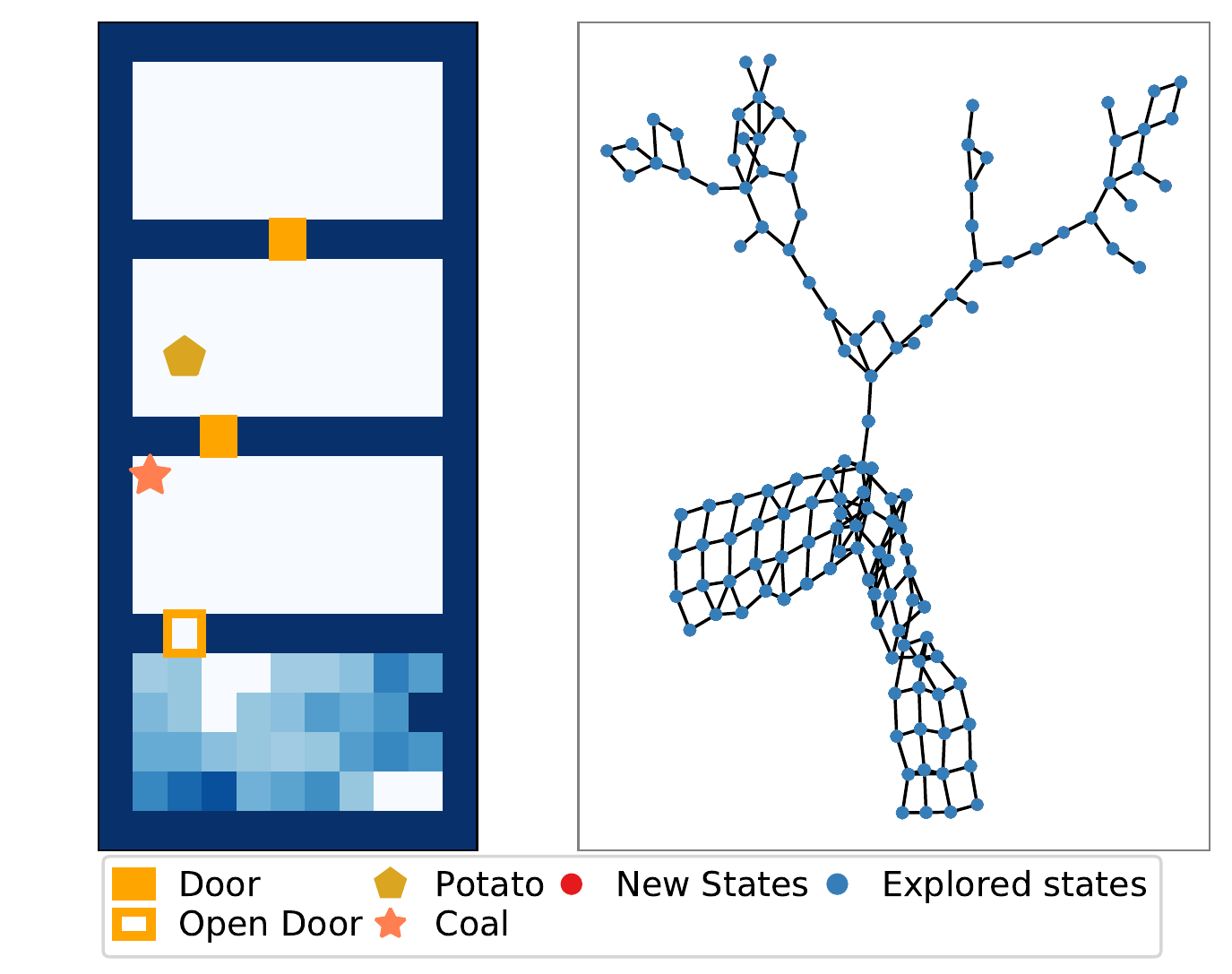}}%
	\hfill
	\subfloat[iteration 17]{\includegraphics[width=0.25\linewidth]{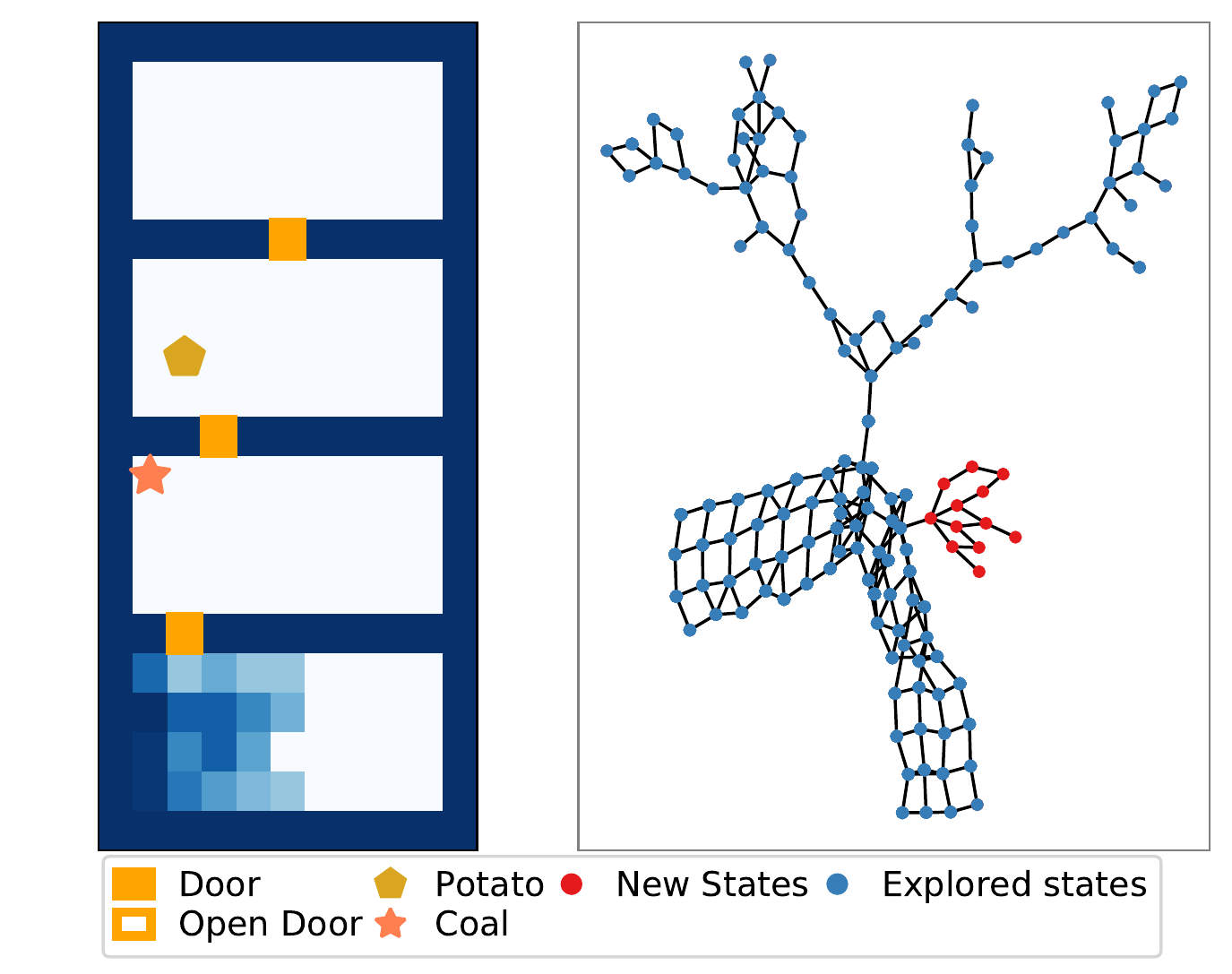}}%
	\hfill
	\subfloat[iteration 18]{\includegraphics[width=0.25\linewidth]{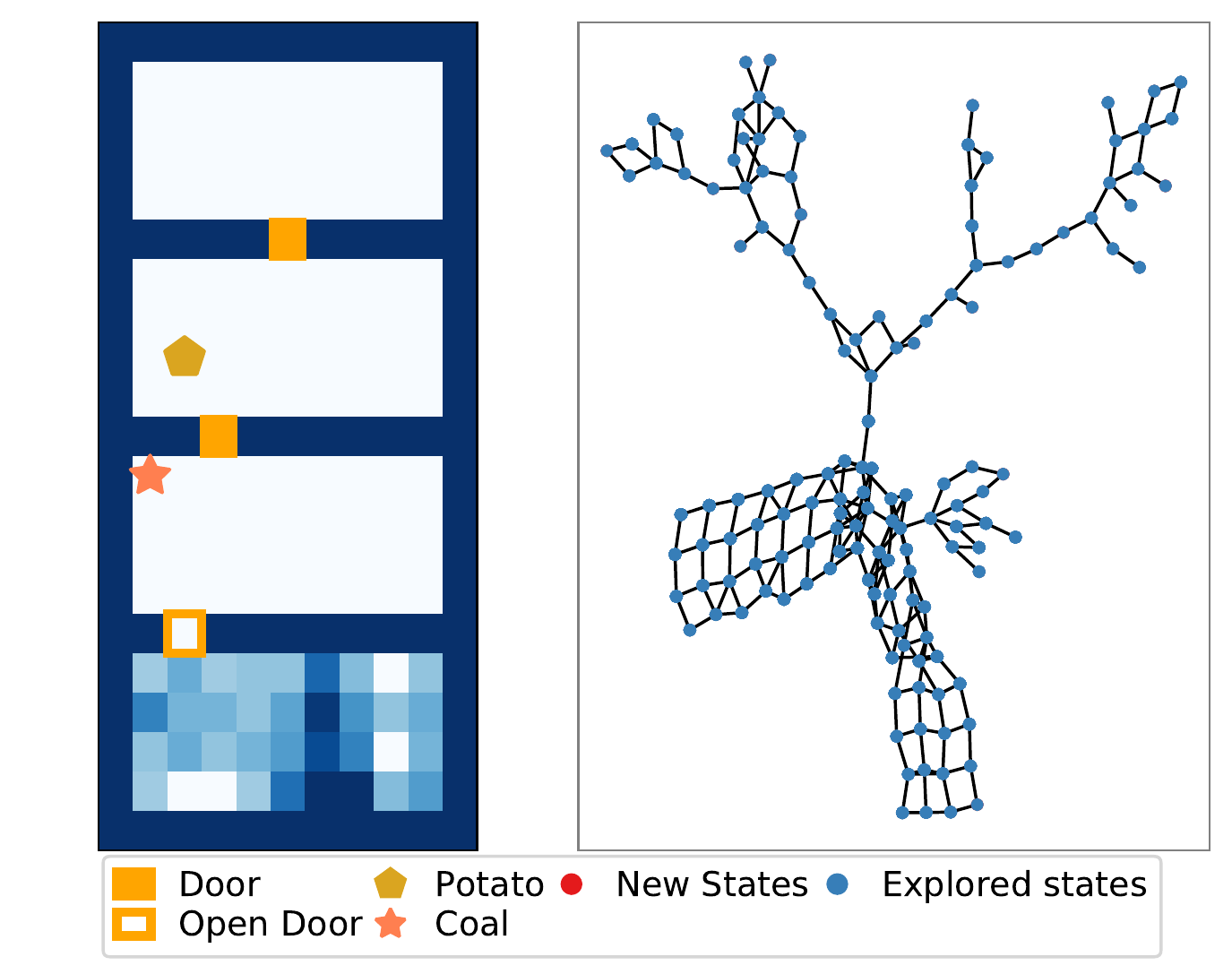}}%
	\hfill
	\subfloat[iteration 19]{\includegraphics[width=0.25\linewidth]{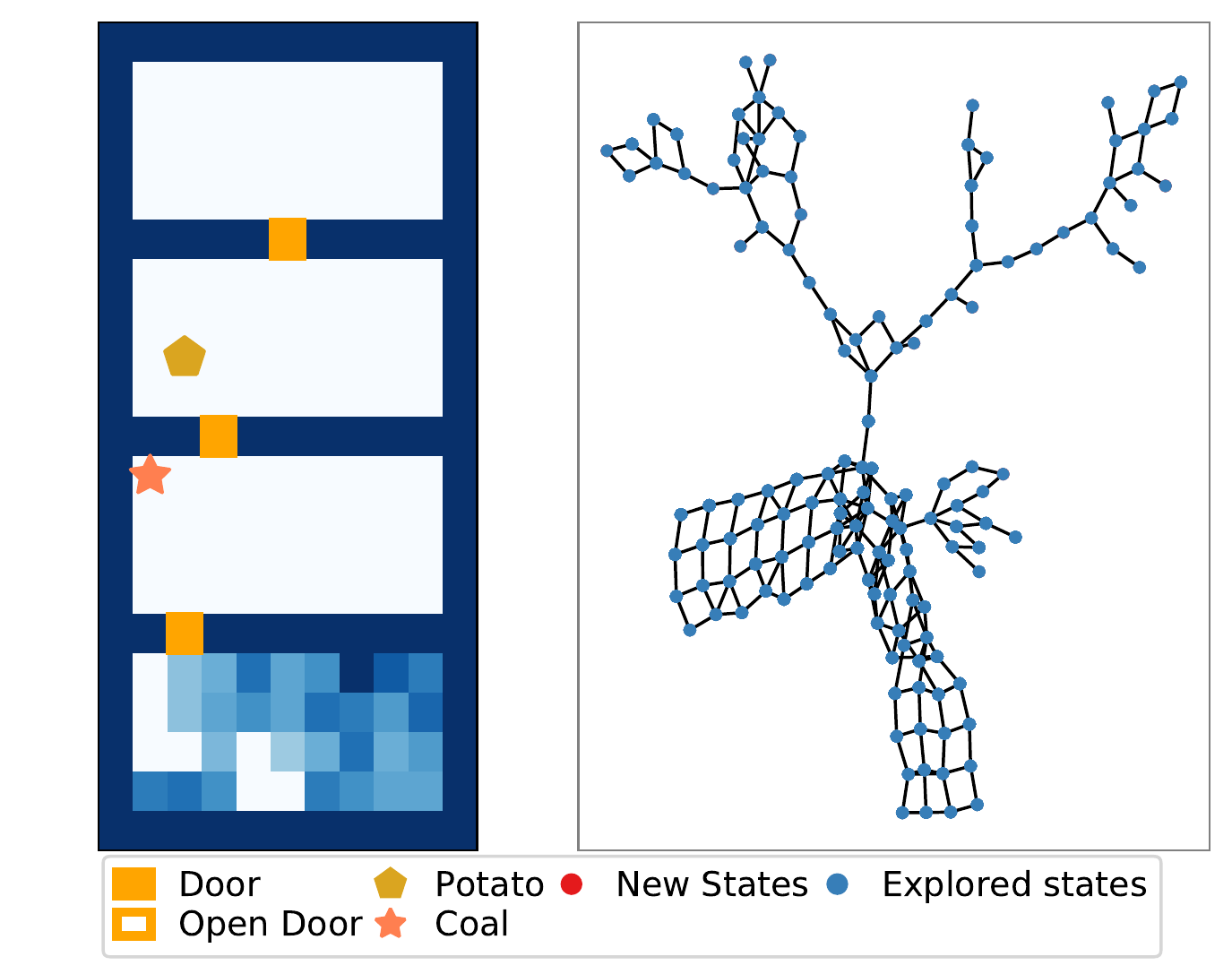}}%
	\hfill
    \caption{(Random Agent) State visitation frequencies and constructed graph per iteration in Minecraft Bake-Rooms}
    \label{fig:state-visitation-bake-rooms-random}
\end{figure*}

\textbf{More details on Minecraft Bake-Rooms:}
Besides Figure~\ref{fig:minecraft_potato}, we now present more details of our option grounding algorithm in environments with unknown transition dynamics in the Minecraft Bake-Rooms experiment. Figures~\ref{fig:state-visitation-bake-rooms-ours},~\ref{fig:state-visitation-bake-rooms-eigen} and~\ref{fig:state-visitation-bake-rooms-random} show the state visitation frequencies and MDP graph constructed over 20 iterations by our algorithm. The agent starts from the bottom room R1, a coal dropper is in R2, a potato dropper is in R3. The rooms are connected by doors which can be opened by the agent. For clarity of presentation, we show the undirected graph constructed. Blue nodes denote explored states and red nodes denote new states explored in the respective iteration. 

The shown figures demonstrate that our agent learns to open the door, and open the door and enter R2 to collect coal in iteration 2, while the eigenoptions agent learns to collect coal in iteration 18, and the random agent collects a coal block in iteration 14. Our agent learns to collect potato in R3 in iteration 3, while the eigenoptions agent learns this in iteration 19, and the random agent has not reached R3 within 20 iterations.

\begin{figure*}[t]
	\centering
	\hfill
	\subfloat[Abstract $\sf$-SMDP]{\includegraphics[width=0.4\linewidth]{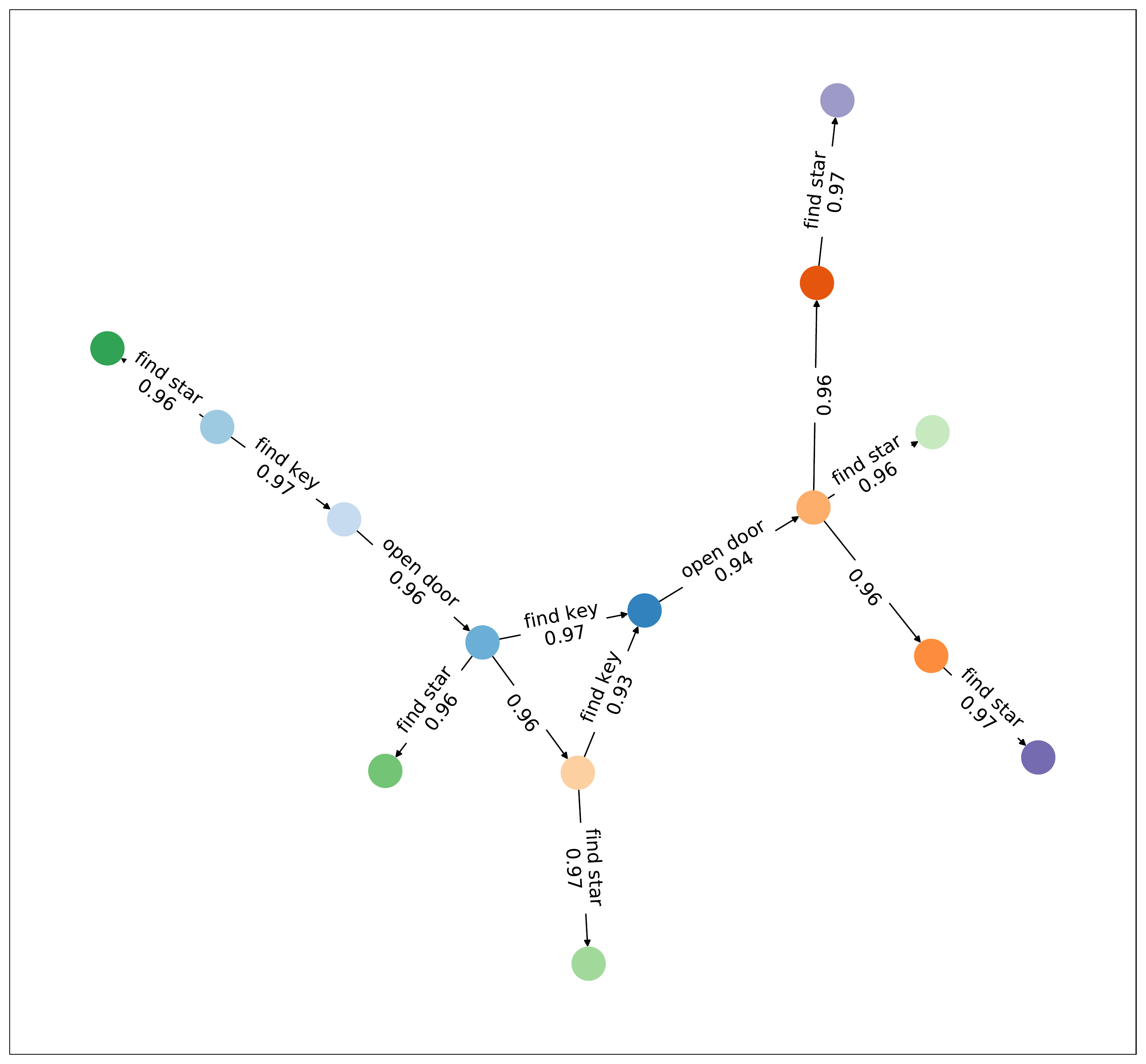}}%
	\hfill
	\subfloat[Ground MDP]{\includegraphics[width=0.4\linewidth]{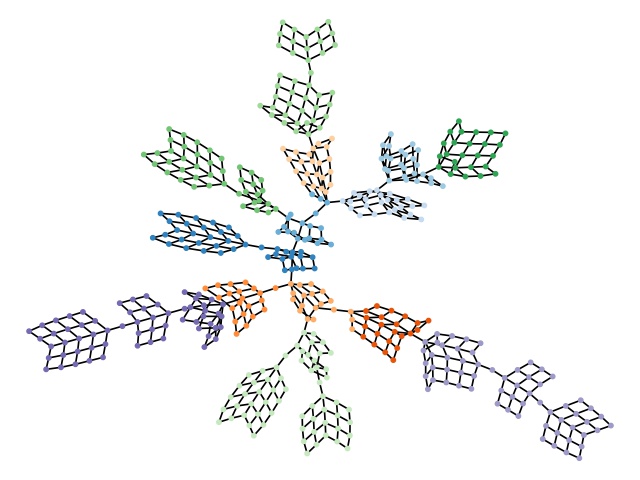}}%
	\hfill
	\vfill
	\hfill
	\subfloat[Q-Abstraction w. All Actions]{\includegraphics[width=0.4\linewidth]{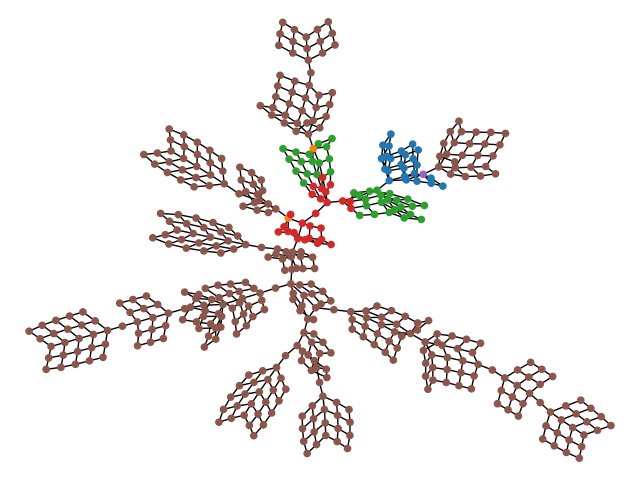}}%
	\hfill
	\subfloat[Q-Abstraction w. Optimal Action]{\includegraphics[width=0.4\linewidth]{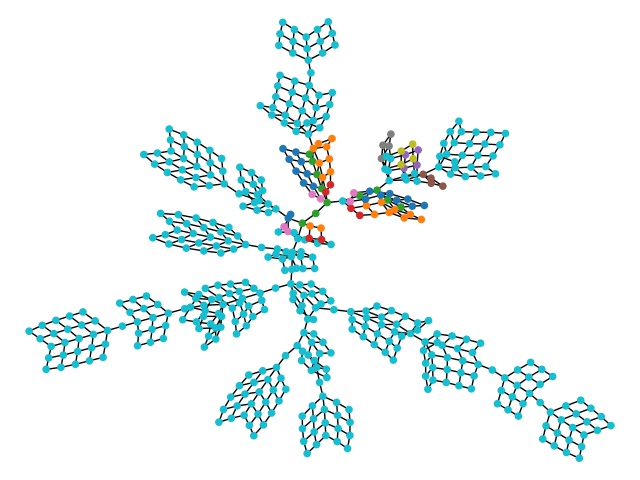}}%
	\hfill
	\caption{Abstract MDP with different abstraction schemes. (a) is the abstract MDP induced by our proposed successor homomorphism and (b) shows how the ground states are mapped to the abstract states. Node colors correspond to the abstract states in (a). (c) and (d) are abstraction induced by the $Q^*$-irrelevance and $a^*$-irrelevance abstraction schemes.}
	\label{fig:abstract_mdp_graphs_larger}
\end{figure*}

\textbf{Additional results on abstraction:}
Figure~\ref{fig:abstract_mdp_graphs_larger} shows the abstract MDPs in the Object-Rooms with $N=3$ rooms. Figure~\ref{fig:abstract_mdp_graphs_larger}(a) is the abstract $\sf$-SMDP model induced by our approximate successor homomorphism, the colors of the nodes match their corresponding ground states in the ground MDP shown in  Figure~\ref{fig:abstract_mdp_graphs_larger}(b). The edges with temporal semantics correspond to abstract successor options and the option transition dynamics. To avoid disconnect graphs, we can augment the abstract successor options with shortest path options, which connect ground states of disconnected abstract states to their nearest abstract states. Figure~\ref{fig:abstract_mdp_graphs_larger}(c) and (d) show the abstract MDPs induced by the $Q^*$-irrelevance (Q-all) and $a^*$-irrelevance (Q-optimal) abstraction methods, for the task \emph{find key}. The distance threshold $\epsilon = 0.1$.

\begin{figure*}[h]
	\centering
	\subfloat[no transfer (dense reward)]{\includegraphics[width=0.25\linewidth]{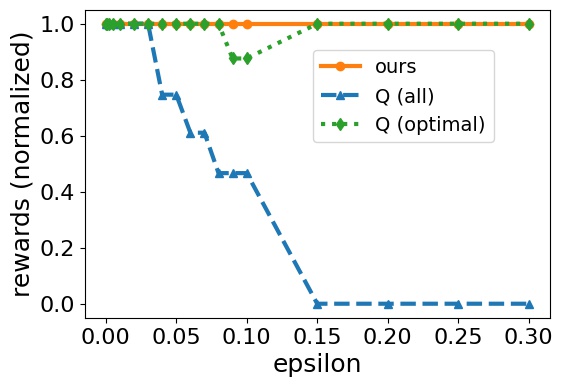}}%
	\hfill
	\subfloat[no transfer (sparse reward)]{\includegraphics[width=0.25\linewidth]{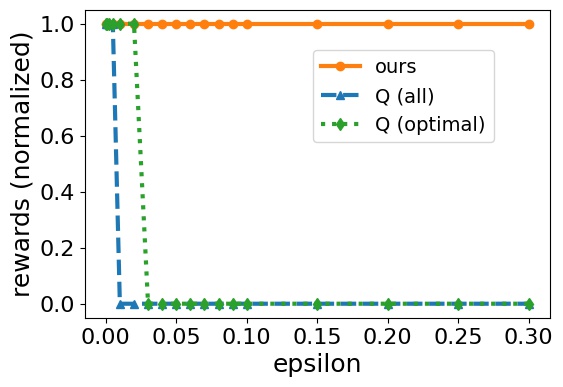}}%
	\hfill
	\subfloat[transfer (w. overlap)]{\includegraphics[width=0.25\linewidth]{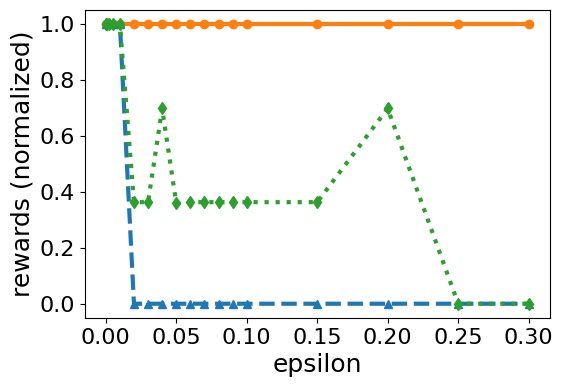}}%
	\hfill
	\subfloat[transfer (w.o. overlap)]{\includegraphics[width=0.25\linewidth]{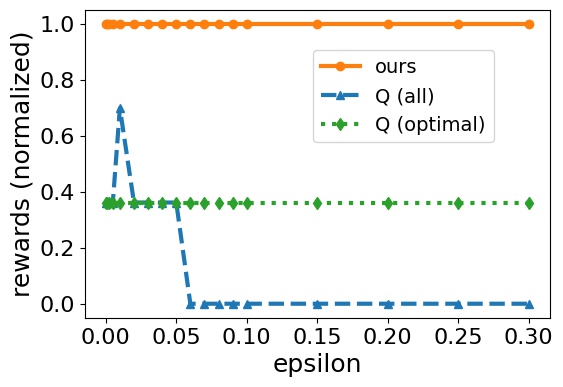}}%
	\vfill
	\subfloat[no transfer (dense reward)]{\includegraphics[width=0.25\linewidth]{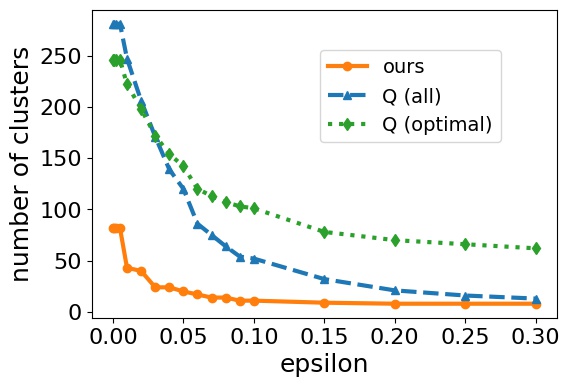}}%
	\hfill
	\subfloat[no transfer (sparse reward)]{\includegraphics[width=0.25\linewidth]{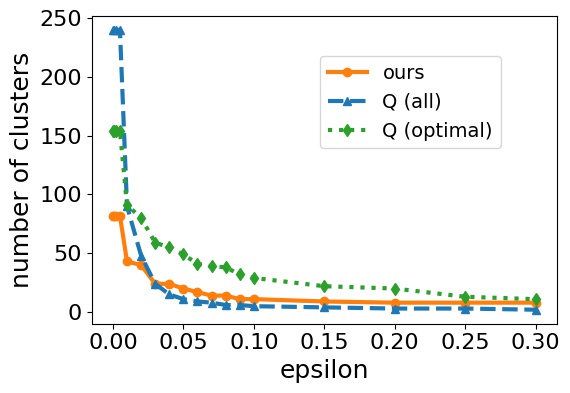}}%
	\hfill
	\subfloat[transfer (w. overlap)]{\includegraphics[width=0.25\linewidth]{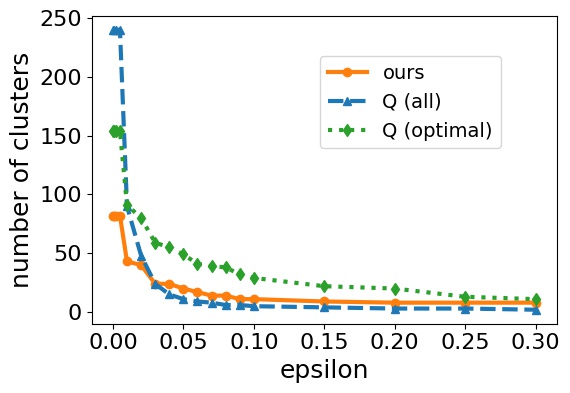}}%
	\hfill
	\subfloat[transfer (w.o. overlap)]{\includegraphics[width=0.25\linewidth]{img/abstraction/abstract_mdp_planning_4R/nclusters_transfer_overlap.jpeg}}%
	
	\caption{Performance of planning with the abstract MDPs. The upper row shows the total rewards (normalized by the maximum possible total rewards) obtained, and the lower row shows the corresponding number of abstract states of the abstract MDP. The x-axes are the distance thresholds $\epsilon$. transfer refers to task transfer (i.e., different reward function)}
	\label{fig:abstract_mdp_planning_4R}
\end{figure*}

Figure~\ref{fig:abstract_mdp_planning_4R} shows the results of using the abstract MDP for planning in the Object-Rooms with $N=4$ rooms. Please refer to Section~\ref{app:settings} for a detailed description of the settings. Our successor homomorphism model performs well across all tested settings with few abstract states (number of clusters). Since successor homomorphism does not depend on rewards, the abstract model can transfer across tasks (with varying reward functions), and is robust under sparse rewards settings. Whereas abstraction schemes based on the reward function perform worse when the source task for performing abstraction is different from the target task where the abstract MDP is used for planning.

\end{document}